\documentclass[smallextended]{svjour3}   
\smartqed

\usepackage{amsmath,amssymb,graphicx}
\usepackage{epsfig}
\usepackage{lscape}
\usepackage{footmisc}
\usepackage{times}
\usepackage{graphicx}
\usepackage{multirow}
\usepackage{lscape}

\begin{document}
\journalname{Computer Science Review}
\title{Background Subtraction in Real Applications: Challenges, Current Models and Future Directions}
\author{Thierry Bouwmans \and Belmar Garcia-Garcia}
\institute{T. Bouwmans \at
              Lab. MIA, Univ. La Rochelle, France \\
              Tel.: +05.46.45.72.02\\
              \email{tbouwman@univ-lr.fr}}
             
\date{Received: date / Accepted: date}

\maketitle

\begin{abstract}
Computer vision applications based on videos often require the detection of moving objects in their first step. Background subtraction is then applied in order to separate the background and the foreground. In literature, background subtraction is surely among the most investigated field in computer vision providing a big amount of publications. Most of them concern the application of mathematical and machine learning models to be more robust to the challenges met in videos. However, the ultimate goal is that the background subtraction methods developed in research could be employed in real applications like traffic surveillance. But looking at the literature, we can remark that there is often a gap between the current methods used in real applications and the current methods in fundamental research. In addition, the videos evaluated in large-scale datasets are not exhaustive in the way that they only covered a part of the complete spectrum of the challenges met in real applications. In this context, we attempt to provide the most exhaustive survey as possible on real applications that used background subtraction in order to identify the real challenges met in practice, the current used background models and to provide future directions. Thus, challenges are investigated in terms of camera, foreground objects and environments. In addition, we identify the background models that are effectively used in these applications in order to find potential usable recent background models in terms of robustness, time and memory requirements. 
\keywords{Background Subtraction \and Background Initialization \and Foreground Detection \and Visual Surveillance }
\end{abstract}

\section{Introduction}
\label{sec:Introduction}
With the rise of the different sensors, background initialization and background subtraction are widely employed in different computer vision applications based on video taken by fixed cameras. These applications involve a big variety of environments with different challenges and different kinds of moving foreground objects of interest. The most well-known and oldest applications are surely intelligent visual surveillance systems of human activities such as traffic surveillance of road, airport and maritime surveillance \cite{P7C1-6010}. But, detection of moving objects are also required for intelligent visual observation systems of animals and insects in order to study the behavior of the observed animals in their environment. However, it requires visual observation in natural environments such as forest, river, ocean and submarine environments with specific challenges. Other miscellaneous applications like optical motion capture, human-machine interaction system, vision-based hand gesture recognition, content-based video coding and background substitution also need in their first step either background initialization and background subtraction. Even if detection of moving objects is widely employed in these real application cases, no full survey can be found in literature that identifies, reviews and groups in one paper the current models and the challenges met in videos taken with fixed cameras for these applications. Furthermore, most of the research are done on large-scale datasets that often consist of videos which are taken in the aim of evaluation by researchers, and thus several challenging situations that appear in real cases are not covered. In addition, research focus on future directions with background subtraction methods with mathematical models, machine learning models, signal processing models and classification models: 1) statistical models \cite{P1C2-MOG-760}\cite{P1C2-MOG-769}\cite{P1C2-MOG-777}\cite{P1C2-MOG-778}, fuzzy models \cite{P2C2-10}\cite{P2C2-11} and Dempster-schafer models \cite{P7C1-10} for mathematical concepts; 2) subspace learning models either reconstructive \cite{P1C4-1}\cite{P1C4-23}\cite{P1C4-25}, discriminative \cite{P2C3-2}\cite{P2C3-3} and mixed \cite{P2C3-100}, robust subspace learning via matrix decomposition \cite{P3C1-PCP-940}\cite{P3C1-PCP-941}\cite{P3C1-PCP-942}\cite{P3C1-SO-14} or tensor decomposition \cite{P3C5-TRPCA-IT-1}\cite{P3C5-TRPCA-IT-3}\cite{P3C5-TRPCA-IT-3}), robust subspace tracking \cite{P3C1-PCP-1030}, support vector machines \cite{P1C3-1}\cite{P1C3-9}\cite{P1C3-101}\cite{P1C3-200}\cite{P1C3-201}, neural networks \cite{P1C5-410}\cite{P1C5-410-1}\cite{P1C5-412}\cite{P1C5-413}\cite{P1C5-414}, and deep learning \cite{P1C5-2100}\cite{P1C5-2110}\cite{P1C5-2150}\cite{P1C5-2162}\cite{P1C5-2163}\cite{P1C5-2163-1}, for machine learning concepts; 3) Wiener filter \cite{P6C2-Dataset-1}, Kalman filter \cite{P1C7-106}, correntropy filter \cite{P1C7-200}, and Chebychev  filter \cite{P1C7-300} for signal processing models; and 4) clustering algorithms \cite{P1C6-1}\cite{P1C6-100}\cite{P1C6-193}\cite{P1C6-194}\cite{P1C6-300} for classification models. Statistical, fuzzy and Dempster-Schafer models allow to handle imprecision, uncertainty and incompleteness in the data due the different challenges while machine learning concepts allow to learn the background pixel representation in an supervised or unsupervised manner. Signal processing models allow to estimate the background value and the classification models attempts to classify pixels as background or foreground. But, in practice, most of the authors in real applications employed basic techniques (temporal median \cite{P1C1-Median-21}\cite{P1C1-Median-40}\cite{P1C1-Median-60}, temporal histogram \cite{P1C1-Histogram-1}\cite{P1C1-Histogram-2}\cite{P1C1-Histogram-10}\cite{P1C1-Histogram-22-1}\cite{P1C1-Histogram-50} and filter \cite{P1C7-101}\cite{P1C7-102}\cite{P1C7-105}\cite{P1C7-106}\cite{P1C7-300}) and relative old techniques like MOG \cite{P1C2-MOG-10} published in 1999, codebook \cite{P1C6-100} in 2004 and ViBe \cite{P2C1-200} in 2009. This fact is due to two main reasons: 1) most of the time the recent advances can not be currently employed in real application cases due their time and memory requirements, and 2) there is also an ignorance in the communities of traffic surveillance and animals surveillance about the recent background subtraction methods with direct real-time ability with low computation and memory requirements.

To address the previous mentioned issues, we first attempt in this review to survey most of the real applications that used background initialization and subtraction in their process by classifying them in terms of aims, environments and objects of interests. We reviewed only the publications that specifically address the problem of background subtraction in real applications with experiments on corresponding videos. To have an overview about the fundamental research in this field, the reader can refer to numerous surveys on background initialization \cite{P0C0-Survey-30}\cite{P0C0-Survey-31}\cite{P0C0-Survey-32}\cite{P0C0-Survey-33} and background subtraction methods \cite{P0C0-Survey-10}\cite{P0C0-Survey-11}\cite{P0C0-Survey-12}\cite{P0C0-Survey-20}\cite{P0C0-Survey-21}\cite{P0C0-Survey-22}\cite{P0C0-Survey-23}\cite{P0C0-Survey-24}\cite{P0C0-Survey-25}. Furthermore, we also highlight recent background subtraction models that can be directly used in real applications. Finally, this paper is intended for researchers and engineers in the field of computer vision (i.e visual surveillance of human activities), and biologist/ethologist (i.e visual surveillance of animals and insects).

The rest of this paper is as follows. First, we provide in Section \ref{Overview} a short reminder on the different key points in background subtraction for novices. In Section \ref{P0C2-sec:Applications}, we provide a preliminary overview of the different real application cases in which background initialization and background subtraction are required. In Section \ref{P0C2:sec:IVS}, we review the background models and the challenges met in current intelligent visual surveillance systems of human activities such as road, airport and maritime surveillance. In Section \ref{P0C2:sec:AIB}, intelligent visual observation systems for animals and insects are reviewed in terms of the challenges related to the behavior of the observed animals and its environment. Then, in Section \ref{P0C2:sec:IVO}, we specifically investigate the challenges met in visual observation of natural environments such as forest, river, ocean and submarine environments. In Section \ref{P0C2:sec:OMA}, we survey other miscellaneous applications like optical motion capture, human-machine interaction system, vision-based hand gesture recognition, content-based video coding and background substitution. In Section \ref{Challenges}, we provide a discussion identifying the solved and unsolved challenges in these different application cases as well as proposing prospective solutions to address them. Finally, in Section \ref{Conclusion}, concluding remarks are given. 

\section{Background Subtraction: A Short Overview}
\label{Overview}
In this section, we remain briefly the aim of background subtraction for segmentation of static and moving foreground objects from a video stream. This task is the fundamental step in many visual surveillance applications for which background subtraction oﬀers a suitable solution which provide a good compromise in terms of quality of detection and computation time. The different steps of background subtraction methods as follows:
\begin{enumerate}
\item \textbf{Background initialization} (also called  \textit{background generation}, \textit{background extraction} and \textit{background reconstruction}) consists in computing the first background image. 
\item \textbf{Background Modeling} (also called \textit{Background Representation}) describes the model use to represent the background. 
\item \textbf{Background Maintenance} concerns the mechanism of update for the model to adapt itself to the changes which appear over time. 
\item \textbf{Classification of pixels in background/moving objects} (also called \textit{Foreground Detection}) consists in  classifying pixels in the class "background" or the class "moving objects".
\end{enumerate}

These different steps employ methods which have different aims and constraints. Thus, they need algorithms with different features. Background initialization requires \textit{"off-line"}  algorithms which are \textit{"batch"}  by taking all the data at one time. On the other hand, background maintenance needs \textit{"on-line"} algorithms which are \textit{"incremental"} algorithms by taking the incoming data one by one. Background initialization, modeling and maintenance require reconstructive algorithms while foreground detection needs discriminative algorithms. \\

A background subtraction process includes the following stages: \textbf{(1)} the background initialization module provides the first background image from $N$ training frames, \textbf{(2)} Foreground detection that
consists in classifying pixels as foreground or background, is  achieved by comparing the background image and the current image. 
\textbf{(3)} Background maintenance module updates the background image by using the previous background, the current image and the foreground detection mask. The steps \textbf{(2)} and \textbf{(3)} are executed repeatedly as time progresses. It is important to see that two images are compared and that for it, the methods compare a sub-entity of the entity background image with its corresponding sub-entity in the current image. This sub-entity can be of the size of a pixel, a region or a cluster. Furthermore, this sub-entity is characterized by a \textit{"feature"} which can be a color feature, edge feature, texture feature, stereo feature or motion feature \cite{P0C0-Survey-26}. Developing a background subtraction method, researchers and engineers must design each step and choose the features in relation to the challenges they want to handle in the concerned applications.

\section{A Preliminary Overview}
\label{P0C2-sec:Applications}
In real application cases, either background initialization and background subtraction are required in video taken by a static camera to generate a clean background image of the filmed scene or to detect static or moving foreground objects.

\subsection{Background Initialization based Applications}
Background initialization provides a a clean background from video sequence, and thus it is required for several applications as developed in Bouwmans et al. \cite{P0C0-Survey-32}:
\begin{enumerate}
\item \textbf{Video inpainting:} Video inpainting (also called video completion) tries to fill-in user defined spatio-temporal holes in a video sequence using information extracted in the existent spatio-temporal volume, according to consistency criteria evaluated both in time and space as in Colombari et al. \cite{P4C1-9}.
\item  \textbf{Privacy protection:} Privacy protection for videos aims to avoid the infringement on the privacy right of people taken in the many videos uploaded to video sharing services, that may contain privacy sensitive information of the people as in Nakashima et al. \cite{P0C0-A-950}.
\item  \textbf{Computational photography:} It concerns the case where the user wants to obtain a clean background plate from a set of input images containing cluttering foreground objects.
\end{enumerate}
These real application cases only need a clean background without detection of moving object. As the reader can found surveys for these applications in literature, we do not review them in this paper.

\subsection{Background Subtraction based Applications}
Segmentation of static and moving foreground objects from a video stream is the fundamental step in many computer vision applications for which background subtraction offers a suitable solution which provide a good compromise in terms of quality of detection and computation time.
\begin{enumerate}
\item \textbf{Visual Surveillance of Human Activities:}  The aim is to identify and track objects of interests in several environments. The most frequent environments are traffic scenes (also called road or highway scenes) for their analysis in order to detect incidents such as stopped vehicles on highways \cite{P0C0-A-9-31}\cite{P0C0-A-9-31-1}\cite{P0C0-A-9-32}\cite{P0C0-A-9-33} or to traffic density estimation on highways which can be then categorized as empty, fluid, heavy and jam. Thus, it is needed to detect and track vehicles \cite{P0C0-A-9-1}\cite{P0C0-A-9-30} or to count the number of vehicles \cite{P0C0-A-9-50}. Background subtraction can be also used for congestion detection \cite{P0C0-A-14}\cite{P0C0-A-7} in urban traffic surveillance, for illegal parking detection
\cite{P0C0-A-9-70}\cite{P0C0-A-9-71}\cite{P0C0-A-9-72}\cite{P0C0-A-9-73}\cite{P0C0-A-9-74} and for the detection of free parking places \cite{P0C0-A-9-75}\cite{P0C0-A-9-76}\cite{P0C0-A-9-77}. It is also important for security in train stations and airports, where unattended luggage can be a main goal. Human activities can be also monitored in maritime scenes to count the number of ships which circulated in a marina or in a harbor \cite{P7C1-3000}\cite{P7C1-3001}\cite{P7C1-3002}\cite{P7C1-3010}, and to detect and track ships in fluvial canals. Other environments are store scenes for the detection and tracking of consumers \cite{P0C0-A-30}\cite{P0C0-A-31}\cite{P0C0-A-32}\cite{P0C0-A-33}. 
\item  \textbf{Visual Observation of Animals and Insects Behaviors:} The system required for intelligent visual observation of animals and insects need to be simple and non-invasive. In this context, a video-based system is suitable to detect and track animals and insects in order to analyze their behavior which is needed \textbf{(1)} to evaluate their interaction inside their group such as in the case of honeybees which are social insects and interact extensively with their mates \cite{P0C0-A-70} and in the case of mice \cite{P3C1-RMC-91}\cite{P3C1-RMC-92}; \textbf{(2)} to have a fine-scale analysis of the interaction of the animals with their environment (plants,etc...) such as in the case of birds in order to have information about the impact of climate change on the ecosystem  \cite{P0C0-A-50}\cite{P0C0-A-51}\cite{P0C0-A-56}\cite{P0C0-A-57}; \textbf{(3)} to study the behavior of animals in different weather conditions such as in the case of fish in presence of typhoons, storms or sea currents for the Fish4Knowledge (F4K\protect\footnotemark[1]) \cite{P0C0-A-60}\cite{P0C0-A-61}\cite{P0C0-A-62}\cite{P0C0-A-63}; \textbf{(4)} for census of either endangered or threatened species like various species of fox, jaguar, mountain beaver, and wolf \cite{P7C1-2000}\cite{P7C1-2001}. In this case, the location and movements of these animals must be acquired, recorded, and made available for review; \textbf{(5)} for livestock and pigs \cite{P0C0-A-77} surveillance in farms; and \textbf{(6)} to design robots that mimics an animal’s locomotion such as in Iwatani et al. \cite{P0C0-A-760}. Detection of animals and insects also allows researchers in biology, ethology and green development to evaluate the impact of the considered animals or insects in the ecosystem, and to protect them such as honeybees that have crucial role in pollination across the world. 
\item \textbf{Visual Observation of Natural Environments:}  The aim is to detect foreign objects in natural environments such as forest, ocean and river to protect the biodiversity in terms of fauna and flora. For example, foreign objects in river and ocean can be floating bottles \cite{P1C7-104}, floating wood \cite{P0C0-A-90} \cite{P0C0-A-91} or mines \cite{P0C0-A-29}.
\item \textbf{Visual Analysis of Human Activities:} Background subtraction is also used in sport \textbf{(1)} when important decisions need to be made quickly as in soccer and in tennis with "Hawk-Eye\protect\footnotemark[2]". It has become a key part of the game; \textbf{(2)} for precise analysis of athletic performance, since it has no physical effect on the athlete as in Tamas et al. \cite{P0C0-A-500} for rowing motion and in John et al. \cite{P0C0-A-502} for aerobic routines; and  \textbf{(3)} for surveillance as in Bastos  \cite{P0C0-A-501} for surfers activities. 
\item \textbf{Visual Hull Computation:}  Visual hull is used for image synthesis to obtain an approximated geometric model of an object which can be static or not. In the first case, it allows a realistic model by provided an image-based model of objects. In the second case, it allows an image-based model of human which can be used for optical motion capture. Practically, visual hull is a geometric entity obtained with shape-from-silhouette 3D reconstruction technique introduced by Laurentini \cite{P0C0-A-400}. First, it uses background subtraction to separate the foreground object from the background to obtain the foreground mask known as a silhouette which is then considered as the 2D projection of the corresponding 3D foreground object. Based on the camera viewing parameters, the silhouette extracted in each view defines a back-projected generalized cone that contains the actual object. This cone is called a silhouette cone. The intersection of all the cones is called a visual hull, which is a bounding geometry of the actual 3D object. In practice, visual hull computation is employed for the following tasks:
\begin{itemize}
\item \textbf{Image-based Modeling:} Traditional modeling in image synthesis is made by a modeler but traditional modeling present several disadvantages: \textbf{(1)} it is a complex task and it requires time, \textbf{(2)} real data of objects are often not known, \textbf{(3)} specifying a practical scene is tedious, and  \textbf{(4)} it presents lack of realism. To address this problem, visual hull is used to extract the model of an object from different images. 
\item \textbf{Optical Motion Capture:} Optical motion capture systems are used \textbf{(1)} to give a character life and personality, and \textbf{(2)} to allow interactions between the real world and virtual world as in games or virtual reality. In the first application, an actor is filmed on up to $200$ cameras which monitor his movements precisely. Then, by using background subtraction, these movements are extracted and translated onto the character. In the second application, the gamer is filmed by a conventional camera or a RGB-D camera such as Microsoft's Kinect. His movements are tracked to interact with virtual objects. 
\end{itemize}
\item \textbf{Human-Machine Interaction (HMI):} In several applications, it requires human-machine interaction such as in arts \cite{P0C0-A-360}, games and ludo-applications \cite{P0C0-A-200}\cite{P0C0-A-201}\cite{P0C0-A-202}. In the case of games, the gamer can observe his own image or silhouette composed into a virtual scene as in PlayStation Eye-Toy. In the case of ludo-multimedia applications, several ones concern the detection of a selected moving object in a video by an user as in the project Aqu@theque \cite{P0C0-A-200}\cite{P0C0-A-201}\cite{P0C0-A-202} which requires to detect the selected fish in video and to recognize it in terms of species. 
\item \textbf{Vision-based Hand Gesture Recognition:}  This application requires to detect, track and recognize hand gesture for several applications such as human-computer interface, behavior studies, sign language interpretation and learning, teleconferencing, distance learning, robotics, games selection and object manipulation in virtual environments. 
\item \textbf{Content-based Video Coding:}  In video coding for transmission such as in teleconferencing, digital movies and video phones, only the key frames are transmitted with the moving objects. For example, the MPEG-4 multimedia communication standard enables the content-based functionality by using the video object plane (VOP) as the basic coding element. Each VOP includes the shape and texture information of a semantically meaningful object in the scene. New functionality like object manipulation and scene composition can be achieved because the video bit stream contains the object shape information. Thus, background subtraction can be used in content-based video coding.
\item \textbf{Background Substitution:} The aim of background substitution (also called background cut and video matting) is to extract the foreground from the input video and then combine it with a new background. Thus, background subtraction can be used in the first step as in Huang et al. \cite{P0C0-A-350}.  
\item \textbf{Miscellaneous applications:} Other applications used background subtraction such as carried baggage detection as in Tzanidou \cite{P0C0-A-600}, fire detection as in Toreyin et al. \cite{P0C0-A-700}, and OLED defect detection as in Wang et al. \cite{P0C0-A-900}.
\end{enumerate}

All these applications require the detection of moving objects in their first step, and possess their own characteristics in terms of challenges due to the location of the camera, the environment and the type of the moving objects. Background subtraction can be applied with one view or a multi-view as in  Diaz et al. \cite{P0C2-6003}. In addition, background subtraction can be also used in applications in which cameras are slowly moving \cite{P0C2-6000}\cite{P0C2-6001}\cite{P0C2-6005}. For example, Taneja et al. \cite{P0C2-6004} proposed to model dynamic scenes recorded with freely moving cameras. Extensions of background subtraction to moving cameras are presented in Yazdi and Bouwmans \cite{P0C3-700}. But, real applications with moving camera is out of the scope of this review as we limited this paper to applications with fixed camera. Table \ref{P0C2:Applications-1}, Table \ref{P0C2:Applications-2}, Table \ref{P0C2:Applications-3}, Table \ref{P0C2:Applications-4} and Table \ref{P0C2:Applications-5} show an overview of the different applications, the corresponding types of moving objects of interest, and specific characteristics.

\footnotetext[1]{http://groups.inf.ed.ac.uk/f4k/}
\footnotetext[2]{http://www.hawkeyeinnovations.co.uk/}

\begin{table}
\scalebox{0.79}{
\begin{tabular}{|l|l|l|} 
\hline
\scriptsize{Sub-categories-Aims} &\scriptsize{Objects of interest-Scenes} &\scriptsize{Authors-Dates} \\
\hline
\hline
\scriptsize{\textbf{1) Road Surveillance}}  &\scriptsize{\textbf{1-Cars}} 				 & \scriptsize{} 	 \\	
\cline{2-3}
\scriptsize{\textbf{1.1) Vehicles Detection}} &\scriptsize{}                      & \scriptsize{}   \\     
\scriptsize{Vehicles Detection} &\scriptsize{Road Traffic}                    & \scriptsize{Zheng et al. (2006) \cite{P0C0-A-1}}  \\
\scriptsize{Vehicles Detection} &\scriptsize{Urban Traffic (Korea)}  	& \scriptsize{Hwang et al. (2009) \cite{P0C0-A-9-48}}     \\ 
\scriptsize{Vehicles Detection} &\scriptsize{Highways Traffic (ATON Project)} & \scriptsize{Wang and Song (2011) \cite{P0C0-A-9-46}} \\ 
\scriptsize{Vehicles Detection} &\scriptsize{Aerial Videos (USA)}   & \scriptsize{Reilly et al. (2012) \cite{P0C0-A-9-79-13}} \\ 
\scriptsize{Vehicles Detection} &\scriptsize{Intersection (CPS) (China)} & \scriptsize{Ding et al. (2012) \cite{P0C0-A-9-44}}  \\ 
\scriptsize{Vehicles Detection} &\scriptsize{Intersection (USA)}  			 & \scriptsize{Hao et al. (2013) \cite{P0C0-A-9-45}}    \\ 
\scriptsize{Vehicles Detection} &\scriptsize{Intersection (Spain)}      & \scriptsize{Milla et al. (2013) \cite{P0C0-A-9-79-3}} \\  
\scriptsize{Vehicles Detection} &\scriptsize{Aerial Videos (VIVID Dataset \cite{P0C0-A-4-1})} &\scriptsize{Teutsch et al. (2014) \cite{P0C0-A-4}} \\
\scriptsize{Vehicles Detection} &\scriptsize{Road Traffic (CCTV cameras)(Korea)} & \scriptsize{Lee et al. (2014) \cite{P0C0-A-9-42}} \\ 
\scriptsize{Vehicles Detection} &\scriptsize{CD.net Dataset 2012 \cite{P6C2-Dataset-1000}} &\scriptsize{Hadi et al. (2014) \cite{P0C0-A-9}}  \\ 
\scriptsize{Vehicles Detection} &\scriptsize{Northern Jutland (Danemark)}   &\scriptsize{Alldieck (2015) \cite{P0C0-A-9-79-12}}   \\ 
\scriptsize{Vehicles Detection} &\scriptsize{Road Traffic (Weather) (Croatia)} &\scriptsize{Vujovic et al. (2014) \cite{P7C1-9001}} \\
\scriptsize{Vehicles Detection} &\scriptsize{Road Traffic (Night) (Hungary)} &\scriptsize{Lipovac et al. (2014) \cite{P7C1-9003}} \\
\scriptsize{Vehicles Detection} &\scriptsize{CD.net Dataset 2012 \cite{P6C2-Dataset-1000}} &\scriptsize{Aqel et al. (2015) \cite{P0C0-A-9-20}}   \\ 
\scriptsize{Vehicles Detection} &\scriptsize{CD.net Dataset 2012 \cite{P6C2-Dataset-1000}} &\scriptsize{Aqel et al. (2016) \cite{P0C0-A-9-21}}   \\ 
\scriptsize{Vehicles Detection} &\scriptsize{CD.net Dataset 2012 \cite{P6C2-Dataset-1000}} & \scriptsize{Wang et al. (2016) \cite{P0C0-A-6}}     \\ 
\scriptsize{Vehicles Detection} &\scriptsize{Urban Traffic (China)}  & \scriptsize{Zhang et al. (2016) \cite{P0C0-A-5}}         \\ 
\scriptsize{Vehicles Detection} &\scriptsize{CCTV cameras (India)}   & \scriptsize{Hargude and Idate (2016) \cite{P0C0-A-9-40}} \\ 
\scriptsize{Vehicles Detection} &\scriptsize{Intersection (USA)}  	  & \scriptsize{Li et al. (2016) \cite{P0C0-A-9-47}}         \\ 
\scriptsize{Vehicles Detection} &\scriptsize{Dhaka city (Bangladesh)} &\scriptsize{Hadiuzzaman et al. (2017) \cite{P0C0-A-9-79-20}} \\
\scriptsize{Vehicles Detection} &\scriptsize{Road Traffic (Weather) (Madrid/Tehran)} &\scriptsize{Ershadi et al. (2018) \cite{P7C1-9002}} \\
\cline{2-3}
\scriptsize{\textbf{1.2) Vehicles Detection/Tracking}} &\scriptsize{} & \scriptsize{}      \\     
\scriptsize{Vehicles Detection/Tracking} &\scriptsize{Urban Traffic/Highways Traffic (Portugal)} &\scriptsize{Batista et al. (2008) \cite{P0C0-A-9-30}} \\ 
\scriptsize{Vehicles Detection/Tracking} &\scriptsize{Intersection (China)}   &\scriptsize{Qu et al. (2010) \cite{P1C2-MOG-514}}  \\
\scriptsize{Vehicles Detection/Tracking} &\scriptsize{Downtown Traffic (Night) (China)} &\scriptsize{Tian et al. (2013) \cite{P7C1-9005}}  \\ 
\scriptsize{Vehicles Detection/Tracking} &\scriptsize{Urban Traffic (China)}  &\scriptsize{Ling et al. (2014) \cite{P0C0-A-3}}     \\ 
\scriptsize{Vehicles Detection/Tracking} &\scriptsize{Highways Traffic (India)} &\scriptsize{Sawalakhe and Metkar (2015) \cite{P0C0-A-9-43}} \\ 
\scriptsize{Vehicles Detection/Tracking} &\scriptsize{Highways Traffic (India)} &\scriptsize{Dey and Praveen (2016) \cite{P0C0-A-9-41}}  \\ 
\scriptsize{Multi-Vehicles Detection/Tracking} &\scriptsize{CD.net Dataset 2012 \cite{P6C2-Dataset-1000}} &\scriptsize{Hadi et al. (2016) \cite{P0C0-A-9-1}} \\  
\scriptsize{Vehicles Tracking} &\scriptsize{NYCDOT video/NGSIM US-101 highway dataset(USA)} & \scriptsize{Li et al. (2016) \cite{P0C0-A-9-44-1}} \\
\cline{2-3}
\scriptsize{\textbf{1.3) Vehicles Counting}} &\scriptsize{} & \scriptsize{}      \\     
\scriptsize{Vehicles Counting/Classification} &\scriptsize{Donostia-San Sebastian (Spain)} & \scriptsize{Unzueta et al. (2012) \cite{P0C0-A-2}} \\
\scriptsize{Vehicles Detection/Counting}      &\scriptsize{Road (Portugal)}             &\scriptsize{Toropov et al. (2015) \cite{P0C0-A-9-79-1}} \\
\scriptsize{Vehicles Counting}      &\scriptsize{Lankershim Boulevard dataset (USA)}    &\scriptsize{Quesada and Rodriguez (2016) \cite{P3C1-PCP-1066}} \\
\cline{2-3}
\scriptsize{\textbf{1.4) Stopped Vehicles}} &\scriptsize{} & \scriptsize{}      \\     
\scriptsize{Stopped Vehicles} &\scriptsize{Portuguese Highways Traffic (24/7)} &\scriptsize{Monteiro et al. (2008) \cite{P0C0-A-9-31}}     \\ 
\scriptsize{Stopped Vehicles} &\scriptsize{Portuguese Highways Traffic (24/7)} &\scriptsize{Monteiro et al. (2008) \cite{P0C0-A-9-31-1}}   \\ 
\scriptsize{Stopped Vehicles} &\scriptsize{Portuguese Highways Traffic (24/7)} &\scriptsize{Monteiro  et al. (2008) \cite{P0C0-A-9-32}}    \\ 
\scriptsize{Stopped Vehicles} &\scriptsize{Portuguese Highways Traffic (24/7)} &\scriptsize{Monteiro  (2009)  \cite{P0C0-A-9-33}}   \\ 
\cline{2-3}
\scriptsize{\textbf{1.5) Congestion Detection}} &\scriptsize{} & \scriptsize{}      \\    
\cline{2-3} 
\scriptsize{}                 &\scriptsize{\textbf{1-Cars}}            &\scriptsize{}  \\
\scriptsize{Congestion Detection}         &\scriptsize{Aerial Videos}  & \scriptsize{Lin et al. (2009) \cite{P0C0-A-14}}          \\
\scriptsize{Free-Flow/Congestion Detection}  &\scriptsize{Urban Traffic (India)} &\scriptsize{Muniruzzaman et al. (2016) \cite{P0C0-A-7}}    \\ 
\cline{2-3} 
\scriptsize{}                 &\scriptsize{\textbf{2-Motorcycles (Motorbikes)}}                       &\scriptsize{}  \\
\scriptsize{Helmet Detection} &\scriptsize{Public Roads (Brazil)}     &\scriptsize{Silva et al. (2013) \cite{P0C0-A-9-49-1}}       \\
\scriptsize{Helmet Detection} &\scriptsize{Naresuan University Campus (Thailand)}  &\scriptsize{Waranusast et al. (2013) \cite{P0C0-A-9-49-1}}  \\
\scriptsize{Helmet Detection} &\scriptsize{Indian Institute of Technology Hyderabad (India)} &\scriptsize{Dahiya et al. (2016) \cite{P0C0-A-9-49}}  \\
\cline{2-3} 
\scriptsize{}                 &\scriptsize{\textbf{3-Pedestrians}}                          &\scriptsize{}  \\
\scriptsize{Pedestrian Abnormal Behavior} &\scriptsize{Public Roads (China)} &\scriptsize{Jiang et al. (2015) \cite{P0C0-A-9-79-5}} \\
\hline
\end{tabular}}
\caption{Intelligent Visual Surveillance of Human Activities: An Overview (Part I)} \centering
\label{P0C2:Applications-1}
\end{table}

\begin{table}
\scalebox{0.79}{
\begin{tabular}{|l|l|l|} 
\hline
\scriptsize{Categories-Aims} &\scriptsize{Objects of interest-Scenes} &\scriptsize{Authors-Dates} \\
\hline
\hline
\scriptsize{\textbf{Airport Surveillance}}        &\scriptsize{\textbf{1-Airplane}} 				 & \scriptsize{} \\
\scriptsize{AVITRACK Project (European)} &\scriptsize{Airport’s Apron}  & \scriptsize{Blauensteiner and Kampel (2004) \cite{P0C0-A-10}}  \\
\scriptsize{AVITRACK Project (European)} &\scriptsize{Airport’s Apron}  & \scriptsize{Aguilera (2005) \cite{P0C0-A-11}} \\
\scriptsize{AVITRACK Project (European)} &\scriptsize{Airport’s Apron}  & \scriptsize{Thirde (2006) \cite{P0C0-A-12}}   \\
\scriptsize{AVITRACK Project (European)} &\scriptsize{Airport’s Apron}  & \scriptsize{Aguilera (2006) \cite{P0C0-A-13}}  \\
\cline{2-3}
\scriptsize{}                &\scriptsize{\textbf{2-Ground Vehicles (Fueling vehicles/Baggage cars)}}  & \scriptsize{}         \\
\scriptsize{AVITRACK Project (European)} &\scriptsize{Airport’s Apron}  & \scriptsize{Blauensteiner and Kampel (2004) \cite{P0C0-A-10}}  \\
\scriptsize{AVITRACK Project (European)} &\scriptsize{Airport’s Apron}  & \scriptsize{Aguilera (2005) \cite{P0C0-A-11}} \\
\scriptsize{AVITRACK Project (European)} &\scriptsize{Airport’s Apron}  & \scriptsize{Thirde (2006) \cite{P0C0-A-12}}   \\
\scriptsize{AVITRACK Project (European)} &\scriptsize{Airport’s Apron}  & \scriptsize{Aguilera (2006) \cite{P0C0-A-13}}  \\ 
\cline{2-3}
\scriptsize{}                            &\scriptsize{\textbf{3-People (Workers)}}& \scriptsize{}  \\ 
\scriptsize{AVITRACK Project (European)} &\scriptsize{Airport’s Apron} & \scriptsize{Blauensteiner and Kampel (2004) \cite{P0C0-A-10}}  \\
\scriptsize{AVITRACK Project (European)}	&\scriptsize{Airport’s Apron}  & \scriptsize{Aguilera (2005) \cite{P0C0-A-11}}  \\
\scriptsize{AVITRACK Project (European)} &\scriptsize{Airport’s Apron}  & \scriptsize{Thirde (2006) \cite{P0C0-A-12}}    \\
\scriptsize{AVITRACK Project (European)} &\scriptsize{Airport’s Apron}  & \scriptsize{Aguilera (2006) \cite{P0C0-A-13}}   \\
\cline{1-3}                                             
\scriptsize{\textbf{Maritime Surveillance}} &\scriptsize{\textbf{1-Cargos}}    & \scriptsize{} \\
\scriptsize{}      			&\scriptsize{Ocean at Miami (USA)} & \scriptsize{Culibrk et al. (2006) \cite{P1C5-70}}         \\
\scriptsize{}      	    &\scriptsize{Harbor Scenes (Ireland)} & \scriptsize{Zhang et al. (2012) \cite{P0C0-A-29-72}}   \\
\cline{2-3}
\scriptsize{}           &\scriptsize{\textbf{2-Boats}}     & \scriptsize{}                                                      \\
\scriptsize{Stationary Camera}   &\scriptsize{Miami Canals (USA)}   & \scriptsize{Socek et al. (2005)  \cite{P0C0-A-29-1}}      \\
\scriptsize{Dock Inspecting Event}     &\scriptsize{Harbor Scenes (China)} & \scriptsize{Ju et al. (2008)  \cite{P1C2-MOG-313}}  \\
\scriptsize{Different Kinds of Targets} &\scriptsize{Baichay Beach (Vietnam)} & \scriptsize{Tran and Le (2016) \cite{P0C0-A-26}}  \\
\scriptsize{Salient Events (Coastal environments)} &\scriptsize{Nantucket Island (USA)} & \scriptsize{Cullen et al. (2012) \cite{P0C0-A-27}}   \\
\scriptsize{Salient Events (Coastal environments)} &\scriptsize{Nantucket Island (USA)} & \scriptsize{Cullen  (2012) \cite{P0C0-A-28}} \\
\scriptsize{Boat ramps surveillance} &\scriptsize{Boat ramps (New Zealand)} & \scriptsize{Pang et al. (2016) \cite{P7C1-9000}} \\
\cline{2-3}
\scriptsize{}                          &\scriptsize{\textbf{3-Sailboats}}          &\scriptsize{}  \\
\scriptsize{Sailboats Detection} &\scriptsize{UCSD Background Subtraction Dataset} & \scriptsize{Sobral et al. (2015)  \cite{P3C1-PCP-917}}   \\
\cline{2-3}
\scriptsize{}                      &\scriptsize{\textbf{4-Ships}}                     &\scriptsize{}  \\
\scriptsize{}            &\scriptsize{Italy}              & \scriptsize{Bloisi et al. (2014) \cite{P0C0-A-20}}      \\
\scriptsize{}            &\scriptsize{Fixed ship-borne camera (China) (IR)} & \scriptsize{Liu et al. (2014) \cite{P0C0-A-21}}  \\
\scriptsize{Different Kinds of Targets} &\scriptsize{Ocean (South Africa)}  & \scriptsize{Szpak and Tapamo (2011) \cite{P0C0-A-29-4}}\\
\scriptsize{Cage Aquaculture}  &\scriptsize{Ocean (Taiwan)}           & \scriptsize{Hu et al. (2011) \cite{P0C0-A-29-5}}         \\
\scriptsize{}                  &\scriptsize{Ocean (Korea)}            & \scriptsize{Arshad et al. (2010) \cite{P0C0-A-29-6}}     \\
\scriptsize{}                  &\scriptsize{Ocean (Korea)}            & \scriptsize{Arshad et al. (2011) \cite{P0C0-A-29-61}}    \\
\scriptsize{}                  &\scriptsize{Ocean (Korea)}            & \scriptsize{Arshad et al. (2014) \cite{P0C0-A-29-62}}    \\
\scriptsize{}                  &\scriptsize{Ocean (Korea)}            & \scriptsize{Saghafi et al. (2012) \cite{P0C0-A-29-7}}    \\
\scriptsize{Overloaded Ship Identification}  &\scriptsize{Ocean (China)}  & \scriptsize{Xie et al. (2012) \cite{P0C0-A-29-71}}      \\
\scriptsize{Ship-Bridge Collision} &\scriptsize{Wuhan Yangtze River (China)} & \scriptsize{Zheng et al. (2013) \cite{P0C0-A-29-73}} \\
\scriptsize{}                  &\scriptsize{Wuhan Yangtze River (China)} & \scriptsize{Mei et al. (2017) \cite{P3C4-DGS-16}} \\
\cline{2-3}
\scriptsize{}          & \scriptsize{\textbf{5-Motor Vehicles}}     &\scriptsize{}   \\
\scriptsize{Salient Events (Coastal environments)} &\scriptsize{Nantucket Island (USA)} & \scriptsize{Cullen et al. (2012) \cite{P0C0-A-27}} \\
\scriptsize{Salient Events (Coastal environments)} &\scriptsize{Nantucket Island (USA)} & \scriptsize{Cullen  (2012) \cite{P0C0-A-28}}       \\
\cline{2-3}
\scriptsize{}                          &\scriptsize{\textbf{6-People (Shoreline)}}      & \scriptsize{}  \\
\scriptsize{Salient Events (Coastal environments)} &\scriptsize{Nantucket Island (USA)} & \scriptsize{Cullen et al. (2012) \cite{P0C0-A-27}} \\
\scriptsize{Salient Events (Coastal environments)} &\scriptsize{Nantucket Island (USA)} & \scriptsize{Cullen  (2012) \cite{P0C0-A-28}}       \\
\cline{2-3}
\scriptsize{}      										  &\scriptsize{\textbf{7-Floating Objects}}  & \scriptsize{}  \\
\scriptsize{Detection of Drifting Mines} &\scriptsize{Floating Test Targets (IR)} & \scriptsize{Borghgraef et al. (2010) \cite{P0C0-A-29}}   \\
\cline{1-3} 
\scriptsize{\textbf{Store Surveillance}}  &\scriptsize{\textbf{People}}   & \scriptsize{} \\
\scriptsize{Apparel Retail Store}  &\scriptsize{Panoramic Camera}  & \scriptsize{Leykin and Tuceryan (2005) \cite{P0C0-A-31}}  \\  
\scriptsize{Apparel Retail Store}  &\scriptsize{Panoramic Camera}  & \scriptsize{Leykin and Tuceryan (2005) \cite{P0C0-A-32}}  \\  
\scriptsize{Apparel Retail Store}  &\scriptsize{Panoramic Camera}  & \scriptsize{Leykin and Tuceryan (2007) \cite{P0C0-A-30}}  \\  
\scriptsize{Retail Store Statistics}  &\scriptsize{Top View Camera}   & \scriptsize{Avinash et al. (2012) \cite{P0C0-A-33}}    \\  
\hline
\end{tabular}}
\caption{Intelligent Visual Surveillance of Human Activities: An Overview (Part II)} \centering
\label{P0C2:Applications-2}
\end{table}

\begin{table}
\scalebox{0.70}{
\begin{tabular}{|l|l|l|} 
\hline
\scriptsize{Categories-Aims} &\scriptsize{Objects of interest-Scenes} &\scriptsize{Authors-Dates} \\
\hline
\hline
\scriptsize{\textbf{Birds Surveillance}}        &\scriptsize{\textbf{Birds}}      & \scriptsize{}   \\  
\scriptsize{Feeder Stations in natural habitats} &\scriptsize{Feeder Station Webcam/Camcorder Datasets}  & \scriptsize{Ko et al. (2008) \cite{P0C0-A-50}}   \\
\scriptsize{Feeder Stations in natural habitats} &\scriptsize{Feeder Station Webcam/Camcorder Datasets} & \scriptsize{Ko et al. (2010) \cite{P0C0-A-51}}	  \\  
\scriptsize{Seabirds}                 &\scriptsize{Cliff Face Nesting Sites}           & \scriptsize{Dickinson et al. (2008) \cite{P0C0-A-56}}	 \\ 
\scriptsize{Seabirds}                 &\scriptsize{Cliff Face Nesting Sites}           & \scriptsize{Dickinson et al. (2010) \cite{P0C0-A-57}}	 \\ 
\scriptsize{Observation in the air}   &\scriptsize{Lakes in Northern Alberta (Canada)} & \scriptsize{Shakeri and Zhang (2012) \cite{P0C0-A-55}} \\ 
\scriptsize{Wildlife@Home}            &\scriptsize{Natural Nesting Stations}           & \scriptsize{Goehner et al. (2015) \cite{P0C0-A-79}}	   \\ 
\cline{1-3}  
\scriptsize{\textbf{Fish Surveillance}} &\scriptsize{\textbf{Fish}}                    & \scriptsize{}  \\
\scriptsize{\textbf{1-Tank}}            &\scriptsize{}                                 & \scriptsize{}  \\
\scriptsize{\textbf{1.1-Ethology}}       &\scriptsize{}                                & \scriptsize{}  \\
\scriptsize{Aqu@theque Project}  & \scriptsize{Aquarium of La Rochelle}  & \scriptsize{Penciuc et al. (2006) \cite{P0C0-A-202}}   \\
\scriptsize{Aqu@theque Project}  & \scriptsize{Aquarium of La Rochelle}  & \scriptsize{Baf et al. (2007) \cite{P0C0-A-201}}       \\
\scriptsize{Aqu@theque Project}  & \scriptsize{Aquarium of La Rochelle}  & \scriptsize{Baf et al. (2007) \cite{P0C0-A-200}}       \\
\scriptsize{\textbf{1.2-Fishing}} &\scriptsize{}                          & \scriptsize{}  \\
\scriptsize{Fish Farming}        & \scriptsize{Japanese rice fish}       & \scriptsize{Abe et al. (2016) \cite{P0C0-A-69-3}}       \\
\scriptsize{\textbf{2-Open Sea}}  &\scriptsize{}                          & \scriptsize{}  \\
\scriptsize{\textbf{2.1-Census/Ethology}} &\scriptsize{}                  & \scriptsize{}  \\
\scriptsize{EcoGrid Project (Taiwan)}  &\scriptsize{Ken-Ding sub-tropical coral reef} & \scriptsize{Spampinato et al. (2008) \cite{P0C0-A-60}}  \\
\scriptsize{EcoGrid Project (Taiwan)}  &\scriptsize{Ken-Ding sub-tropical coral reef} & \scriptsize{Spampinato et al. (2010) \cite{P0C0-A-61}}	 \\  
\scriptsize{Fish4Knowledge (European)} &\scriptsize{Taiwan’s coral reefs} & \scriptsize{Kavasidis and Palazzo (2012) \cite{P0C0-A-68-1}}        \\
\scriptsize{Fish4Knowledge (European)} &\scriptsize{Taiwan’s coral reefs} & \scriptsize{Spampinato et al. (2014) \cite{P0C0-A-62}}  \\
\scriptsize{Fish4Knowledge (European)} &\scriptsize{Taiwan’s coral reefs} & \scriptsize{Spampinato et al. (2014) \cite{P0C0-A-63}}  \\
\scriptsize{UnderwaterChangeDetection (European)} &\scriptsize{Underwater Scenes (Germany)} & \scriptsize{Radolko et al. (2016) \cite{P6C2-Dataset-2025}} \\
\scriptsize{-}                         &\scriptsize{Simulated Underwater Environment} & \scriptsize{Liu et al. (2016) \cite{P0C0-A-65}}         \\ 
\scriptsize{Fish4Knowledge (European)} &\scriptsize{Taiwan’s coral reefs}     & \scriptsize{Seese et al. (2016) \cite{P0C0-A-68}}  \\
\scriptsize{Fish4Knowledge (European)} &\scriptsize{Taiwan’s coral reefs}     & \scriptsize{Rout et al. (2017) \cite{P1C2-MOG-776}}  \\
\scriptsize{\textbf{2.2-Fishing}}      &\scriptsize{}                         & \scriptsize{}  \\
\scriptsize{Rail-based fish catching}  &\scriptsize{Open Sea Environment}     & \scriptsize{Huang et al. (2016) \cite{P0C0-A-67}}   \\
\scriptsize{Fish Length Measurement} &\scriptsize{Chute Multi-Spectral Dataset \cite{P0C0-A-69-2}} & \scriptsize{Huang et al. (2016) \cite{P0C0-A-69-2}}  \\
\scriptsize{Fine-Grained Fish Recognition}  &\scriptsize{Cam-Trawl Dataset \cite{P0C0-A-69-1}/Chute Multi-Spectral Dataset \cite{P0C0-A-69-2}} & \scriptsize{Wang et al. (2016) \cite{P0C0-A-69}}   \\
\cline{1-3}  
\scriptsize{\textbf{Dolphins Surveillance}}  &\scriptsize{\textbf{Dolphins}}  & \scriptsize{}  \\
\scriptsize{Social marine mammals}  &\scriptsize{Open sea environments}  & \scriptsize{Karnowski et al. (2015) \cite{P0C0-A-66}}  \\
\cline{1-3} 
\scriptsize{\textbf{Lizards Surveillance}}  &\scriptsize{\textbf{Lizards}}  & \scriptsize{}  \\
\scriptsize{Endangered lizard  species}  &\scriptsize{Natural environments}  & \scriptsize{Nguwi et al. (2016) \cite{P0C0-A-750}}  \\
\cline{1-3} 
\scriptsize{\textbf{Mice Surveillance}}  &\scriptsize{\textbf{Mice}}  & \scriptsize{}  \\
\scriptsize{Social behavior}  &\scriptsize{Caltech mice dataset \cite{P6C2-Dataset-20000}}  & \scriptsize{Rezaei and Ostadabbas (2017) \cite{P3C1-RMC-91}}  \\
\scriptsize{Social behavior}  &\scriptsize{Caltech mice dataset \cite{P6C2-Dataset-20000}}  & \scriptsize{Rezaei and Ostadabbas (2018) \cite{P3C1-RMC-92}}  \\
\cline{1-3} 
\scriptsize{\textbf{Pigs Surveillance}}  &\scriptsize{\textbf{Pigs}}  & \scriptsize{}  \\
\scriptsize{Farming}  &\scriptsize{Farming box (piglets)}  & \scriptsize{Mc Farlane and  Schofield (1995) \cite{P1C1-Median-1}}  \\
\scriptsize{Farming}  &\scriptsize{Farming box}  & \scriptsize{Guo et al. (2014) \cite{P0C0-A-76}}  \\
\scriptsize{Farming}  &\scriptsize{Farming box}  & \scriptsize{Tu et al. (2014) \cite{P0C0-A-77}}  \\
\scriptsize{Farming}  &\scriptsize{Farming box}  & \scriptsize{Tu et al. (2015) \cite{P0C0-A-77-1}}  \\
\cline{1-3} 
\scriptsize{\textbf{Hinds Surveillance}}  &\scriptsize{\textbf{Hinds}}  & \scriptsize{}  \\
\scriptsize{Animal Species Detection}  &\scriptsize{Forest Environment}  & \scriptsize{Khorrami et al. (2012) \cite{P0C0-A-75}}  \\
\cline{1-3} 
\scriptsize{\textbf{Insects Surveillance}}  &\scriptsize{\textbf{1) Honeybees}}  & \scriptsize{}  \\
\scriptsize{Hygienic Bees}  &\scriptsize{Institute of Apiculture in  Hohen-Neuendorf (Germany)}  & \scriptsize{Knauer  et al. (2005) \cite{P0C0-A-70}}  \\
\scriptsize{Honeybee Colonies}  &\scriptsize{Hive Entrance} & \scriptsize{Campbell et al. (2008)  \cite{P0C0-A-71}}	         \\  
\scriptsize{Honeybees Behaviors}      &\scriptsize{Flat Surface - Karl-Franzens-Universität in Graz (Austria)}  & \scriptsize{Kimura et al. (2012)  \cite{P0C0-A-72}} \\  
\scriptsize{Pollen Bearing Honeybees}  &\scriptsize{Hive Entrance}    & \scriptsize{Babic et al. (2016)  \cite{P0C0-A-73}}     \\
\scriptsize{Honeybees Detection}       &\scriptsize{Hive Entrance}    & \scriptsize{Pilipovic et al. (2016)  \cite{P0C0-A-74}}  \\ 
\cline{2-3}  
\scriptsize{}                          &\scriptsize{\textbf{2) Spiders}}  & \scriptsize{}  \\
\scriptsize{Spiders Detection}         &\scriptsize{Observation Box}      & \scriptsize{Iwatani  et al. (2016)  \cite{P0C0-A-760}} \\ 
\hline
\end{tabular}}
\caption{Intelligent Visual Observation of Animal/Insect Behaviors: An Overview (Part III)} \centering
\label{P0C2:Applications-3}
\end{table}

\begin{table}
\scalebox{0.9}{
\begin{tabular}{|l|l|l|} 
\hline
\scriptsize{Categories-Aims} &\scriptsize{Objects of interest-Scenes} &\scriptsize{Authors-Dates} \\
\hline
\hline
\scriptsize{\textbf{1-Forest Environments}}        &\scriptsize{\textbf{Woods}}          & \scriptsize{} \\
\scriptsize{Human Detection}  &\scriptsize{Omnidirectional cameras} & \scriptsize{Boult et al. (2003) \cite{P0C0-Sensors-SC-2}}    \\
\scriptsize{Animal Detection}  &\scriptsize{Illumination change dataset \cite{P7C1-1}} & \scriptsize{Shakeri and Zhang (2017) \cite{P7C1-1}}    \\
\scriptsize{Animal Detection}  &\scriptsize{Camera-trap dataset \cite{P7C1-2} } & \scriptsize{Yousif et al. (2017) \cite{P7C1-2}}    \\
\cline{1-3} 
\scriptsize{\textbf{2-River Environments}}        &\scriptsize{\textbf{Woods}}          & \scriptsize{} \\
\scriptsize{Floating Bottles Detection}  &\scriptsize{Dynamic Texture Videos} & \scriptsize{Zhong et al. (2003) \cite{P1C7-104}}    \\
\scriptsize{Floating wood detection}     &\scriptsize{River Videos}           & \scriptsize{Ali et al. (2012) \cite{P0C0-A-90}}     \\
\scriptsize{Floating wood detection}     &\scriptsize{River Videos}           & \scriptsize{Ali et al. (2013) \cite{P0C0-A-91}}     \\
\cline{1-3} 
\scriptsize{\textbf{3-Ocean Environments}}        &\scriptsize{}          & \scriptsize{} \\
\scriptsize{Mine Detection}  &\scriptsize{Open Sea Environments} & \scriptsize{Borghgraef et al. (2010) \cite{P0C0-A-29}}    \\
\scriptsize{Intruders Detection}  &\scriptsize{Open Sea  Environments} & \scriptsize{Szpak and Tapamo (2011) \cite{P0C0-A-29-4}}    \\
\scriptsize{Boats Detection}  &\scriptsize{Singapore Marine dataset} & \scriptsize{Prasad et al. (2016) \cite{P0C0-A-29-3}}    \\
\scriptsize{Boats Detection}  &\scriptsize{Singapore Marine dataset} & \scriptsize{Prasad et al. (2017) \cite{P0C0-A-29-2}}    \\
\cline{1-3} 
\scriptsize{\textbf{4-Submarine Environments}}                &\scriptsize{}          & \scriptsize{} \\
\scriptsize{\textbf{4.1- Swimming Pools Surveillance}}        &\scriptsize{}          & \scriptsize{} \\
\scriptsize{Human Detection}  &\scriptsize{Public Swimming Pool} & \scriptsize{Eng et al. (2003) \cite{P0C0-A-40}}    \\
\scriptsize{Human Detection}  &\scriptsize{Public Swimming Pool} & \scriptsize{Eng et al. (2004) \cite{P0C0-A-40-1}}   \\
\scriptsize{Human Detection}  &\scriptsize{Public Swimming Pool} & \scriptsize{Lei and Zhao (2010) \cite{P1C7-118}}    \\
\scriptsize{Human Detection}  &\scriptsize{Public Swimming Pool} & \scriptsize{Fei et al. (2009) \cite{P0C0-A-41}}     \\
\scriptsize{Human Detection}  &\scriptsize{Public Swimming Pool} & \scriptsize{Chan (2011) \cite{P0C0-A-42-1}}         \\
\scriptsize{Human Detection}  &\scriptsize{Public Swimming Pool} & \scriptsize{Chan (2013) \cite{P0C0-A-42}}           \\
\scriptsize{Human Detection}  &\scriptsize{Private Swimming Pool} & \scriptsize{Peixoto al. (2012) \cite{P0C0-A-44}}    \\
\scriptsize{\textbf{4.2- Tank Environments}}        &\scriptsize{}          & \scriptsize{} \\
\scriptsize{Fish Detection}  & \scriptsize{Aquarium of La Rochelle}  & \scriptsize{Penciuc et al. (2006) \cite{P0C0-A-202}}   \\
\scriptsize{Fish Detection}  & \scriptsize{Aquarium of La Rochelle}  & \scriptsize{Baf et al. (2007) \cite{P0C0-A-201}}       \\
\scriptsize{Fish Detection}  & \scriptsize{Aquarium of La Rochelle}  & \scriptsize{Baf et al. (2007) \cite{P0C0-A-200}}       \\
\scriptsize{Fish Detection}  & \scriptsize{Fish Farming}       & \scriptsize{Abe et al. (2016) \cite{P0C0-A-69-3}}       \\
\scriptsize{\textbf{4.3- Open Sea Environments}}                 &\scriptsize{}          & \scriptsize{} \\
\scriptsize{Fish Detection}  & \scriptsize{Taiwan’s coral reefs}  & \scriptsize{Kasavidis and Palazzo (2012) \cite{P0C0-A-68-1}}  \\
\scriptsize{Fish Detection}  & \scriptsize{Taiwan’s coral reefs}  & \scriptsize{Spampinato et al. (2014) \cite{P0C0-A-63}} \\
\scriptsize{Fish Detection}  & \scriptsize{UnderwaterChangeDetection Dataset}   & \scriptsize{Radolko et al. (2017) \cite{P6C2-Dataset-2025}}       \\
\hline
\end{tabular}}
\caption{Intelligent Visual Observation of Natural Environments: An Overview (Part IV)} \centering
\label{P0C2:Applications-4}
\end{table}

\begin{table}
\scalebox{0.70}{
\begin{tabular}{|l|l|l|l|} 
\hline
\scriptsize{Categories} &\scriptsize{Sub-categories-Aims} &\scriptsize{Objects of interest} &\scriptsize{Authors-Dates} \\
\hline
\hline
\multirow{1}{*}{\scriptsize{Visual Hull Computing}}
&\scriptsize{\textbf{Image-based Modeling}}         & \scriptsize{\textbf{Object}} & \scriptsize{}  \\
& \scriptsize{Marker Free}           & \scriptsize{Indoor Scenes}   & \scriptsize{Matusik et al. (2000) \cite{P0C0-A-151}}       \\
&\scriptsize{\textbf{Optical Motion Capture}}       & \scriptsize{\textbf{People}} & \scriptsize{}   \\
& \scriptsize{Marker Free}             & \scriptsize{Indoor Scenes}   & \scriptsize{Wren et al. (1997) \cite{P1C2-SG-1}}           \\
& \scriptsize{Marker Free}             & \scriptsize{Indoor Scenes}   & \scriptsize{Horprasert et al. (1998) \cite{P0C0-A-110}}    \\
& \scriptsize{Marker Free}             & \scriptsize{Indoor Scenes}   & \scriptsize{Horprasert et al. (1999) \cite{P0C0-A-111}}    \\
& \scriptsize{Marker Free}             & \scriptsize{Indoor Scenes}   & \scriptsize{Horprasert et al. (2000) \cite{P0C0-A-112}}    \\
& \scriptsize{Marker Free}             & \scriptsize{Indoor Scenes}   & \scriptsize{Mikic et al. (2002) \cite{P0C0-A-130}}         \\
& \scriptsize{Marker Free}             & \scriptsize{Indoor Scenes}   & \scriptsize{Mikic et al. (2003) \cite{P0C0-A-131}}         \\
& \scriptsize{Marker Free}             & \scriptsize{Indoor Scenes}   & \scriptsize{Chu et al. (2003) \cite{P0C0-A-140}}           \\
& \scriptsize{Marker Free}             & \scriptsize{Indoor Scenes}   & \scriptsize{Carranza et al. (2003) \cite{P0C0-A-100}}      \\
& \scriptsize{Marker Detection}        & \scriptsize{Indoor Scenes}   & \scriptsize{Guerra-Filho (2005) \cite{P0C0-A-101}}         \\
& \scriptsize{Marker Free}             & \scriptsize{Indoor Scenes}   & \scriptsize{Kim et al. (2007) \cite{P0C0-A-141}}           \\
& \scriptsize{Marker Free}             & \scriptsize{Indoor Scenes}   & \scriptsize{Park et al. (2009) \cite{P0C0-A-120}}          \\
\hline
\multirow{1}{*}{\scriptsize{Human-Machine Interaction (HMI)}}
&\scriptsize{\textbf{Arts}}             & \scriptsize{\textbf{People}}              & \scriptsize{}                                \\
&\scriptsize{Static Camera}             & \scriptsize{Interaction real/virtual}     & \scriptsize{Levin (2006)\cite{P0C0-A-360}}   \\
\cline{2-3}
&\scriptsize{\textbf{Games}}             & \scriptsize{\textbf{People}}              & \scriptsize{}                               \\
&\scriptsize{RGB-D Camera}               & \scriptsize{Interaction real/virtual}     & \scriptsize{Microsoft Kinect}               \\
\cline{2-3}  
&\scriptsize{\textbf{Ludo-Multimedia}}  & \scriptsize{}                         & \scriptsize{}   \\
&\scriptsize{}                          & \scriptsize{\textbf{Fish}}            & \scriptsize{}  \\
&\scriptsize{Aqu@theque Project} & \scriptsize{Aquarium of La Rochelle}  & \scriptsize{Penciuc et al. (2006)\cite{P0C0-A-202}}   \\
&\scriptsize{Aqu@theque Project} & \scriptsize{Aquarium of La Rochelle}  & \scriptsize{Baf et al. (2007)\cite{P0C0-A-201}}       \\
&\scriptsize{Aqu@theque Project} & \scriptsize{Aquarium of La Rochelle}  & \scriptsize{Baf et al. (2007)\cite{P0C0-A-200}}       \\
\hline
\hline
\multirow{1}{*}{\scriptsize{Vision-based Hand Gesture Recognition}}
&\scriptsize{\textbf{Human-Computer Interface (HCI}} & \scriptsize{\textbf{Hands}}    & \scriptsize{}   \\
&\scriptsize{Augmented Screen}     & \scriptsize{Indoor Scenes}     & \scriptsize{Park and Hyun (2013) \cite{P0C0-A-370}}         \\
&\scriptsize{Hand Detection}       & \scriptsize{Indoor Scenes}     & \scriptsize{Stergiopoulou  et al. (2014) \cite{P0C0-A-371}} \\
&\scriptsize{\textbf{Behavior Analysis}}       & \scriptsize{\textbf{Hands}}    & \scriptsize{} \\
&\scriptsize{Hand Detection}      & \scriptsize{Indoor Car Scenes} &  \scriptsize{Perrett et al. (2016) \cite{P0C0-A-374}}       \\
&\scriptsize{\textbf{Sign Language Interpretation and Learning}}  & \scriptsize{\textbf{Hands}}   & \scriptsize{}  \\
&\scriptsize{Hand Gesture Segmentation}    & \scriptsize{Indoor/Outdoor Scenes}   & \scriptsize{Elsayed  et al. (2015) \cite{P0C0-A-372}}     \\
&\scriptsize{\textbf{Robotics}}  & \scriptsize{\textbf{Hands}}                    & \scriptsize{}   \\
&\scriptsize{Control robot movements}      & \scriptsize{Indoor Scenes}    & \scriptsize{Khaled et al. (2015) \cite{P0C0-A-373}} \\
\hline
\hline
\multirow{1}{*}{\scriptsize{Content based Video Coding}}
& \scriptsize{\textbf{Video Content}}              & \scriptsize{\textbf{Objects}}    & \scriptsize{}  \\
&\scriptsize{Static Camera}            & \scriptsize{MPEG-4}      & \scriptsize{Chien et al. (2012) \cite{P0C0-A-300}}       \\
&\scriptsize{Static Camera}            & \scriptsize{H.264/AVC}   & \scriptsize{Paul et al. (2010)\cite{P0C0-A-310-1}}       \\
&\scriptsize{Static Camera}            & \scriptsize{H.264/AVC}   & \scriptsize{Paul et al. (2013)\cite{P5C1-VC-1}}          \\
&\scriptsize{Static Camera}            & \scriptsize{H.264/AVC}   & \scriptsize{Paul et al. (2013)\cite{P0C0-A-310-2}}       \\
&\scriptsize{Static Camera}            & \scriptsize{H.264/AVC}   & \scriptsize{Zhang et al. (2010)\cite{P0C0-A-320-1}}      \\
&\scriptsize{Static Camera}            & \scriptsize{H.264/AVC}   & \scriptsize{Zhang et al. (2012)\cite{P0C0-A-320}}        \\
&\scriptsize{Static Camera}            & \scriptsize{H.264/AVC}   & \scriptsize{Chen et al. (2012)\cite{P0C0-A-321}}         \\
&\scriptsize{Moving Camera}            & \scriptsize{H.264/AVC}   & \scriptsize{Han et al. (2012)\cite{P0C0-A-322}}          \\
&\scriptsize{Static Camera}            & \scriptsize{H.264/AVC}   & \scriptsize{Zhang et al. (2012)\cite{P0C0-A-323}}        \\
&\scriptsize{Static Camera}            & \scriptsize{H.264/AVC}   & \scriptsize{Geng et al. (2012)\cite{P0C0-A-324}}         \\
&\scriptsize{Static Camera}            & \scriptsize{H.264/AVC}   & \scriptsize{Zhang et al. (2014)\cite{P5C1-VC-10}}        \\
&\scriptsize{Static Camera}            & \scriptsize{HEVC}     	  & \scriptsize{Zhao et al. (2014)\cite{P5C1-VC-11}}         \\
&\scriptsize{Static Camera}            & \scriptsize{HEVC}     	  & \scriptsize{Zhang et al. (2014)\cite{P0C0-A-327}}        \\
&\scriptsize{Static Camera}            & \scriptsize{HEVC}     	  & \scriptsize{Chakraborty et al. (2014)\cite{P5C1-VC-2}}   \\
&\scriptsize{Static Camera}            & \scriptsize{HEVC}     	  & \scriptsize{Chakraborty et al. (2014)\cite{P5C1-VC-3}}   \\
&\scriptsize{Static Camera}            & \scriptsize{HEVC}     	  & \scriptsize{Chakraborty et al. (2017)\cite{P0C0-A-312}}   \\
&\scriptsize{Static Camera}            & \scriptsize{H.264/AVC}   & \scriptsize{Chen et al. (2012) \cite{P3C1-OA-32}}         \\
&\scriptsize{Static Camera}            & \scriptsize{H.264/AVC}   & \scriptsize{Guo et al. (2013)  \cite{P3C1-OA-31}}         \\
&\scriptsize{Static Camera}            & \scriptsize{HEVC}        & \scriptsize{Zhao et al. (2013) \cite{P3C1-OA-30}}         \\
\hline
\end{tabular}}
\caption{Miscellaneous applications: An Overview (Part V)} \centering
\label{P0C2:Applications-5}
\end{table}

\section{Intelligent Visual Surveillance}
\label{P0C2:sec:IVS}
This is the main application of background modeling and foreground detection, and in this section we only reviewed papers that appeared after 1997 because before the techniques used to detect static or moving objects were based on two or three frames differences due to the limitations of the computer. Practically, the goal in visual surveillance is to automatically detect static or moving foreground objects as follows:
\begin{itemize}
\item \textbf{Static Foreground Objects (SFO):} Detection of abandoned objects is needed to assure the security of the concerned area. A representative approach can be found in Porikli et al. \cite{P0C0-A-5000} and a full survey on background model to detect stationary objects can be found in Cuevas et al. \cite{P0C0-Survey-100}. 
\item \textbf{Moving Foreground Objects (MFO):} Detection of moving objects is needed to compute statistics on the traffic such as in road \cite{P0C0-A-1}\cite{P0C0-A-2}\cite{P0C0-A-3}\cite{P0C0-A-4}, airport \cite{P0C0-A-10} or maritime surveillance \cite{P0C0-A-20}\cite{P0C0-A-21}.  The objects of interest are very different such as vehicles, airplanes, boats, persons and luggages. Surveillance can be more specific as in the case of the study for consumer behavior in stores \cite{P0C0-A-31}\cite{P0C0-A-32}\cite{P0C0-A-33}\cite{P0C0-A-34}{P0C0-A-35}. 
\end{itemize}

\subsection{Traffic surveillance}
Traffic surveillance videos present their own characteristics in terms of locations of the camera, environments and types of the moving objects as follows:
\begin{itemize}
\item \textbf{Location of the cameras:} There are three kinds of videos in traffic videos surveillance: \textbf{(1)} videos taken by a fixed camera as in most of the cases, \textbf{(2)} aerial videos as in Reilly \cite{P0C0-A-9-79-13}, in Teutsch et al. \cite{P0C0-A-4}, and in ElTantawy and Shehata \cite{P3C1-PCP-944}\cite{P3C1-PCP-944-1}\cite{P3C1-PCP-944-2}\cite{P3C1-PCP-944-3}\cite{P3C1-PCP-944-4}, and \textbf{(3)} very high resolution satellite videos as in Kopsiaftis and Karantzalos  \cite{P0C0-A-9-79-4}. In the first case, the camera can be highly-mounted \cite{P0C0-A-9-60} or not as in most of the cases. Furthermore, the camera can be mono-directional or omnidirectional \cite{P0C0-Sensors-SC-2}\cite{P7C1-10000}. 
\item \textbf{Quality of the cameras:} Most of the time CCTV cameras are used but the quality of the cameras can varied from low-quality to high quality (HD Cameras). Low quality cameras which generate one and two frames per second and 100k pixels/frame are used to reduce both the data and the price. Indeed, a video generated by a low quality city camera is roughly estimated to be about 1GB to 10GB of data per day. 
\item \textbf{Environments:} Traffic scenes present highways, roads and urban traffic environments with their different challenges. In highways scenes, there are often shadows and illumination changes. Road scenes often present environments with trees, and their foliage moves with the wind. In urban traffic scenes, there are often illumination changes such as highlight.
\item \textbf{Foreground Objects:} Foreground objects are all road users which have different appearance in terms of color, shape and behavior. Thus, moving foreground objects of interest are \textbf{(1)} any kind of moving vehicles such as cars, trucks, motorcycles (motorbikes), etc.., \textbf{(2)} cyclists on bicycles, and \textbf{(3)} pedestrians on a pedestrian crossing. 
\end{itemize}

\indent Practically, all these characteristics generate intrinsic specificities and challenges as developed in Song and Tai \cite{P1C1-Histogram-4} and Hao et al. \cite{P0C0-A-9-45}, and they can be classified as follows: 
\begin{enumerate}
\item \textbf{Background Values:} The intensity of background scene is generally the most frequently recorded one at its pixel position. So, the background intensity can be determined by analyzing the intensity histogram. However, sensing variation and noise from image acquisition devices may result in erroneous estimation and cause a foreground object to have the maximum intensity frequency in the histogram. 
\item \textbf{Challenges due the cameras:} 
\begin{itemize}
\item In the case of cameras placed on a tall tripod, tripod may moves due the wind \cite{P1C2-KDE-45}\cite{P1C2-KDE-46}. In the case of aerial videos, the detection has particular challenges due to high object distance, simultaneous object and camera motion, shadows, or weak contrast \cite{P0C0-A-9-79-13}\cite{P0C0-A-4}\cite{P3C1-PCP-944}\cite{P3C1-PCP-944-1}\cite{P3C1-PCP-944-2}\cite{P3C1-PCP-944-3}\cite{P3C1-PCP-944-4}. In the case of satellite videos, small size of the objects and the weak contrast are the main challenges \cite{P0C0-A-9-79-4}. 
\item In the case of low quality cameras, the video is low quality, noisy, with compression artifacts, and low frame rate as developed in Toropov et al. \cite{P0C0-A-9-79-1}. 
\end{itemize}
\item \textbf{Challenges due the environments:} 
\begin{itemize}
\item Traffic surveillance needs to work 24/7 with different weather conditions in day and night scenes. Thus, a background scene dramatically changes over time by the shadow of background objects (e.g., trees) and varying illumination. 
\item Shadows may move with the wind in the trees, which may makes the detection result too noisy. 
\end{itemize}
\item  \textbf{Challenges due the foreground objects:} 
\begin{itemize}
\item The moving objects may have similar colors to those of the road and the shadow. Then, the background may be falsely
detected as an object or vice-versa. 
\item Vehicles may stop occasionally at intersections because of traffic light or control signals. Such kind of transient stops 
increase the weight of non-background Gaussian and seriously degrade the background estimation quality of a traffic image sequence. 
\item In scenarios where vehicles are driving on busy streets, this is even more challenging due to possible merged detections.  
\item False detection are caused by vehicle headlights during nighttime as developed in Li et al. \cite{P0C0-A-9-47}. 
\end{itemize}
\item \textbf{Challenges in the implementation:} The computation time needs to be low as possible because most of the applications require real-time detection. 
\end{enumerate}
Table \ref{P0C2:IVS-1} and Table \ref{P0C2:IVS-2} show an overview of the different publications in the field of traffic surveillance with information about the background model, the background maintenance, the foreground detection, the color space and the strategies used by the authors. Authors used uni-modal model or multi-modal model following where the camera is placed. For example, if the camera mainly filmed the road, the most used models are uni-modal models like the median, the histogram and the single Gaussian while if the camera is in a dynamic environment with waving trees, the most models used are multi-modal models like MOG models.  For the color space, the authors often used the RGB color space but intensity and YCrCb are also employed to be more robust against illumination changes.  For additional strategies, it concerns most of the time shadows detection because it is the most met challenges in this kind of applications.

\subsection{Airport surveillance}
Airport visual surveillance mainly concerns the area where aircrafts are parked and maintained by specialized ground vehicles such as fueling vehicles and baggage cars as well as tracking of individuals such as workers. The need of visual surveillance is given due to the following reasons: \textbf{(1)} an airport’s apron is a security relevant area, \textbf{(2)} it helps to improve transit time, i.e. the time the aircraft is parking on the apron, and \textbf{(3)} it helps to minimize costs for the company operating the airport, as personal can be deployed more efficiently, and to minimize latencies for the passengers, as the time needed for accomplishing ground services decreases. Practically, airport surveillance videos present their own characteristics as follows: 
\begin{enumerate}
\item  \textbf{Challenges due the environments:}  Weather and light changes are very challenging problems.  
\item  \textbf{Challenges due the foreground objects:}  
\begin{itemize}
\item Ground support vehicles may change their shape in an extended degree during the tracking period, e.g. a baggage car. Strictly rigid motion and object models may not work on tracking those vehicles. 
\item  Vehicles may build blobs, either with the aircraft or with other vehicles, for a longer period of time, e.g. a fueling vehicle during refueling process. 
\end{itemize}
\end{enumerate}
Table \ref{P0C2:IVS-3} shows an overview of the different publications in the field of airport surveillance with information about the background model, the background maintenance, the foreground detection, the color space and the strategies used by the authors. We can see that only the median or the single Gaussian are employed for the background model as the tarmac is an uni-modal background. The color space used is the RGB color space for all the works.  

\subsection{Maritime surveillance}
Maritime surveillance can be achieved in visible spectrum  \cite{P0C0-A-20}\cite{P0C0-A-25} or IR spectrum  \cite{P0C0-A-21}. The idea is to count, to track and to recognize boats in fluvial canals, in river, or in open sea. For fluvial canals of Miami, Socek et al. \cite{P0C0-A-29-1} proposed a hybrid foreground detection approach which combined a Bayesian background subtraction framework with an image color segmentation technique to improve accuracy. In an other work, Bloisi et al. \cite{P0C0-A-20} used the Independent Multimodal Background Subtraction (IMBS\cite{P2C0-200}) which has been designed for dealing with highly
dynamic scenarios characterized by non-regular and high frequency noise, such as water background. IMBS is a per-pixel, non-recursive, and non-predictive model. In the case of open sea environments, Culibrk et al. \cite{P1C5-70} used the original MOG \cite{P1C2-MOG-10} implemented on General Regression Neural Network (GRNN) to detect cargos. In an other work, Zhang et al. \cite{P0C0-A-25} used the median model to detect ships to track them. To detect foreign floating objects, Borghgraef et al. \cite{P0C0-A-29} employed the improved MOG model called Zivkovic-Heijden GMM  \cite{P1C2-MOG-36-1}. To detect many kinds of vessels, Szpak and Tapamo used the single Gaussian model \cite{P0C0-A-29-4}, and can tracked in their experiments jet-skis, sailboats, rigid-hulled inflatable boats, tankers, ferries and patrol boats. In infrared video, Liu et al. \cite{P0C0-A-21} employed a modified histogram model. To detect sailboats, Sobral et al. \cite{P3C1-PCP-917} developed a double constrained RPCA based on saliency detection. In a comparison work, Tran and Le \cite{P0C0-A-26} compared the original MOG and ViBe to detect boats, and they concluded that ViBe is a suitable algorithm for detecting different kinds of boats such as cargo ships, fishing boats, cruise ships, and canoes which have very different appearance  in terms of size, shape, texture and structure. To automatically detects and tracks ships (intruders) in the case of cage aquaculture, Hu et al. \cite{P0C0-A-29-5} used an approximated median based on AM \cite{P1C1-Median-1} with a wave ripple removal. In Table \ref{P0C2:IVS-3}, we show an overview of the different publications in the field of maritime surveillance with information about the background model, the background maintenance, the foreground detection, the color space and the strategies used by the authors. We can remark that the authors prefer to use multi-modal background models (MOG, Zivkovic-Heijden GMM, etc...) in this context because water presents a dynamic aspect. For the color space, RGB is often used in diurnal conditions as well as infrared for night conditions. Most of the time, additional strategies are employed like morphological processing and saliency detection to deal with the false positive detections due to the movement of the water.

\subsection{Coastal Surveillance} 
Cullen et al. \cite{P0C0-A-27}\cite{P0C0-A-28} used the behavior subtraction model developed by Jodoin et al. \cite{P5C1-FSI-210} to detect salient events in coastal environments\protect\footnotemark[4] which can be interesting for many organizations to learn about the wildlife, land erosion, impact of humans on the environment, etc. For example, biologists interested in marine mammal protection wish to know whether humans have come too close to seals on a beach. US Fish and Wildlife Service wish to know how many people and cars have been on the beach each day, and whether they have disturbed the fragile sand dunes. Practically, Cullen et al. \cite{P0C0-A-27}\cite{P0C0-A-28} detected boats, motor vehicles and people appearing close to the shoreline. In Table \ref{P0C2:IVS-3}, the reader can find an overview of the different publications in the field of coastal surveillance with information about the background model, the background maintenance, the foreground detection, the color space and the strategies used by the authors. 

\footnotetext[4]{http://vip.bu.edu/projects/vsns/coastal-surveillance/}

\subsection{Store surveillance}
The interest is more and more coming across detection of human in stores because marketing researchers in academia and industry seek for tools to aid their decision making. Unlike other types of sensors, vision presents an ability to observe customer experience without separating it from
the environment. By tracking the path traveled by the customer along the store, important pieces of information, such as customer dwell time 
and product interaction statistics can be collected. One of the most important customer statistics is the information about the
shopper groups such as in Leykin and Tuceryan \cite{P0C0-A-30}\cite{P0C0-A-31}\cite{P0C0-A-32} and  Avinash et al. \cite{P0C0-A-33}.  Table \ref{P0C2:IVS-3} shows an overview of the different publications in the field of store surveillance with information about the background model, the background maintenance, the foreground detection, the color space and the strategies used by the authors. Either a uni-modal model and multi-modal model in RGB color space are used because the videos are generally filmed in indoor scenes.\\

\subsection{Military surveillance}
Detection of moving objects in military surveillance is often referred as target detection in literature. Most of the time, it used specific sensors like infrared cameras \cite{P4C2-F-62} and Synthetic-aperture radar (SAR) imaging \cite{P7C1-20000}. In practice, the goal is to detect persons and/or vehicles in challenging environments (forest, etc...) and challenging conditions (night scenes, etc...). In literature, several authors employed background subtraction methods for target detection either in infrared or SAR imaging as follows: 
\begin{itemize}
\item \textbf{Infrared imaging:} For example, Baf et al. \cite{P4C2-F-62}\cite{P4C2-F-62-2} used a median background subtraction algorithm with a Choquet integral approach to classify objects as background or foreground in infrared video whilst Baf et al. \cite{P4C2-F-62-1} used a Type-2Fuzzy MOG (T2-FMOG) model. 
\item \textbf{SAR imaging:} A sub-aperture image can be considered as combination of background image that contains clutter and
foreground image that contains moving target. For clutter, its scattered field is varying slowly in limit angular sector. For moving target, its image position is changing along azimuth viewing angle because of circular flight. Thus, target signature is moving in consecutive sub-aperture images. Thus, value of certain pixel would have sudden change when target signature is moving onto and -7leaving it. Based on this idea, Shen et al. \cite{P7C1-20000} employed a temporal median to obtain the background image. A log-ratio operator is then integrated into the process. The operator can be defined as the logarithm of the ratio of two images. This is equivalent to subtract two logarithm images. Finally, the moving targets are detected by applying Constant False Alarm Rate (CFAR) detector. 
\end{itemize}
Because in this field videos are very confidential, authors present results on publicly civil datasets for publication.
\footnotetext[5]{http://vcipl-okstate.org/pbvs/bench/}

\begin{landscape}
\begin{table}
\scalebox{0.70}{
\begin{tabular}{|l|l|l|l|l|l|l|} 
\hline
\scriptsize{Human Activities}        &\scriptsize{Type}            &\scriptsize{Background model}    &\scriptsize{Background Maintenance}    &\scriptsize{Foreground Detection}   &\scriptsize{Color Space}     &\scriptsize{Strategies}           \\
\hline
\hline
\multirow{1}{*}{\scriptsize{Traffic Surveillance}} 																                                       
&\scriptsize{\textbf{1) Road Surveillance}}                 &\scriptsize{}                      &\scriptsize{}     &\scriptsize{}  
&\scriptsize{}     &\scriptsize{}   \\
\cline{2-7}
&\scriptsize{\textbf{1.1) Road/Highways Traffic}}           &\scriptsize{}                      &\scriptsize{}     &\scriptsize{}    
&\scriptsize{}     &\scriptsize{}   \\
&\scriptsize{Zheng et al. (2006) \cite{P0C0-A-1}} &\scriptsize{Histogram mode} &\scriptsize{Blind Maintenance} &\scriptsize{Difference}             
&\scriptsize{RGB}     &\scriptsize{Aggregation for small range}   \\
&\scriptsize{Batista et al. (2006) \cite{P0C0-A-9-30}}      &\scriptsize{Average}       &\scriptsize{Running Average}     &\scriptsize{Minimum}       &\scriptsize{RGB}     &\scriptsize{Double Backgrounds}   \\
&\scriptsize{Monteiro et al. (2008) \cite{P0C0-A-9-31}}     &\scriptsize{Median}        &\scriptsize{Sliding Median}     &\scriptsize{Minimum}        &\scriptsize{RGB}     &\scriptsize{Double Backgrounds-Shadow/Highlight Removal}   \\
&\scriptsize{Monteiro et al. (2008) \cite{P0C0-A-9-31-1}}   &\scriptsize{Median}        &\scriptsize{Sliding Median}     
&\scriptsize{Minimum}       &\scriptsize{RGB}     &\scriptsize{Double Backgrounds-Shadow/Highlight Removal}   \\
&\scriptsize{Monteiro et al. (2008) \cite{P0C0-A-32}} &\scriptsize{Codebook \cite{P1C6-100}} &\scriptsize{Idem Codebook} 
&\scriptsize{Idem Codebook}  &\scriptsize{RGB}     &\scriptsize{Shadow/Highlight Detection \cite{P5C1-CF-300}}   \\
&\scriptsize{Monteiro (2009) \cite{P0C0-A-9-33}}            &\scriptsize{Sliding Median}            &\scriptsize{}       
&\scriptsize{Minimum}          &\scriptsize{RGB}     &\scriptsize{Double Backgrounds-Shadow/Highlight Removal}   \\
&\scriptsize{Hao et al. (2013) \cite{P1C2-KDE-159}}         &\scriptsize{KDE \cite{P1C2-KDE-46}}    &\scriptsize{Idem KDE} 
&\scriptsize{Idem KDE}     &\scriptsize{Joint Domain-Range Features \cite{P1C2-KDE-46}}&\scriptsize{Foreground Model}   \\
&\scriptsize{Ling et al. (2014) \cite{P0C0-A-3}}            &\scriptsize{Dual Layer Approach \cite{P0C0-A-3}}  &\scriptsize{} 
&\scriptsize{}   &\scriptsize{}     &\scriptsize{}  \\
&\scriptsize{Sawalakhe and Metkar (2014) \cite{P0C0-A-9-43}} &\scriptsize{Spatio-Temporal BS/FD (ST-BSFD)} &\scriptsize{-}  
&\scriptsize{AND} &\scriptsize{RGB}     &\scriptsize{-} \\ 
&\scriptsize{Huang and Chen (2013) \cite{P0C0-A-9-62}}      &\scriptsize{Cerebellar-Model-Articulation-Controller (CMAC) \cite{P0C0-A-9-62}} &\scriptsize{-} &\scriptsize{Threshold}  &\scriptsize{YCrCb}     &\scriptsize{Block} \\
&\scriptsize{Chen and Huang (2014) \cite{P0C0-A-9-61}}      &\scriptsize{PCA-based RBF Network  \cite{P0C0-A-9-61}} 
&\scriptsize{Selective Maintenance} &\scriptsize{Euclidean distance} &\scriptsize{YCrCb}     &\scriptsize{Block} \\
&\scriptsize{Lee et al (2015) \cite{P0C0-A-9-42}}  				  &\scriptsize{GDSM with Running Average \cite{P0C0-A-9-42}} &\scriptsize{Selective Maintenance}  &\scriptsize{AND} &\scriptsize{RGB}     &\scriptsize{Shadow Detection \cite{P0C1-90-1}}  \\
&\scriptsize{Aqel et al. (2015) \cite{P0C0-A-9-20}} 		      &\scriptsize{SG on Intensity Transition} &\scriptsize{Idem SG}     
&\scriptsize{Idem SG}  &\scriptsize{RGB}     &\scriptsize{-} \\
&\scriptsize{Aqel et al. (2016) \cite{P0C0-A-9-21}} 		      &\scriptsize{SG on Intensity Transition} &\scriptsize{Idem SG}    
&\scriptsize{Idem SG}  &\scriptsize{RGB}     &\scriptsize{Shadow Detection} \\
&\scriptsize{Wang et al. (2016) \cite{P5C1-MulF-9}} 		    &\scriptsize{Median} &\scriptsize{}     &\scriptsize{}
&\scriptsize{}     &\scriptsize{}  \\
&\scriptsize{Dey and Praveen (2016) \cite{P0C0-A-9-41}}     &\scriptsize{GDSM with background subtraction \cite{P0C0-A-9-1}} &\scriptsize{Yes}     &\scriptsize{AND}     &\scriptsize{RGB}     &\scriptsize{Post-processing} \\
&\scriptsize{Hadiuzzaman et al. (2017) \cite{P0C0-A-9-79-20}}    &\scriptsize{Median} &\scriptsize{Yes}  &\scriptsize{Idem Median}   
&\scriptsize{Intensity}     &\scriptsize{Shadow Detection} \\
\cline{2-7}
&\scriptsize{\textbf{1.2)} Urban Traffic}                   &\scriptsize{}   &\scriptsize{}     &\scriptsize{}       
&\scriptsize{}     &\scriptsize{} \\
&\scriptsize{\textbf{Conventional camera}}                           &\scriptsize{}   &\scriptsize{}     &\scriptsize{}       
&\scriptsize{}     &\scriptsize{} \\
&\scriptsize{Hwang et al. (2009) \cite{P0C0-A-9-48}}        &\scriptsize{MOG-ALR \cite{P0C0-A-9-48}} &\scriptsize{Adaptive Learning Rate}     &\scriptsize{Idem MOG}    &\scriptsize{RGB}     &\scriptsize{-} \\
&\scriptsize{Intachak and Kaewapichai (2011) \cite{P0C0-A-9-79-2}}  &\scriptsize{Mean (Clean images)}  &\scriptsize{Selective Maintenance}     &\scriptsize{Idem Mean} &\scriptsize{RGB}  &\scriptsize{Illumination Adjustment} \\
&\scriptsize{Milla et al. (2013) \cite{P0C0-A-9-79-3}}      &\scriptsize{$\Sigma-\Delta$ filter} &\scriptsize{$\Sigma-\Delta$ filter}
&\scriptsize{$\Sigma-\Delta$ filter}    &\scriptsize{Intensity}     &\scriptsize{Short-term/Long-term backgrounds} \\
&\scriptsize{Toropov et al. (2015) \cite{P0C0-A-9-79-1}} &\scriptsize{MOG \cite{P1C2-MOG-10}} &\scriptsize{Idem MOG}  &\scriptsize{Idem MOG}    &\scriptsize{Color} &\scriptsize{Brightness Adjustment} \\
&\scriptsize{Zhang et al. (2016) \cite{P1C2-MOG-710}} 		  &\scriptsize{GMMCM \cite{P1C2-MOG-710}}  &\scriptsize{Idem MOG \cite{P1C2-MOG-10}}     &\scriptsize{Confidence Period} &\scriptsize{Intensity}     &\scriptsize{Classification of traffic density} \\
&\scriptsize{\textbf{Highly-mounted camera}}      &\scriptsize{}   &\scriptsize{}     &\scriptsize{}  &\scriptsize{}     
&\scriptsize{} \\
&\scriptsize{Gao et al. (2014) \cite{P0C0-A-9-60}}          &\scriptsize{SG \cite{P1C2-SG-1}} &\scriptsize{Idem SG}     &\scriptsize{Idem SG}				
&\scriptsize{YCrCb}     &\scriptsize{Shadow detection \cite{P0C1-90}} \\
&\scriptsize{Quesada and Rodriguez (2016) \cite{P3C1-PCP-1066}}          &\scriptsize{incPCP \cite{P3C1-PCP-1063}} &\scriptsize{Idem incPCP}     &\scriptsize{Idem incPCP}				&\scriptsize{Intensity}     &\scriptsize{-} \\
&\scriptsize{\textbf{Headlight Removal}}                         &\scriptsize{}   &\scriptsize{}     &\scriptsize{}                
&\scriptsize{}     &\scriptsize{} \\
&\scriptsize{Li et al. (2016) \cite{P0C0-A-9-47}}           &\scriptsize{GMM \cite{P1C2-MOG-10}}  &\scriptsize{Idem GMM}  &\scriptsize{Idem GMM}   &\scriptsize{RGB}     &\scriptsize{Headlight/Shadow Removal} \\
&\scriptsize{\textbf{Intersection}}                         &\scriptsize{}   &\scriptsize{}     &\scriptsize{} &\scriptsize{}     &\scriptsize{} \\
&\scriptsize{Ding et al. (2012) \cite{P0C0-A-9-44}} &\scriptsize{CPS based GMM \cite{P0C0-A-9-44}} &\scriptsize{Idem GMM} &\scriptsize{Idem GMM}   
&\scriptsize{RGB}     &\scriptsize{Cyber Physical System} \\
&\scriptsize{Ding et al. (2012) \cite{P0C0-A-9-44}} &\scriptsize{CPS based FGD \cite{P6C2-Dataset-10}} &\scriptsize{Idem FGD} &\scriptsize{Idem FGD} &\scriptsize{RGB}     &\scriptsize{Cyber Physical System} \\
&\scriptsize{Alldieck (2015) \cite{P0C0-A-9-79-12}}   &\scriptsize{Zivkovic-Heijden GMM  \cite{P1C2-MOG-36-1}} &\scriptsize{Idem GMM} &\scriptsize{Idem GMM}  &\scriptsize{RGB/IR}          &\scriptsize{Multimodal cameras} \\
&\scriptsize{Mendoca et Oliveira (2015) \cite{P0C0-A-9-79-6}}   &\scriptsize{Context supported
ROad iNformation (CRON) \cite{P0C0-A-9-79-6}} &\scriptsize{Selective Maintenance}     &\scriptsize{Idem AM}  &\scriptsize{RGB}          &\scriptsize{} \\
&\scriptsize{Li et al. (2016) \cite{P0C0-A-9-44-1}}   &\scriptsize{incPCP \cite{P3C1-PCP-1062}} &\scriptsize{Idem incPCP}     &\scriptsize{Idem incPCP}     &\scriptsize{RGB}     &\scriptsize{-} \\
&\scriptsize{\textbf{Obstacle Detection}}          &\scriptsize{}   &\scriptsize{}     &\scriptsize{}                    
&\scriptsize{}     &\scriptsize{} \\
&\scriptsize{Lan et al. (2015) \cite{P1C2-MOG-761}}  				&\scriptsize{SUOG \cite{P1C2-MOG-761}} &\scriptsize{Selective Maintenance}&\scriptsize{Idem GMM}  &\scriptsize{RGB}  &\scriptsize{Obstacle Detection Model} \\
\hline
\end{tabular}}
\caption{Intelligent Visual Surveillance of Human Activities: An Overview (Part 1). "-" indicated that the corresponding step used in not indicated in the paper.} \centering
\label{P0C2:IVS-1}
\end{table}
\end{landscape}

\begin{landscape}
\begin{table}
\scalebox{0.70}{
\begin{tabular}{|l|l|l|l|l|l|l|} 
\hline
\scriptsize{Human Activities}        &\scriptsize{Type}            &\scriptsize{Background model}    &\scriptsize{Background Maintenance}    &\scriptsize{Foreground Detection}   &\scriptsize{Color Space}     &\scriptsize{Strategies}           \\
\hline
\hline
&\scriptsize{\textbf{2)Vehicle Counting}}                   &\scriptsize{}   &\scriptsize{}     &\scriptsize{}               
&\scriptsize{}     &\scriptsize{} \\
&\scriptsize{Unzueta et al. (2012) \cite{P0C0-A-2}}         &\scriptsize{Multicue approach \cite{P0C0-A-2}} &\scriptsize{} &\scriptsize{}  
&\scriptsize{}     &\scriptsize{} \\
&\scriptsize{Virginas-Tar et al. (2014) \cite{P0C0-A-9-50}} &\scriptsize{MOG-EM \cite{P1C2-MOG-90}}  &\scriptsize{Idem MOG-EM} &\scriptsize{Idem MOG-EM}	 &\scriptsize{}     &\scriptsize{Shadow Detection \cite{P5C1-CF-300}} \\
\cline{2-7}
&\scriptsize{\textbf{3) Vehicle Detection}}                 &\scriptsize{}  &\scriptsize{}     &\scriptsize{}                      
&\scriptsize{}     &\scriptsize{} \\
&\scriptsize{\textbf{3.1) Conventional Video}}              &\scriptsize{} &\scriptsize{}      &\scriptsize{}                   
&\scriptsize{}     &\scriptsize{} \\
&\scriptsize{Wang and Song (2011) \cite{P0C0-A-9-46}} &\scriptsize{GMM with Spatial Correlation Method  \cite{P0C0-A-9-46}} &\scriptsize{Idem GMM}    &\scriptsize{Idem GMM}  &\scriptsize{HSV}     &\scriptsize{Spatial Correlation Method} \\
&\scriptsize{Hadi et al. (2014) \cite{P0C0-A-9}} 		        &\scriptsize{Histogram mode} &\scriptsize{Blind Maintenance} &\scriptsize{Absolute Difference} &\scriptsize{RGB}     &\scriptsize{Morphological Processing} \\
&\scriptsize{Hadi et al. (2017) \cite{P0C0-A-9-1}} 		      &\scriptsize{Histogram mode} &\scriptsize{Blind Maintenance} &\scriptsize{Absolute Difference} &\scriptsize{RGB}     &\scriptsize{Morphological Processing} \\
&\scriptsize{\textbf{3.2) Aerial Video}}                    &\scriptsize{}               &\scriptsize{}     &\scriptsize{}           
&\scriptsize{}     &\scriptsize{} \\
& \scriptsize{Lin et al. (2009) \cite{P0C0-A-14}}           &\scriptsize{Two CFD}        &\scriptsize{-}     &\scriptsize{-} 
&\scriptsize{RGB}   &\scriptsize{Translation}   \\
&\scriptsize{Reilly (2012) \cite{P0C0-A-9-79-13}}         &\scriptsize{Median} &\scriptsize{Idem Median}     &\scriptsize{Idem Median}&\scriptsize{Intensity}   &\scriptsize{-}   \\
&\scriptsize{Teutsch et al. (2014) \cite{P0C0-A-4}}         &\scriptsize{Independent Motion Detection \cite{P0C0-A-4}} &\scriptsize{}
&\scriptsize{} &\scriptsize{}   &\scriptsize{}   \\
&\scriptsize{\textbf{3.3) Satellite Video}}                    &\scriptsize{}       &\scriptsize{}     &\scriptsize{}               
&\scriptsize{}     &\scriptsize{} \\
&\scriptsize{Kopsiaftis and Karantzalos (2015) \cite{P0C0-A-9-79-4}} &\scriptsize{Mean}  &\scriptsize{Idem Mean}     &\scriptsize{Idem Mean}         &\scriptsize{Intensity}     &\scriptsize{-} \\
&\scriptsize{Yang et al. (2016) \cite{P0C0-A-9-79-41}}  &\scriptsize{Local Saliency based Background Model based on ViBe (LS-ViBe) \cite{P0C0-A-9-79-41}}   &\scriptsize{Idem ViBe}     &\scriptsize{Idem ViBe} &\scriptsize{Intensity}     &\scriptsize{-} \\
\cline{2-7}
&\scriptsize{\textbf{4) Illegally Parked Vehicles}}         &\scriptsize{}         &\scriptsize{}     &\scriptsize{}                
&\scriptsize{}     &\scriptsize{} \\
&\scriptsize{Lee et al. (2007) \cite{P0C0-A-9-70}}          &\scriptsize{Median}   &\scriptsize{Selective Maintenance} &\scriptsize{Difference}   &\scriptsize{RGB}     &\scriptsize{Morphological Processing} \\
&\scriptsize{Zhao et al. (2013)\cite{P0C0-A-9-71}}          &\scriptsize{Average}  &\scriptsize{Running Average} &\scriptsize{Difference}             &\scriptsize{HIS}     &\scriptsize{Morphological Processing-Salient Object Detection} \\
&\scriptsize{Saker et al. (2015)\cite{P0C0-A-9-72}}          &\scriptsize{GMM-PUC \cite{P1C2-MOG-762}} &\scriptsize{Idem GMM} &\scriptsize{Idem GMM}  &\scriptsize{RGB}  &\scriptsize{Detection of Stationary Object} \\
&\scriptsize{Chunyang et al. (2015) \cite{P0C0-A-9-73}}      &\scriptsize{MOG \cite{P1C2-MOG-10}} &\scriptsize{Idem MOG}     &\scriptsize{Idem MOG}   &\scriptsize{RGB}  &\scriptsize{Morphological Processing} \\
&\scriptsize{Wahyono et al. (2015) \cite{P0C0-A-9-74}}    &\scriptsize{Running Average} &\scriptsize{Selective Maintenance}  &\scriptsize{Difference} &\scriptsize{RGB}  &\scriptsize{Dual background models} \\
\cline{2-7}
&\scriptsize{\textbf{5) Vacant parking area}}               &\scriptsize{}         &\scriptsize{}     &\scriptsize{}                 
&\scriptsize{}     &\scriptsize{} \\
&\scriptsize{Postigo et al. (2015) \cite{P0C0-A-9-75}}   &\scriptsize{MOG-EM \cite{P1C2-MOG-90}} &\scriptsize{Idem MOG-EM} &\scriptsize{Idem MOG-EM}  &\scriptsize{RGB}     &\scriptsize{Transience Map} \\
&\scriptsize{Neuhausen (2015) \cite{P0C0-A-9-76}}         &\scriptsize{SG}  &\scriptsize{Selective Maintenance}     &\scriptsize{Choquet Integral \cite{P4C2-F-70-1}} &\scriptsize{YCrCb-ULBP \cite{P5C1-TF-26}}  &\scriptsize{Adaptive weight on Illumination Normalization} \\
\cline{2-7}
&\scriptsize{\textbf{6) Motorcycle (Motorbike) Detection}}   &\scriptsize{}               &\scriptsize{}     &\scriptsize{}          
&\scriptsize{}     &\scriptsize{} \\
&\scriptsize{Silva et al. (2013) \cite{P0C0-A-9-49-2}}       &\scriptsize{Zivkovic-Heijden GMM  \cite{P1C2-MOG-36-1}}  &\scriptsize{Idem GMM}     &\scriptsize{Idem GMM}                                       &\scriptsize{Intensity}     &\scriptsize{-}     \\
&\scriptsize{Waranusast et al. (2013) \cite{P0C0-A-9-49-1}}       &\scriptsize{Zivkovic-Heijden GMM  \cite{P1C2-MOG-36-1}}  &\scriptsize{Idem GMM}     &\scriptsize{Idem GMM}                                      &\scriptsize{RGB}     &\scriptsize{Morphological Operators}     \\
&\scriptsize{Dahiya et al. (2016) \cite{P0C0-A-9-49}}       &\scriptsize{Zivkovic-Heijden GMM  \cite{P1C2-MOG-36-1}}  &\scriptsize{Idem GMM}     &\scriptsize{Idem GMM}                                      &\scriptsize{Intensity}     &\scriptsize{Detection Bike-riders}     \\
\hline
\end{tabular}}
\caption{Intelligent Visual Surveillance of Human Activities: An Overview (Part 2). "-" indicated that the corresponding step used in not indicated in the paper.} \centering
\label{P0C2:IVS-2}
\end{table}
\end{landscape}

\begin{landscape}
\begin{table}
\scalebox{0.70}{
\begin{tabular}{|l|l|l|l|l|l|l|} 
\hline
\scriptsize{Human Activities}        &\scriptsize{Type}            &\scriptsize{Background model}    &\scriptsize{Background Maintenance}    &\scriptsize{Foreground Detection}   &\scriptsize{Color Space}     &\scriptsize{Strategies}           \\
\hline
\hline
\multirow{1}{*}{\scriptsize{Airport Surveillance}} 															                  
&\scriptsize{Blauensteiner and Kampel (2004) \cite{P0C0-A-10}}  &\scriptsize{Median}          &\scriptsize{Idem Median} &\scriptsize{Idem Median}  &\scriptsize{RGB}     &\scriptsize{-} \\
&\scriptsize{Aguilera et al. (2005) \cite{P0C0-A-11}}           &\scriptsize{Single Gaussian} &\scriptsize{Idem SG}     &\scriptsize{Idem SG}  &\scriptsize{RGB}     &\scriptsize{-} \\
&\scriptsize{Thirde et al. (2006) \cite{P0C0-A-12}}             &\scriptsize{Single Gaussian} &\scriptsize{Idem SG}     &\scriptsize{Idem SG}  &\scriptsize{RGB}     &\scriptsize{-} \\
&\scriptsize{Aguilera et al. (2006) \cite{P0C0-A-13}}           &\scriptsize{Single Gaussian} &\scriptsize{Idem SG}     &\scriptsize{Idem SG}  &\scriptsize{RGB}     &\scriptsize{-} \\
\hline
\multirow{1}{*}{\scriptsize{Maritime Surveillance}}
&\scriptsize{\textbf{1) Fluvial canals environment}}          &\scriptsize{}                  &\scriptsize{}     &\scriptsize{}  &\scriptsize{}     &\scriptsize{} \\
&\scriptsize{Socek et al. (2005) \cite{P0C0-A-29-1}}          &\scriptsize{Two CFD}           &\scriptsize{-}  &\scriptsize{Bayesian Decision \cite{P0C0-A-31}}  &\scriptsize{RGB}     &\scriptsize{Color Segmentation} \\
&\scriptsize{Bloisi et al. (2014) \cite{P0C0-A-20}}           &\scriptsize{IMBS \cite{P2C0-200}}                 &\scriptsize{}  &\scriptsize{}  
&\scriptsize{}     &\scriptsize{} \\
\cline{2-7}
&\scriptsize{\textbf{2) River environment}}             &\scriptsize{}                   &\scriptsize{}    &\scriptsize{}  &\scriptsize{} &\scriptsize{}        \\
&\scriptsize{Zheng et al. (2013) \cite{P0C0-A-29-73}}   &\scriptsize{LBP Histogram \cite{P5C1-TF-11}}           &\scriptsize{LBP Histogram \cite{P5C1-TF-11}}  &\scriptsize{LBP Histogram \cite{P5C1-TF-11}}  &\scriptsize{RGB}     &\scriptsize{-} \\
&\scriptsize{Mei et al. (2017) \cite{P3C4-DGS-16}}   &\scriptsize{EAdaDGS \cite{P3C4-DGS-16}}    &\scriptsize{Improved mechanism}  &\scriptsize{Idem AdaDGS \cite{P3C4-DGS-10}}          &\scriptsize{RGB}     &\scriptsize{Multi-resolution} \\
\cline{2-7}
&\scriptsize{\textbf{3) Open sea environment}}                &\scriptsize{}                   &\scriptsize{}    &\scriptsize{}  &\scriptsize{} &\scriptsize{}        \\
&\scriptsize{Culibrk et al. (2006) \cite{P1C5-70}}            &\scriptsize{MOG-GNN \cite{P1C2-MOG-10}}           &\scriptsize{Idem MOG}  &\scriptsize{Idem MOG}  &\scriptsize{Intensity}     &\scriptsize{-} \\
&\scriptsize{Zhang et al. (2009) \cite{P0C0-A-25}}            &\scriptsize{Median}                               &\scriptsize{}  &\scriptsize{} 
&\scriptsize{}     &\scriptsize{} \\
&\scriptsize{Borghgraef et al. (2010) \cite{P0C0-A-29}}       &\scriptsize{Zivkovic-Heijden GMM  \cite{P1C2-MOG-36-1}}  &\scriptsize{Idem Zivkovic-Heijden GMM}  &\scriptsize{Idem Zivkovic-Heijden GMM} &\scriptsize{IR}   &\scriptsize{-} \\
&\scriptsize{Arshad et al. (2010) \cite{P0C0-A-29-6}}      &\scriptsize{-}  &\scriptsize{-}     &\scriptsize{-} &\scriptsize{RGB}   &\scriptsize{Morphological Processing} \\
&\scriptsize{Arshad et al. (2011) \cite{P0C0-A-29-61}}     &\scriptsize{-}  &\scriptsize{-}     &\scriptsize{-} &\scriptsize{RGB}   &\scriptsize{Morphological Processing} \\
&\scriptsize{Szpak and Tapamo (2011) \cite{P0C0-A-29-4}}      &\scriptsize{Single Gaussian \cite{P1C2-SG-1}}  &\scriptsize{Idem SG}  
&\scriptsize{Idem SG} &\scriptsize{Intensity}   &\scriptsize{Spatial Smoothness} \\
& \scriptsize{Hu et al. (2011) \cite{P0C0-A-29-5}}            &\scriptsize{Modified AM \cite{P0C0-A-29-5}}     &\scriptsize{Idem Modified AM}    &\scriptsize{Idem Modified AM} &\scriptsize{RGB}              &\scriptsize{Fast 4-Connected Component Labeling} \\
&\scriptsize{Saghafi et al. (2012) \cite{P0C0-A-29-7}}            &\scriptsize{Modified ViBe \cite{P0C0-A-29-7}}     &\scriptsize{Modified ViBe}    &\scriptsize{Modified ViBe} &\scriptsize{RGB}              &\scriptsize{Backwash Cancellation Algorithm} \\
&\scriptsize{Xie et al. (2012) \cite{P0C0-A-29-71}}            &\scriptsize{Three CFD}     &\scriptsize{Selective Maintenance}  &\scriptsize{AND} &\scriptsize{Intensity} &\scriptsize{} \\
&\scriptsize{Zhang et al. (2012) \cite{P0C0-A-29-72}}      &\scriptsize{PCA \cite{P1C4-1}/GMM \cite{P1C2-MOG-10}}     &\scriptsize{Idem PCA \cite{P1C4-1}/GMM \cite{P1C2-MOG-10}}    &\scriptsize{Idem PCA \cite{P1C4-1}/GMM \cite{P1C2-MOG-10}} &\scriptsize{Intensity} &\scriptsize{-} \\
&\scriptsize{Arshad et al. (2014) \cite{P0C0-A-29-62}}     &\scriptsize{-}  &\scriptsize{-}     &\scriptsize{-} &\scriptsize{RGB}   &\scriptsize{Morphological Processing} \\
&\scriptsize{Liu et al. (2014) \cite{P0C0-A-21}}              &\scriptsize{Modified Histogram \cite{P0C0-A-21}}              &\scriptsize{Idem Histogram} &\scriptsize{Idem Histogram}  &\scriptsize{IR}     &\scriptsize{Adaptive Row Mean Filter} \\
&\scriptsize{Sobral et al. (2015) \cite{P3C1-PCP-917}}        &\scriptsize{Double Constrained RPCA \cite{P3C1-PCP-917}}      &\scriptsize{Idem RPCA}  &\scriptsize{Idem RPCA} &\scriptsize{RGB}   &\scriptsize{Saliency Detection} \\
&\scriptsize{Tran and Le (2016) \cite{P0C0-A-26}}             &\scriptsize{ViBe \cite{P2C1-200}}  &\scriptsize{Idem ViBe}    &\scriptsize{Idem ViBe} &\scriptsize{RGB}   &\scriptsize{Saliency Detection} \\
\hline
\multirow{1}{*}{\scriptsize{Store Surveillance}}
&\scriptsize{Leykin and Tuceryan (2007) \cite{P0C0-A-30}}  &\scriptsize{Codebook \cite{P1C6-100}} &\scriptsize{Idem Codebook}     &\scriptsize{Idem Codebook}  &\scriptsize{RGB}   &\scriptsize{-} \\
&\scriptsize{Leykin and Tuceryan (2005) \cite{P0C0-A-31}}  &\scriptsize{Codebook \cite{P1C6-100}} &\scriptsize{Idem Codebook}     &\scriptsize{Idem Codebook}  &\scriptsize{RGB}   &\scriptsize{-} \\
&\scriptsize{Leykin and Tuceryan (2005) \cite{P0C0-A-32}}  &\scriptsize{Codebook \cite{P1C6-100}} &\scriptsize{Idem Codebook}     &\scriptsize{Idem Codebook}  &\scriptsize{RGB}   &\scriptsize{-} \\
&\scriptsize{Avinash et al. (2012) \cite{P0C0-A-33}}       &\scriptsize{Single Gaussian \cite{P1C2-SG-1}}            &\scriptsize{Idem SG}  &\scriptsize{Idem SG} &\scriptsize{RGB}     &\scriptsize{-} \\
&\scriptsize{Zhou et al. (2017) \cite{P0C0-A-35}}     &\scriptsize{MOG \cite{P1C2-MOG-1}}     &\scriptsize{Idem MOG}  &\scriptsize{Idem MOG} &\scriptsize{RGB}     &\scriptsize{-} \\
\hline
\multirow{1}{*}{\scriptsize{Coastal Surveillance}}
&\scriptsize{Cullen et al. (2012) \cite{P0C0-A-27}}           &\scriptsize{Behavior Subtraction \cite{P5C1-FSI-210}}  &\scriptsize{Idem BS}     &\scriptsize{Idem BS} &\scriptsize{RGB}   &\scriptsize{-} \\
&\scriptsize{Cullen (2012) \cite{P0C0-A-28}}                  &\scriptsize{Behavior Subtraction \cite{P5C1-FSI-210}}  &\scriptsize{Idem BS}     &\scriptsize{Idem BS} &\scriptsize{RGB}   &\scriptsize{-} \\
\hline
\multirow{1}{*}{\scriptsize{Swimming Pools Surveillance}}
&\scriptsize{\textbf{1) Online videos}}                    &\scriptsize{}    &\scriptsize{}     &\scriptsize{}       &\scriptsize{} &\scriptsize{} \\
&\scriptsize{1.1) Top view videos}                         &\scriptsize{}    &\scriptsize{}     &\scriptsize{}       &\scriptsize{} &\scriptsize{} \\
&\scriptsize{Eng et al. (2003) \cite{P0C0-A-40}}           &\scriptsize{Block-based median \cite{P0C0-A-40}}         &\scriptsize{Sliding Mean} &\scriptsize{Block-based Difference}                       &\scriptsize{CIELa*b*} &\scriptsize{Partial Occlusion Handling} \\       
&\scriptsize{Eng et al. (2004) \cite{P0C0-A-40-1}}         &\scriptsize{Region-based single multivariate Gaussian \cite{P0C0-A-40-1}} &\scriptsize{Idem SG}  &\scriptsize{Region based SG}       &\scriptsize{CIELa*b*}     &\scriptsize{Handling Specular Reflection} \\
&\scriptsize{Chan (2011) \cite{P0C0-A-42-1}}               &\scriptsize{SG/optical flow \cite{P0C0-A-42-1}}      &\scriptsize{Idem SG} &\scriptsize{Designed Distance}                            &\scriptsize{RGB}     &\scriptsize{-} \\
&\scriptsize{Chan (2014) \cite{P0C0-A-42}}                 &\scriptsize{MOG/optical flow \cite{P0C0-A-42}}           &\scriptsize{Idem MOG} &\scriptsize{Idem MOG}                                     &\scriptsize{RGB}     &\scriptsize{-} \\
&\scriptsize{1.2) Underwater videos}                   &\scriptsize{}    &\scriptsize{} &\scriptsize{}       &\scriptsize{} &\scriptsize{} \\
&\scriptsize{Fei et al. (2009) \cite{P0C0-A-41}}       &\scriptsize{MOG} &\scriptsize{Idem MOG}              &\scriptsize{} &\scriptsize{} &\scriptsize{Shadow Removal} \\
&\scriptsize{Lei and Zhao (2010) \cite{P1C7-118}}      &\scriptsize{Kalman filter \cite{P1C7-103}}               &\scriptsize{Idem Kalman filter} &\scriptsize{Idem Kalman filter} &\scriptsize{RGB}     &\scriptsize{} \\
&\scriptsize{Zhang et al. (2015) \cite{P0C0-A-45}}     &\scriptsize{-}  &\scriptsize{-} &\scriptsize{-}	 &\scriptsize{RGB}     &\scriptsize{Inter-frame based denoising} \\
\cline{2-7}
&\scriptsize{\textbf{2) Archived videos}}                  &\scriptsize{}  &\scriptsize{}     &\scriptsize{}   &\scriptsize{}    &\scriptsize{} \\
&\scriptsize{Sha et al. (2014) \cite{P0C0-A-43}}           &\scriptsize{-} &\scriptsize{-}    &\scriptsize{-}	 &\scriptsize{RGB} &\scriptsize{-} \\
\cline{2-7}
&\scriptsize{\textbf{3) Private swimming pools}}            &\scriptsize{}                              &\scriptsize{} &\scriptsize{} &\scriptsize{} &\scriptsize{} \\
&\scriptsize{Peixoto et al. (2012) \cite{P0C0-A-44}}        &\scriptsize{Mean/Two CFD \cite{P0C0-A-44}} &\scriptsize{Selective Maintenance} &\scriptsize{Mean Distance/Two CFD Distance} &\scriptsize{HSV}  &\scriptsize{} \\
\hline
\end{tabular}}
\caption{Intelligent Visual Surveillance of Human Activities: An Overview (Part 3). "-" indicated that the background model used in not indicated in the paper.} \centering
\label{P0C2:IVS-3}
\end{table}
\end{landscape}

\section{Intelligent visual observation of animals and insects} 
\label{P0C2:sec:AIB}
Surveillance with fixed cameras can also concern census and activities of animals in open protected areas (river, ocean, forest, etc...), as well as ethology in closed areas (zoo, cages, etc...). In these application, the detection done by background subtraction is followed by a tracking phase or a recognition phase \cite{P7C1-200}\cite{P7C1-210}\cite{P7C1-211}. In practice, the objects of interest are then animals such as birds \cite{P0C0-A-50}\cite{P0C0-A-51}\cite{P0C0-A-55}\cite{P0C0-A-56}\cite{P0C0-A-57}, fish \cite{P0C0-A-60}\cite{P0C0-A-61}, honeybees \cite{P0C0-A-70}\cite{P0C0-A-71}\cite{P0C0-A-72}\cite{P0C0-A-73}, hinds \cite{P0C0-A-75}\cite{P3C1-PCP-903}, squirrels \cite{P7C1-2000}\cite{P7C1-2001}, mice \cite{P3C1-RMC-91}\cite{P3C1-RMC-92} or pigs \cite{P0C0-A-77}. In practice, animals live and evolve in different environments that can be classified in three main categories: 1) Natural environments like forest, river and ocean, 2) study's environment like tank for fish and cages for mice, 3) farm environments for surveillance of pigs \cite{P1C1-Median-1} and  livestock \cite{P7C1-300}. Thus, videos for visual observation of animals and insects present their own characteristics due the intrinsic appearance and behavior of the detected animals or insects, and the environments in which they are filmed.  In these application, the detection done by background subtraction is followed by a tracking phase or a recognition phase \cite{P7C1-200}\cite{P7C1-210}\cite{P7C1-211}. Therefore, there are generally no a priori on the shape and the color of the objects when background subtraction is applied. In addition of these different intrinsic appearances in terms of size, shape, color and texture, and behavior in terms of velocity, videos which contains animals or insects present challenging characteristics due the background in natural environments which are developed in the Section \ref{P0C2:sec:IVO}. Pratically, events such as VAIB (Visual observation and analysis of animal and insect behavior) in conjunction with ICPR addressed the problem of the detection in animals and insects.

\subsection{Birds Surveillance} 
Detection of birds is a crucial problem for multiple applications such as aviation safety, avian  protection, and ecological science of migrant bird species. There are three kinds of bird observations: \textbf{(1)} observations at human made feeder stations \cite{P0C0-A-50}\cite{P0C0-A-51},  \textbf{(2)} observation at natural nesting stations \cite{P0C0-A-79}\cite{P0C0-A-56}\cite{P0C0-A-57}, and \textbf{(3)} observation in the air with camera looking at the roof-top of a building or recorded footages on lakes \cite{P0C0-A-55}. In the first case, as developed in Ko et al. \cite{P0C0-A-50}, birds at a feeder station present a larger per-pixel variance due to changes in the background generated by the presence of a bird. Rapid background adaptation fails because birds, when present, are often moving less than the background and often end up being incorporated into it. To address this challenge, Ko et al. \cite{P0C0-A-50}\cite{P0C0-A-51} designed a background subtraction based on distributions. In the second case, Goehner et al. \cite{P0C0-A-79} compared three background models (MOG \cite{P1C2-MOG-115}, ViBe \cite{P2C1-200}, PBAS \cite{P2C1-230}) to detect events of interest within uncontrolled outdoor avian nesting video for the Wildlife@Home\protect\footnotemark[6] project. The video are taken in environments which require background models that can handle quick correction of camera lighting problems while still being sensitive enough to detect the motion of a small to medium sized animal with cryptic coloration. To address these problems, Goehner et al. \cite{P0C0-A-79} added modifications to both the ViBe and PBAS algorithms by making these algorithms second-frame-ready and by adding a morphological opening and closing filter to quickly adjust to the noise present in the videos. Moreover, Goehner et al. \cite{P0C0-A-79} also added a convex hull around the connected foreground regions to help counter cryptic coloration.

\footnotetext[6]{http://csgrid.org/csg/wildlife/}

\subsection{Fish Surveillance} 
Automated video analysis of fish is required in several type of applications: \textbf{(1)} detection of fish for recognition of species in a tank as in Baf et al. \cite{P0C0-A-201}\cite{P0C0-A-202} and in open sea as in Wang  et al. \cite{P0C0-A-69}, \textbf{(2)} detection and tracking of fish for counting in a tank in the case of industrial aquaculture as in Abe et al. \cite{P0C0-A-69-3}, \textbf{(3)} detection and tracking of fish in open sea to study their behaviors in different weather conditions as in Spampinato et al.\cite{P0C0-A-60}\cite{P0C0-A-61}\cite{P0C0-A-62}\cite{P0C0-A-63}, and \textbf{(4)} in fish catch tracking and size measurement as in Huang et al. \cite{P0C0-A-67}. In all these applications, the appearances of fish are variable as they are non-rigid and deformable objects, and therefore it makes their identification of very complex. 

\indent In 2007, Baf et al. \cite{P0C0-A-201}\cite{P0C0-A-202} studied different models (SG \cite{P1C2-SG-1}, MOG \cite{P1C2-MOG-10} and KDE \cite{P1C2-KDE-1}) for the Aqu@theque project, and concluded that SG and MOG offer good performance in time consuming and memory requirement, and that  KDE  is too slow for the application and requires too much memory. Because MOG gives better results than SG, MOG is revealed to be the most suitable for fish detection in a tank. In the EcoGrid project in 2008, Sampinato et al. \cite{P0C0-A-60}\cite{P0C0-A-62} used both a sliding Average and the Zivkovic-Heijden GMM  \cite{P1C2-MOG-36-1} for fish detection in submarine video. In 2014, Sampinato et al. \cite{P0C0-A-61} developed a specific model called textons based KDE model to address the problem that the movement of fish is fairly erratic with frequent direction changes. For videos taken by Remotely Operated Vehicles (ROVs) or Autonomous Underwater Vehicles (AUVs), Liu et al. \cite{P0C0-A-65} proposed to combine with a logical "AND" the foreground detection obtained by the running average and the three CFD. In an other work, Huang et al. \cite{P0C0-A-67} proposed a live tracking of rail-based fish catching by combining background subtraction and motion trajectories techniques in highly noisy sea surface environment. First, the  
foreground masks are obtained using SuBSENSE \cite{P2C0-40}. Then, the fish are tracked and separated from noise based on their trajectories. In 2017, Huang et al. \cite{P7C1-1000} detected moving deep-sea fishes with the MOG model and tracked
them with an algorithm combining Camshift with Kalman filtering. By analyzing the directions and trails of targets,  both distance and velocity are determined.

\subsection{Dolphins Surveillance}
Detecting and tracking social marine mammals such as bottle nose dolphins, allow researchers such as ethologists and ecologists to explain their social dynamics, predict their behavior, and measure the impact of human interference. Practically, multi-camera systems give an opportunity to record the behavior of captive dolphins with a massive dataset from which long term statistics can be extracted. In this context, Karnowski et al. \cite{P0C0-A-66} used a background subtraction method to detect dolphins over time, and to visualize the paths by which dolphins regularly traverse their environment. \\

\subsection{Honeybees Surveillance} 
Honeybees are generally filmed at the entrance of a hive to track and count different goal as follows:  \textbf{(1)} detection of external honeybee parasites as in Knauer et al. \cite{P0C0-A-70}, \textbf{(2)}  monitoring arrivals and departures at the hive entrance as in Campbell et al. \cite{P0C0-A-71}, \textbf{(3)} study of their sociability as in Kimura et al. \cite{P0C0-A-72}, and \textbf{(4)} remote pollination monitoring as in Babic et al. \cite{P0C0-A-73}. There are several reasons why honeybees detection is a difficult computer vision problem as developed in Campbell et al. \cite{P0C0-A-71} and in Babic et al. \cite{P0C0-A-73}.
\begin{enumerate}
\item Honeybees are small. In a typical image acquired from a hive-mounted camera a single bee occupies only a very small portion of the image (approximately $6 \times 14$ pixels). Honeybee detection can be easier with higher-resolution cameras or with multiple cameras placed closer to the hive entrance, but only at a substantial increase in cost as well as physical and computational complexity, limiting utility in practical setting. 
\item Honeybees are fast targets. Indeed, honeybees cover a significant distance between frames. This movement  complicates  frame-to-frame  matching  as  worker bees from a hive are virtually identical in appearance. 
\item Honeybees present motion which appears to be chaotic. Indeed, honeybees transition quickly between loitering, crawling, and flying modes of movement and change directions unpredictably; this makes it impossible to track them using one unimodal motion model. 
\item Bee hives are in outdoor scenes where lighting conditions vary significantly with the time of day, season and weather. Moreover, shadows are cast by the camera enclosure, moving bees, and moving foliage overhead. Even if it is possible to have clear lighting in the hive entry area, it demands onerous hive placement constraints vis-a-vis trees, buildings, and compass points. Artificial lighting is difficult to place in the field and could affect honeybee behavior. 
\item The scene in front of a hive is often cluttered with honeybees grouping, occluding and/or overlapping each other. Thus, the moving objects detection aware of the fact that the detected moving object can sometimes contain more than one honeybee.
\item In the case of the pollen assessment \cite{P0C0-A-73}, it is needed to at least obtain additional information that is whether the group of honeybees has a pollen load or not, when it is not possible to segment individual honeybees. 
\end{enumerate}
In 2016, Pilipovic et al. \cite{P0C0-A-74} studied different background models (frame differencing, median model  \cite{P1C1-Median-1}, MOG model \cite{P1C2-MOG-10} and Kalman filter \cite{P1C7-100}) in this field, and concluded that MOG is best suited for detection honeybees in hive entrance video. In 2017, Yang \cite{P0C0-A-74-1} confirmed the adequacy of MOG by using a modified version \cite{P1C2-MOG-562-2}.

\subsection{Spiders Surveillance} 
Iwatani et al. \cite{P0C0-A-760} proposed to design a hunting robot that mimics the spider’s hunting locomotion. To estimate the two-dimensional position and direction of a wolf spider in an observation box from video imaging, a simple background subtraction method is used because the environment is controlled. Practically, a gray scale image without the spider is taken in advance. Then, a pixel in each captured image is selected as a spider component candidate, if the difference between the background image and the captured image in grayscale is larger than a given threshold. 
 
\subsection{Lizards Surveillance}
To census of endangered lizard  species, it is important to automatically them from videos. In this context, Nguwi et al. \cite{P0C0-A-750} used a background subtraction method to detect lizards in video, followed by experiments comparing thresholding values and methods. 

\subsection{Mice Surveillance}
Mice surveillance concern mice engaging in social behavior \cite{P6C2-Dataset-20000}\cite{P7C1-100}. Most of the time experts reviewed frame-by-frame the behavior but it is time consuming. In practice, the main challenges in this kind of surveillance is the big amount of videos and thus it requires fast algorithms to detect and recognize behaviors. In this context, Rezaei and Ostadabbas \cite{P3C1-RMC-91}\cite{P3C1-RMC-92} provided a fast Robust Matrix Completion (fRMC) algorithm for in-cage mice detection using the Caltech resident intruder mice dataset \cite{P6C2-Dataset-20000}. 

\subsection{Pigs Surveillance}
Behavior analysis of livestock animals such as pigs under farm conditions is an important task to allow better management and climate regulation to improve the life of the animals. There are three main problems in farrowing pens as developed in Tu et al. \cite{P0C0-A-77}:
\begin{itemize}
\item \textbf{Dynamic background objects:} The nesting materials in the farrowing pen are often moved around because
of movements of the sow and piglets. The nesting materials can be detected as moving backgrounds. 
\item \textbf{Light changes:} Lights are often switched on and off in the pig house. In the worst case, the whole segmented image
often appears as foreground in most statistical models when the strong global illumination change suddenly occurs. 
\item \textbf{Motionless foreground objects:} Pigs and sows often sleeps over a long period. In this case, a foreground object that becomes motionless can be incorporated in the background. 
\end{itemize}

First, Guo et al. \cite{P0C0-A-76} used a Prediction mechanism-Mixture of Gaussians algorithm called PM-MOG for detection of group-housed pigs in overhead views. In an other work, Tu et al. \cite{P0C0-A-77} employed a combination of modified MOG model and
the Dyadic Wavelet Transform (DWT) in gray scale videos. This algorithm accurately extracts the shapes of a sow under complex environments. In a further work, Tu et al. \cite{P0C0-A-77-1} proposed an illumination and reflectance estimation by using an Homomorphic Wavelet Filter (HWF) and a Wavelet Quotient Image (WQI) model based on DWT.  Based on this illumination and reflectance estimation,  Tu et al. \cite{P0C0-A-77-1} used  the CFD algorithm of Li and Leung \cite{P5C1-MulF-6} which combined intensity and texture differences to detect sows in gray scale video. 

\subsection{Hinds Surveillance}
There are three main problems to detect hinds as developed in Khorrami et al. \cite{P0C0-A-75}:
\begin{itemize}
\item\textbf{Camouflage:} Hinds  may blend with the forest background by necessity 
\item \textbf{Motionless foreground objects:} Hinds may sleep over a long period. In this case, hinds can be incorporated in the background. 
\item \textbf{Rapid Motion:} Hinds  can quicly move to escape a predator. 
\end{itemize}
In camera-trap sequences, Giraldo-Zuluaga et al. \cite{P3C1-PCP-903}\cite{P3C1-PCP-904} used a multi-layer RPCA to detect hinds in forest in Colombia. Experimental results \cite{P3C1-PCP-903} against other RPCA models show the robustness of the multi-layer RPCA model in presence of challenges such as illumination changes.

\section{Intelligent Visual Observation of Natural Environments} 
\label{P0C2:sec:IVO}
There is a general use tool to detect motion in ecological environments called MotionMeerkat \cite{P0C0-A-79-1}. MotionMeerkat\protect\footnotemark[7] alleviates the process of video stream data analysis by extracting frames with motion from a video. MotionMeerkat can either use Running Gaussian Average or MOG as background model. MotionMeerkat is successful in many ecological environments but is still subject to problems such as rapid lighting changes, and camouflage. In a further work, Weinstein \cite{P7C1-4} proposed to employ a background modeling based on convolutional neural networks and developed the software DeepMeerkat\protect\footnotemark[8] for biodiversity detection. For marine environment, there is a open source framework called Video and Image Analytics for a Marine Environment (VIAME) but it currently no contains background subtraction algorithms. Thus, advanced and designed background models are needed in natural environments. Practically, natural environments such as the forest canopy, river and ocean present an extreme challenge because the foreground objects may blend with the background by necessity. Furthermore, the background itself mainly changes following its characteristics as described in the following sections.

\footnotetext[7]{http://benweinstein.weebly.com/motionmeerkat.html}
\footnotetext[8]{http://benweinstein.weebly.com/deepmeerkat.html}

\begin{landscape}
\begin{table}
\scalebox{0.60}{
\begin{tabular}{|l|l|l|l|l|l|l|} 
\hline
\scriptsize{Animal and Insect Behaviors} &\scriptsize{Type}   &\scriptsize{Background model} &\scriptsize{Background Maintenance}    &\scriptsize{Foreground Detection}       &\scriptsize{Color Space}     &\scriptsize{Strategies}        \\
\hline
\hline
\multirow{1}{*}{\scriptsize{Birds Surveillance}}
&\scriptsize{\textbf{Birds}}                                           &\scriptsize{}  &\scriptsize{}        &\scriptsize{}  &\scriptsize{} &\scriptsize{} \\
\cline{2-7}
&\scriptsize{\textbf{1) Feeder stations}}                              &\scriptsize{}  &\scriptsize{}        &\scriptsize{}  &\scriptsize{} &\scriptsize{} \\ &\scriptsize{Ko et al. (2008) \cite{P0C0-A-50}}               &\scriptsize{KDE  \cite{P1C2-KDE-45}} &\scriptsize{Blind Maintenance}  &\scriptsize{Bhattacharyya Distance} &\scriptsize{RGB} &\scriptsize{Temporal Consistency} \\
&\scriptsize{Ko et al. (2010) \cite{P0C0-A-51}}               &\scriptsize{Set of Warping Layer \cite{P0C0-A-51}}  &\scriptsize{Blind Maintenance} &\scriptsize{Bhattacharyya distance}   &\scriptsize{UYV} &\scriptsize{-} \\ 
\cline{2-7}
&\scriptsize{\textbf{2) Birds in air}}                                 &\scriptsize{}  &\scriptsize{}        &\scriptsize{}  &\scriptsize{} &\scriptsize{} \\ &\scriptsize{Shakeri and Zhang (2012) \cite{P0C0-A-55}}       &\scriptsize{Zivkovic-Heijden GMM  \cite{P1C2-MOG-36-1}} &\scriptsize{Idem GMM} &\scriptsize{Idem GMM}  &\scriptsize{RGB} &\scriptsize{Correspondence Component} \\
&\scriptsize{}          &\scriptsize{} &\scriptsize{} &\scriptsize{}  &\scriptsize{} &\scriptsize{based on Point-Tracking} \\
&\scriptsize{Nazir et al. (2017) \cite{P0C0-A-79-2}}  &\scriptsize{OpenCV Background Subtraction (WiseEye)} &\scriptsize{-} &\scriptsize{-}&\scriptsize{-} &\scriptsize{-} \\
\cline{2-7}
&\scriptsize{\textbf{3) Avian nesting}}                                &\scriptsize{}  &\scriptsize{}        &\scriptsize{}   &\scriptsize{} &\scriptsize{} \\ &\scriptsize{Goehner et al. (2015) \cite{P0C0-A-79}}          &\scriptsize{MOG \cite{P1C2-MOG-115}, ViBe \cite{P2C1-200}, PBAS \cite{P2C1-230}} &\scriptsize{Idem MOG/PBAS/ViBe}     &\scriptsize{Idem MOG/PBAS/ViBe}  &\scriptsize{RGB} &\scriptsize{Morphological Processing} \\
&\scriptsize{Dickinson et al. (2008) \cite{P0C0-A-56}}          &\scriptsize{MOG \cite{P0C0-A-56}} &\scriptsize{Idem MOG}     &\scriptsize{Idem MOG}  &\scriptsize{RGB} &\scriptsize{Spatially Coherent Segmentation \cite{P0C0-A-56-1} } \\
&\scriptsize{Dickinson et al. (2010) \cite{P0C0-A-57}}          &\scriptsize{MOG \cite{P0C0-A-56}} &\scriptsize{Idem MOG}     &\scriptsize{Idem MOG}  &\scriptsize{RGB} &\scriptsize{Spatially Coherent Segmentation \cite{P0C0-A-56-1} } \\
\hline
\multirow{1}{*}{\scriptsize{Fish Surveillance}}
&\scriptsize{\textbf{1) Tank environment}}  &\scriptsize{} &\scriptsize{}        &\scriptsize{}      &\scriptsize{} &\scriptsize{} \\ 
\cline{2-7}
&\scriptsize{\textbf{1.1) Species Identification}} &\scriptsize{} &\scriptsize{}  &\scriptsize{}      &\scriptsize{} &\scriptsize{} \\ 
&\scriptsize{Penciuc et al. (2006) \cite{P0C0-A-200}}   &\scriptsize{MOG  \cite{P1C2-MOG-10}} &\scriptsize{Idem MOG}     &\scriptsize{Idem MOG}  &\scriptsize{RGB} &\scriptsize{-} \\    
&\scriptsize{Baf et al. (2007) \cite{P0C0-A-201}}       &\scriptsize{MOG  \cite{P1C2-MOG-10}} &\scriptsize{Idem MOG}     &\scriptsize{Idem MOG}  &\scriptsize{RGB} &\scriptsize{-} \\    
&\scriptsize{Baf et al. (2007) \cite{P0C0-A-202}}       &\scriptsize{MOG  \cite{P1C2-MOG-10}} &\scriptsize{Idem MOG}     &\scriptsize{Idem MOG}  &\scriptsize{RGB} &\scriptsize{-} \\   
\cline{2-7}
&\scriptsize{\textbf{1.2) Industrial Aquaculture}}      &\scriptsize{}        &\scriptsize{}  &\scriptsize{} &\scriptsize{} &\scriptsize{}    \\ 
&\scriptsize{Pinkiewicz (2012) \cite{P0C0-A-69-5}}      &\scriptsize{Average/Median} &\scriptsize{Idem Average/Idem Median}     &\scriptsize{Idem Average/Idem Median}  &\scriptsize{RGB} &\scriptsize{-} \\  
&\scriptsize{Abe et al. (2017) \cite{P0C0-A-69-3}}      &\scriptsize{Average} &\scriptsize{Idem average}     &\scriptsize{Idem average}  &\scriptsize{RGB} &\scriptsize{-} \\  
&\scriptsize{Zhou et al. (2017) \cite{P0C0-A-69-4}}     &\scriptsize{Average} &\scriptsize{Idem average}     &\scriptsize{Idem average}  &\scriptsize{Near Infrared} &\scriptsize{-} \\  
\cline{2-7}
&\scriptsize{\textbf{2) Open sea environment}}          &\scriptsize{} &\scriptsize{}     &\scriptsize{}         &\scriptsize{} &\scriptsize{} \\ 
&\scriptsize{Spampinato et al. (2008) \cite{P0C0-A-60}}         &\scriptsize{Sliding Average/Zivkovic-Heijden GMM  \cite{P1C2-MOG-36-1}}   &\scriptsize{-}     &\scriptsize{-}  &\scriptsize{-} &\scriptsize{AND} \\    
&\scriptsize{Spampinato et al. (2010) \cite{P0C0-A-62}}         &\scriptsize{Sliding Average/Zivkovic-Heijden GMM  \cite{P1C2-MOG-36-1}}   &\scriptsize{-}     &\scriptsize{-}  &\scriptsize{-} &\scriptsize{AND} \\    
&\scriptsize{Spampinato et al. (2014) \cite{P0C0-A-63}}         &\scriptsize{GMM \cite{P1C2-MOG-10}, APMM \cite{P2C1-20}, IM \cite{P1C1-Median-75}, Wave-Back \cite{P2C5-TDM-2}}   &\scriptsize{-}     &\scriptsize{-}  &\scriptsize{-} &\scriptsize{Fish Detector} \\    
&\scriptsize{Spampinato et al. (2014) \cite{P0C0-A-61}}         &\scriptsize{Textons based KDE \cite{P0C0-A-61}}   &\scriptsize{}     &\scriptsize{} &\scriptsize{} &\scriptsize{} \\      
&\scriptsize{Liu et al. (2016) \cite{P0C0-A-65}}                &\scriptsize{Running Average-Three CFD \cite{P0C0-A-65}}  &\scriptsize{Idem RA-TCFD} &\scriptsize{Idem RA-TCFD}   &\scriptsize{RGB} &\scriptsize{AND} \\
&\scriptsize{Huang et al. (2016) \cite{P0C0-A-67}}              &\scriptsize{SuBSENSE \cite{P2C0-40}} &\scriptsize{Idem SuBSENSE} &\scriptsize{Idem SuBSENSE}   &\scriptsize{RGB-LBSP \cite{P5C1-TF-110}} &\scriptsize{Trajectory Feedback} \\
&\scriptsize{Seese et al. (2006) \cite{P0C0-A-68}}              &\scriptsize{MOG \cite{P1C2-MOG-10}/Kalman filter \cite{P1C7-103}}    &\scriptsize{Idem MOG/Kalman filter} &\scriptsize{Idem MOG/Kalman filter} &\scriptsize{Intensity} &\scriptsize{Intersection} \\
&\scriptsize{Wang et al. (2006) \cite{P0C0-A-69}}               &\scriptsize{GMM \cite{P1C2-MOG-10}}  &\scriptsize{Idem GMM} &\scriptsize{Idem GMM}  &\scriptsize{RGB} &\scriptsize{Double Local Thresholding} \\
&\scriptsize{Huang et al. (2017) \cite{P7C1-1000}}  &\scriptsize{GMM  \cite{P1C2-MOG-10}}   &\scriptsize{Idem GMM}     &\scriptsize{Idem GMM}  &\scriptsize{RGB} &\scriptsize{-} \\  
\hline
\multirow{1}{*}{\scriptsize{Dolphins Surveillance}}
&\scriptsize{Karnowski et al. (2015) \cite{P0C0-A-66}}         &\scriptsize{RPCA  \cite{P3C1-PCP-1}}   &\scriptsize{-} &\scriptsize{-}  &\scriptsize{Intensity} &\scriptsize{-} \\                
\hline
\multirow{1}{*}{\scriptsize{Honeybees Surveillance}}
&\scriptsize{Knauer et al. (2005) \cite{P0C0-A-70}}       &\scriptsize{K-Clusters \cite{P1C6-1}}      &\scriptsize{Idem K-Clusters} &\scriptsize{Idem K-Clusters}   &\scriptsize{intensity} &\scriptsize{-} \\
&\scriptsize{Campbell et al. (2008) \cite{P0C0-A-71}}     &\scriptsize{MOG \cite{P1C2-MOG-10}}         &\scriptsize{Idem MOG} &\scriptsize{Idem MOG}   &\scriptsize{RGB} &\scriptsize{-} \\
&\scriptsize{Kimura et al. (2012) \cite{P0C0-A-72}}       &\scriptsize{-}                              &\scriptsize{-} &\scriptsize{-} &\scriptsize{-} &\scriptsize{-} \\
&\scriptsize{Babic et al. (2016) \cite{P0C0-A-73}}        &\scriptsize{MOG  \cite{P1C2-MOG-10}}        &\scriptsize{Idem MOG} &\scriptsize{Idem MOG} &\scriptsize{RGB} &\scriptsize{-} \\
&\scriptsize{Pilipovic et al. (2016) \cite{P0C0-A-74}}    &\scriptsize{MOG  \cite{P1C2-MOG-10}}        &\scriptsize{Idem MOG} &\scriptsize{Idem MOG} &\scriptsize{RGB} &\scriptsize{-} \\
\hline
\multirow{1}{*}{\scriptsize{Spiders Surveillance}}
&\scriptsize{Iwatani et al. (2016) \cite{P0C0-A-760}}       &\scriptsize{Image without foreground objects}                            &\scriptsize{-} &\scriptsize{Threshold}   &\scriptsize{Intensity} &\scriptsize{-} \\
\hline
\multirow{1}{*}{\scriptsize{Lizards Surveillance}}
&\scriptsize{Nguwi et al. (2016) \cite{P0C0-A-750}}         &\scriptsize{-}                            &\scriptsize{-} &\scriptsize{-}&\scriptsize{-} &\scriptsize{}- \\
\hline
\multirow{1}{*}{\scriptsize{Pigs Surveillance}}
&\scriptsize{McFarlane and Schofield (1995) \cite{P1C1-Median-1}}  &\scriptsize{Approximated Median}    &\scriptsize{YEs} &\scriptsize{Idem Median} &\scriptsize{Intensity} &\scriptsize{-} \\
&\scriptsize{Guo et al. (2015) \cite{P0C0-A-76}}         &\scriptsize{PM-MOG \cite{P0C0-A-76}}         &\scriptsize{Idem MOG} &\scriptsize{Idem MOG} &\scriptsize{RGB} &\scriptsize{Prediction Mechanism} \\     
&\scriptsize{Tu et al. (2014) \cite{P0C0-A-77}}          &\scriptsize{MOG-DWT \cite{P0C0-A-77}}        &\scriptsize{Idem MOG} &\scriptsize{Idem MOG} &\scriptsize{Intensity/Texture} &\scriptsize{OR} \\
&\scriptsize{Tu et al. (2015) \cite{P0C0-A-77-1}}        &\scriptsize{CFD \cite{P5C1-MulF-6}}          &\scriptsize{-} &\scriptsize{Difference}   &\scriptsize{RGB} &\scriptsize{Illumination and Reflectance} \\ 
&\scriptsize{}    &\scriptsize{} &\scriptsize{} &\scriptsize{}  &\scriptsize{} &\scriptsize{ Estimation} \\ 
\hline
\multirow{1}{*}{\scriptsize{Hinds  Surveillance}}
&\scriptsize{Khorrami et al. (2012) \cite{P0C0-A-75}}       &\scriptsize{RPCA  \cite{P3C1-PCP-1}}      &\scriptsize{-} &\scriptsize{-}&\scriptsize{Intensity} &\scriptsize{-} \\
&\scriptsize{Giraldo-Zuluaga et al. (2017) \cite{P3C1-PCP-903}}  &\scriptsize{Multi-Layer RPCA  \cite{P3C1-PCP-903}}      &\scriptsize{-} &\scriptsize{-}   &\scriptsize{RGB} &\scriptsize{-} \\
&\scriptsize{Giraldo-Zuluaga et al. (2017) \cite{P3C1-PCP-904}}  &\scriptsize{Multi-Layer RPCA  \cite{P3C1-PCP-903}}      &\scriptsize{-} &\scriptsize{-}   &\scriptsize{RGB} &\scriptsize{-} \\
\hline
\end{tabular}}
\caption{Background models used for intelligent visual observation of animals and insects: An Overview. "-" indicated that the background model used in not indicated in the paper.} \centering
\label{P0C2:AIB}
\end{table}
\end{landscape}

\subsection{Forest Environments} 
The aim is to detect humans or animals but the motion of the foliage generate rapid transitions between light and shadow. Furthermore, humans or animals can be partially occluded by branches. First, Boult et al. \cite{P0C0-Sensors-SC-2} addressed these problems to detect humans into the woods with an omnidirectional cameras. In 2017, Shakeri and Zhang \cite{P7C1-1} employed a robust PCA method to detect animals in clearing zones. A camera trap is used to capture the videos. Experiments show that the proposed method outperforms most of the previous RPCA methods on the illumination change dataset (ICD). Yousif et al. \cite{P7C1-2} designed a joint background modeling and deep learning classification to detect and distinguish human and animals. The block-wise background modeling employ three features (intensity, histogram, Local Binary Pattern \cite{P0C0-Survey-26}, and Histogram of Oriented Gradient (HOG) \cite{P0C0-Survey-26}) to be robust against waving trees. Janzen et al. \cite{P7C1-3} detected movement in a region via background subtraction, image thresholding, and fragment histogram analysis. This system reduced the number of images for human consideration to one third of the input set with a success rate of proper identification of ungulate crossing between $60\%$ and $92\%$, which is suitable in larger study context.

\subsection{River Environments} 
The idea is to detect foreign objects in the river (bottles, floating woods, etc... ) for \textbf{(1)} the preservation of the river or for \textbf{(2)} the preservation of the civil infrastructures such as bridges and dams on the rivers \cite{P0C0-A-90}\cite{P0C0-A-91}. In the first case, foreign objects pollute the environment and then animals are affected \cite{P1C7-104}. In the second case, foreign objects such as fallen trees, bushes, branches of fallen trees and other small pieces of wood can damage bridges and dams on the rivers. The risk of damage by trees is directly proportional to their size. Larger fallen trees are more dangerous than the smaller parts of fallen trees. These trees often remain wedged against the pillars of bridges  and help in the accumulation of small branches, bushes and debris around. In these river environments, both background water waves and floating objects of interest are in motion. Moreover, the flow of river water varies from the normal flow during floods, and thus it causes more motion. In addition, small waves and water surface illumination, cloud shadows, and similar phenomena add non-periodical background changes. In 2003, Zhong et al. \cite{P1C7-104} used Kalman filter to detect bottles in waving river for an application of type (1). In 2012, Ali et al. \cite{P0C0-A-90} used a space-time spectral mode for an application of type (2). In a further work, Ali et al. \cite{P0C0-A-91} added to the MOG model a rigid motion model to detect floating woods. 

\subsection{Ocean Environments} 
The aim is to detect ships \textbf{(1)} ships for the optimization of the traffic which is the case in most of the applications,  \textbf{(2)} foreign objects to avoid the collision with foreign objects \cite{P0C0-A-29}, and \textbf{(3)} foreign people (intruders) because ships are in danger of pirate attacks both in open waters and in a harbor environment \cite{P0C0-A-29-4}. These scenes are more challenging than calm water scenes because of the waves breaking near the shore. Moreover boat wakes and weather issues contribute to generate a highly dynamic background. Practically, the motion of the waves generate false positive detections in the foreground detection \cite{P0C0-A-95}. In a valuable, Prasad et al. \cite{P0C0-A-29-2} \cite{P0C0-A-29-3} provided a list of the challenges met in videos acquired in maritime environments, and applied on the Singapore-Marine dataset the 34 algorithms that participated in the ChangeDetection.net competition. Experimental results \cite{P0C0-A-29-2} show that all these methods are ineffective, and produced false positives in the water region or false negatives while suppressing
water background. Practically, the challenges can classified as follows as developed in Prasad et al. \cite{P0C0-A-29-2}:
\begin{itemize}
\item Weather and illumination conditions such as bright sunlight, twilight conditions, night, haze, rain, fog, etc..
\item The solar angles induce different speckle and glint conditions in the water. 
\item Tides also influence the dynamicity of water. 
\item Situations that affect the visibility influence the contrast, statistical distribution of sea and water, and visibility of
far located objects. 
\item Effects such as speckle and glint create non-uniform background statistics which need extremely
complicated modeling such that foreground is not detected as the background and vice versa. 
\item Color gamuts for illumination conditions such as night (dominantly dark), sunset (dominantly yellow and red), and bright daylight
(dominantly blue), and hazy conditions (dominantly gray) vary significantly. 
\end{itemize}

\subsection{Submarine Environments} 
There are three kinds of aquatic underwater environments (also called underwater environments): \textbf{(1)} swimming pools, \textbf{(2)} tank environments for fish, and \textbf{(3)} open sea environments. 

\subsubsection{Swimming pools}
In swimming pools, there are water ripples, splashes and specular reflections. First, Eng et al. \cite{P0C0-A-40} designed a block-based median background model and used the CIELa*b* color space for outdoor pool environments to detect swimmers under amid reflections, ripples, splashes and rapid lighting changes. Partial occlusions are resolved using a Markov Random Field framework that enhances the tracking capability of the system. In a further work, Eng et al. \cite{P0C0-A-40-1} employed a region-based single multivariate Gaussian model to detect swimmers, and used the CIELa*b* color space as in Eng et al. \cite{P0C0-A-40}. Then, a spatio-temporal filtering scheme enhanced the detection because swimmers are often partially hidden by specular reflections of artificial nighttime lighting.  In an other work, Lei and Zhao \cite{P1C7-118} employed a Kalman filter \cite{P1C7-103} to deal with light spot and water ripple. In a further work, Chan et al. \cite{P0C0-A-42} detected swimmers by computing dense optical flow and the MOG model on video sequences captured at daytime, and nighttime, and of different swimming styles (breaststroke, freestyle, backstroke). For private swimming, Peixoto et al. \cite{P0C0-A-44} combined the mean model and the two CFD, and used the HSV invariant color model space. 

\subsubsection{Tank environments}
There are three kinds of tanks: \textbf{(1)} tanks in aquarium which reproduce the maritime environment, \textbf{(2)} tanks for industrial aquaculture, and \textbf{(3)} tanks for studies of fish's behaviors. The challenges met in tank environments can be classified as follows
\begin{itemize}
\item \textbf{Challenges due the environments:} Illumination changes are owed to the ambient light, the spotlights which light the tank from the inside and from the outside, the movement of the water due to fish and the continuous renewal of the water. In addition for tank in aquarium, moving algae generate false detections as developed in Baf et al. \cite{P0C0-A-201}\cite{P0C0-A-202}.
\item \textbf{Challenges due fish:} The movement of fish is different due to their species and the kind of tank. Furthermore, their number is different. In tank for aquarium, there are different species as in Baf et al. \cite{P0C0-A-201} where there are ten species of tropical fish. However, fish of the same species tend to have the same behavior. But, there are several species which swim at different depths. Furthermore, they can be occluded by algae or other fish. In an other way in industrial aquaculture, the number of fish is bigger than aquarium, and all the fish are from the same species and thus they have the same behavior. For example in Abe et al. \cite{P0C0-A-69-3}, there were 250 Japanese rice fish, of which 170-190 were detected by naked eye observation in the visible area during the recording period. Furthermore, these fish tend to swim at various depths in the tank. 
\end{itemize}

\subsubsection{Open sea environments}
In underwater open sea environments, the degree of luminosity and water flow vary depending upon the weather and the time of the day. The water may 
also have varying degrees of clearness and cleanness as developed in Spampinato et al. \cite{P0C0-A-60}. In addition in subtropical waters, algae 
grow rapidly and appear on camera lens. Consequently, different degrees of greenish and bluish videos are produced. In order to reduce the presence of the algae, lens is frequently and manually cleaned. Practically, the challenges in open sea environments can classified as developed in Kavasidis and Palazzo  \cite{P0C0-A-68-1} and Spampinato et al. \cite{P0C0-A-63}: 
\begin{itemize}
\item \textbf{Light changes:} Every possible lighting conditions need to be taken into account because the video feeds are captured during the
whole day and the background subtraction algorithms should consider the light transition. 
\item \textbf{Physical phenomena:} Image contrast is influenced by various physical phenomena such as typhoons, storms or sea currents which can easily compromise the contrast and the clearness of the acquired videos. 
\item \textbf{Murky water:} It is important to consider that the clarity of the water during the
day could change due to the drift and the presence of plankton in order to investigate the movements of fish in their natural
habitat. Under these conditions, targets that are not fish might be detected as false positives. 
\item \textbf{Grades of freedom:} In underwater videos the moving objects can move in all three dimensions whilst videos containing traffic 
or pedestrians videos are virtually confined in two dimensions. 
\item  \textbf{Algae formation on camera lens:} The contact of sea water with the camera’s lens facilitates the quick formation of algae on top of camera. 
\item \textbf{Periodic and multi-modal background:} Arbitrarily moving objects such as stones and periodically moving ojects such as plants subject to flood-tide and drift can generated false positive detections.
\end{itemize}

\indent In a comparative evaluation done in 2012 on the Fish4Knowledge dataset \cite{P6C2-Dataset-2020}, Kavasidis and Palazzo \cite{P0C0-A-68-1} evaluated the performance of six state-of-the-art background subtraction algorithms (GMM \cite{P1C2-MOG-10}, APMM \cite{P2C1-20}, IM \cite{P1C1-Median-75}, Wave-Back \cite{P2C5-TDM-2}, Codebook \cite{P1C6-100} and ViBe \cite{P2C1-200}) in the task of fish detection in unconstrained and underwater video. Kavasidis and Palazzo \cite{P0C0-A-68-1} concluded that:
\begin{enumerate}
\item At the blob level, the performance is generally good in videos which presented scenes under normal weather and lighting conditions and more or less static backgrounds, except from the Wave-Back algorithm. On the other hand, the ViBe algorithm excelled in nearly all the videos. GMM and APMM performed somewhere in the middle, with the GMM algorithm resulting slightly better than the PMM algorithm. The codebook algorithm gave the best results in the high resolution videos. \ 
\item At the pixel level, all the algorithms show a good pixel detection rate, i.e. they are able to correctly identify pixels belonging to an object, with values in the range between $83.2\%$ for the APMM algorithm and $83.4\%$ for ViBe. But, they provide a relatively high pixel false alarm rate, especially the Intrinsic Model and Wave-back algorithms when the contrast of the video was low, a condition encountered during low light scenes and when violent weather phenomena were present (typhoons and storms). 
\end{enumerate}

\indent In a further work in 2017, Radolko et al. \cite{P6C2-Dataset-2025} identified the following five main difficulties:
\begin{itemize}
\item \textbf{Blur:} It is due to the forward scattering in water and makes it impossible to get a sharp image. 
\item \textbf{Haze:} Small particles in the water cause back scatter. The effect is similar to a sheer veil in front of the
scene.
\item \textbf{Color Attenuation:} Water absorbs light stronger than air. Also, the absorption effect depends on the wavelength of
the light and this leads to underwater images with strongly distorted and mitigated colors.
\item \textbf{Caustics:} Light reflections on the ground caused by ripples on the water surface. They are similar to strong, fast moving shadows which makes them very hard to differentiate from dark objects. 
\item \textbf{Marine Snow:} Small floating particles which strongly reflect light. Mostly they are small enough that they are
filtered out during the segmentation process, however, they still corrupt the image and complicate for example the modeling of the static background. 
\end{itemize}
Experimental results  \cite{P6C2-Dataset-2025} done on the dataset UnderwaterChangeDetection.eu \cite{P6C2-Dataset-2025} show that GSM \cite{P1C2-MOG-765} gives better performance than MOG-EM \cite{P1C2-MOG-90}, Zivkovic-Heijden GMM (also called ZHGMM) \cite{P1C2-MOG-36-1} and EA-KDE (also called KNN) \cite{P1C2-KDE-30}. In an other work, Rout et al. \cite{P1C2-MOG-776} designed a spatio-Contextual GMM (SC-GMM) that outperforms 18 background subtraction algorithms such as \textbf{1)} classical algorithms like the original MOG, the original KDE and the original PCA, and \textbf{2)} advanced algorithms like DPGMM (VarDMM) \cite{P2C1-11}, PBAS \cite{P2C1-230}, PBAS-PID \cite{P2C1-231}, SuBSENSE \cite{P2C0-40}  and SOBS \cite{P1C5-404} both on the Fish4Knowledge dataset \cite{P6C2-Dataset-2020} and the dataset UnderwaterChangeDetection.eu \cite{P6C2-Dataset-2025}. \\

Thus, natural environments involve multi-modal backgrounds, and changes of the background structure need to be captured from the background model to avoid a big amount of false detection rate. Practically, events such as CVAUI (Computer Vision for Analysis of Underwater Imagery) 2015 and 2016 in conjunction with ICPR addressed the problem of the detection in ocean surface and underwater environments. \\

\section{Miscellaneous Applications} 
\label{P0C2:sec:OMA}

\subsection{Visual Analysis of Human Activities} 
Background subtraction is also used for visual analysis of human activities like in sport \textbf{(1)} when important decisions need to be made quickly, \textbf{(2)} for precise analysis of athletic performance, and \textbf{(3)} for surveillance in dangerous activities. For example, John et al. \cite{P0C0-A-502} provide a system that allow a coach to obtain real-time feedbacks to ensure that the routine is performed in a correct manner. During the initialisation stage, the stationary camera captures the background image without the user, and then each current image is subtracted to obtain the silhouette. This method is the simplest way to obtain  the silhouette and is useful in this context as it is indoor scene with control on the light. In practice, the method was implemented inside an Augmented Reality (AR) desktop app that employs a single RGB camera. The detected body pose image is compared against the exemplar body pose image at specific intervals. The pose matching function is reported to have an accuracy of $93.67\%$. In an other work, Tamas \cite{P0C0-A-500} designed a system for estimating the pose of athletes exercising on indoor rowing machines. Practically, Zivkovic-Heijden GMM \cite{P1C2-MOG-36-1} and Zivkovic-KDE \cite{P1C2-KDE-30} available in OpenCV were tested with success but with the main drawbacks that these methods mark the shadows projecting to the background as foreground. Then, Tamas \cite{P0C0-A-500} developed a fast and accurate background subtraction method which allows to extract the head, shoulder, elbow, hand, hip, knee, ankle and back positions of a rowing athlete. For surveillance, Bastos \cite{P0C0-A-501} employed the original MOG to detect surfers on waves in Ribeira d’Ilhas beach (Portugal). The proposed system obtains a false positive rate of 1.77 for a detection rate of $90\%$ but the amount of memory and computational time required to process a video sequence is the main drawback of this system.

\begin{landscape}
\begin{table}
\scalebox{0.60}{
\begin{tabular}{|l|l|l|l|l|l|l|} 
\hline
\scriptsize{Applications} &\scriptsize{Type}   &\scriptsize{Background model} &\scriptsize{Background Maintenance}    &\scriptsize{Foreground Detection}  &\scriptsize{Color Space}     &\scriptsize{Strategies}        \\
\hline
\hline
\scriptsize{\textbf{Visual Hull Computing}}&\scriptsize{}  &\scriptsize{}   &\scriptsize{}  &\scriptsize{}    &\scriptsize{} &\scriptsize{} \\
\scriptsize{\textbf{1) Image-based Modeling}}&\scriptsize{}  &\scriptsize{}   &\scriptsize{}  &\scriptsize{}    &\scriptsize{} &\scriptsize{} \\
\scriptsize{Matusik et al. (2000) \cite{P1C2-SG-1}}  &\scriptsize{Marker Free}  &\scriptsize{MOG \cite{P1C2-MOG-1}} &\scriptsize{Idem MOG}  &\scriptsize{Idem MOG} &\scriptsize{RGB} &\scriptsize{-} \\
\scriptsize{\textbf{2) Optical Motion Capture}}  &\scriptsize{}  &\scriptsize{}   &\scriptsize{}  &\scriptsize{}    &\scriptsize{} &\scriptsize{} \\
\scriptsize{Wren et al. (1997) \cite{P1C2-SG-1}}        &\scriptsize{Marker Free (Pfinder \cite{P1C2-SG-1})}  &\scriptsize{SG \cite{P1C2-SG-1}}   &\scriptsize{Idem SG}  &\scriptsize{Idem SG} &\scriptsize{YUV} &\scriptsize{-} \\
\scriptsize{Horprasert et al. (1998) \cite{P0C0-A-110}} &\scriptsize{Marker Free}  &\scriptsize{W4 \cite{P0C0-A-110-1}}  &\scriptsize{Idem W4}  &\scriptsize{Idem W4} &\scriptsize{Intensity} &\scriptsize{-} \\
\scriptsize{Horprasert et al. (1999) \cite{P0C0-A-111}} &\scriptsize{Marker Free}  &\scriptsize{Codebook \cite{P0C0-A-111}} &\scriptsize{Idem Codebook \cite{P0C0-A-111}}  &\scriptsize{Idem Codebook \cite{P0C0-A-111}} &\scriptsize{RGB} &\scriptsize{Shadow Detection} \\
\scriptsize{Horprasert et al. (2000) \cite{P0C0-A-112}} &\scriptsize{Marker Free}  &\scriptsize{Codebook \cite{P0C0-A-111}} &\scriptsize{Idem Codebook \cite{P0C0-A-111}}  &\scriptsize{Idem Codebook \cite{P0C0-A-111}} &\scriptsize{RGB} &\scriptsize{Shadow Detection} \\
\scriptsize{Mikic et al. (2002) \cite{P0C0-A-130}} &\scriptsize{Marker Free}  &\scriptsize{Codebook \cite{P0C0-A-111}} &\scriptsize{Idem Codebook \cite{P0C0-A-111}}  &\scriptsize{Idem Codebook \cite{P0C0-A-111}} &\scriptsize{RGB} &\scriptsize{Shadow Detection} \\
\scriptsize{Mikic et al. (2003) \cite{P0C0-A-131}} &\scriptsize{Marker Free}  &\scriptsize{Codebook \cite{P0C0-A-111}} &\scriptsize{Idem Codebook \cite{P0C0-A-111}}  &\scriptsize{Idem Codebook \cite{P0C0-A-111}} &\scriptsize{RGB} &\scriptsize{Shadow Detection} \\
\scriptsize{Chu et al. (2003) \cite{P0C0-A-140}} &\scriptsize{Marker Free}  &\scriptsize{SG \cite{P1C2-SG-10}} &\scriptsize{Idem SG}  &\scriptsize{Idem SG} &\scriptsize{HSV} &\scriptsize{-} \\
\scriptsize{Carranza et al. (2003) \cite{P0C0-A-100}}   &\scriptsize{Marker Free}  &\scriptsize{SG \cite{P1C2-SG-1}}        &\scriptsize{Idem SG}  &\scriptsize{Idem SG} &\scriptsize{YUV}  &\scriptsize{Shadow Detecion} \\
\scriptsize{Guerra-Filho (2005) \cite{P0C0-A-101}}      &\scriptsize{Marker Detection}  &\scriptsize{Median}     &\scriptsize{Idem Median}  &\scriptsize{Idem Median} &\scriptsize{RGB } &\scriptsize{-} \\
\scriptsize{Kim et al. (2007) \cite{P0C0-A-141}}       &\scriptsize{Marker Free}  &\scriptsize{SGG (GGF) \cite{P1C2-SGG-1}}     &\scriptsize{Idem SGG} &\scriptsize{Idem SGG} &\scriptsize{RGB}  &\scriptsize{Small regions suppression} \\
\scriptsize{Park et al. (2009) \cite{P0C0-A-120}}       &\scriptsize{Photorealistic Avatars}  &\scriptsize{Codebook with online MoG}     &\scriptsize{Idem Codebook} &\scriptsize{Idem Codebook} &\scriptsize{RGB}  &\scriptsize{Markov Random Field (MRF)} \\
\hline
\hline
\scriptsize{\textbf{Human-Machine Interaction (HMI)}}  &\scriptsize{}  &\scriptsize{} &\scriptsize{}  &\scriptsize{}  &\scriptsize{} &\scriptsize{} \\
\scriptsize{\textbf{1) Arts}} &\scriptsize{}  &\scriptsize{} &\scriptsize{}  &\scriptsize{}  &\scriptsize{} &\scriptsize{} \\
\scriptsize{Levin (2006) \cite{P0C0-A-360}}   &\scriptsize{Art}  &\scriptsize{-} &\scriptsize{-}  &\scriptsize{-}  &\scriptsize{-} &\scriptsize{-} \\
\scriptsize{\textbf{2) Games}} &\scriptsize{}  &\scriptsize{} &\scriptsize{}  &\scriptsize{}  &\scriptsize{} &\scriptsize{} \\
\scriptsize{\textbf{3) Ludo-Multimedia Applications}} &\scriptsize{}  &\scriptsize{} &\scriptsize{}  &\scriptsize{}  &\scriptsize{} &\scriptsize{} \\
\scriptsize{Penciuc et al. (2006) \cite{P0C0-A-202}} &\scriptsize{Fish Detection} &\scriptsize{MOG} &\scriptsize{Idem MOG} &\scriptsize{Idem MOG} &\scriptsize{RGB}  &\scriptsize{-} \\
\scriptsize{Baf et al.(2007) \cite{P0C0-A-201}}      &\scriptsize{Fish Detection} &\scriptsize{MOG} &\scriptsize{Idem MOG}  &\scriptsize{Idem MOG} &\scriptsize{RGB}  &\scriptsize{-} \\
\scriptsize{Baf et al.(2007) \cite{P0C0-A-200}}      &\scriptsize{Fish Detection} &\scriptsize{MOG} &\scriptsize{Idem MOG}  &\scriptsize{Idem MOG} &\scriptsize{RGB}  &\scriptsize{-} \\
\hline
\hline
\scriptsize{\textbf{Vision-based Hand Gesture Recognition}}  &\scriptsize{}  &\scriptsize{} &\scriptsize{}  &\scriptsize{}  &\scriptsize{} &\scriptsize{}   \\
\scriptsize{\textbf{1) Human Computer Interface (HCI)}} &\scriptsize{}  &\scriptsize{} &\scriptsize{}  &\scriptsize{}  &\scriptsize{} &\scriptsize{} \\
\scriptsize{Park and Hyun (2013) \cite{P0C0-A-370}}   &\scriptsize{Hand Detection}  &\scriptsize{Average} &\scriptsize{Selective Maintenance}  &\scriptsize{Idem Running Average}  &\scriptsize{Intensity} &\scriptsize{-} \\
\scriptsize{Stergiopoulou et al. (2014) \cite{P0C0-A-371}}   &\scriptsize{Hand Detection}  &\scriptsize{Three CFD/BGS \cite{P0C0-A-4-10}} &\scriptsize{Idem CFD/BGS\cite{P0C0-A-4-10}}  &\scriptsize{Idem CFD/BGS\cite{P0C0-A-4-10}}  &\scriptsize{RGB} &\scriptsize{-} \\
\scriptsize{\textbf{2) Behavior Analysis}} &\scriptsize{}  &\scriptsize{} &\scriptsize{}  &\scriptsize{}  &\scriptsize{} &\scriptsize{} \\
\scriptsize{Perrett et al. (2016) \cite{P0C0-A-374}}   &\scriptsize{Hand Detection}  &\scriptsize{PBAS \cite{P2C1-230}} &\scriptsize{Idem PBAS \cite{P2C1-230}}  &\scriptsize{PBAS \cite{P2C1-230}}  &\scriptsize{Intensity} &\scriptsize{Post-processing (Median filter)} \\
\scriptsize{\textbf{3) Sign Language Interpretation and Learning}} &\scriptsize{}  &\scriptsize{} &\scriptsize{}  &\scriptsize{}  &\scriptsize{} &\scriptsize{} \\
\scriptsize{Elsayed  et al. (2015) \cite{P0C0-A-372}}   &\scriptsize{Hand Detection}  &\scriptsize{First Frame without Foreground Objects} &\scriptsize{Running Average}  &\scriptsize{Idem Running Average}  &\scriptsize{YCrCb} &\scriptsize{-} \\
\scriptsize{\textbf{4) Robotics}} &\scriptsize{}  &\scriptsize{} &\scriptsize{}  &\scriptsize{}  &\scriptsize{} &\scriptsize{} \\
\scriptsize{Khaled et al. (2015) \cite{P0C0-A-373}}   &\scriptsize{Hand Detection}  &\scriptsize{Average} &\scriptsize{Running Average}  &\scriptsize{Idem Running Average}  &\scriptsize{Intensity/Color} &\scriptsize{Contour Extraction Algorithm} \\\hline
\hline
\scriptsize{\textbf{Video Coding}}  &\scriptsize{}  &\scriptsize{}  &\scriptsize{} &\scriptsize{} &\scriptsize{}     &\scriptsize{} \\
\scriptsize{Chien et al. (2002) \cite{P0C0-A-300}}  &\scriptsize{MPEG-4 (QCIF Format)}    &\scriptsize{Progressive Generation (CFD \cite{P5C1-CDFeatureI-300})}                        &\scriptsize{Progressive Maintenance} &\scriptsize{CD \cite{P5C1-CDFeatureI-300}} &\scriptsize{Intensity}  &\scriptsize{Post Processing/Shadow Detection} \\
\scriptsize{Paul et al. (2010) \cite{P0C0-A-310-1}}    &\scriptsize{H.264/AVC (CIF Format)}  &\scriptsize{MOG on decoded pixel intensities}     &\scriptsize{Selective Maintenance} &\scriptsize{Block-based Difference} &\scriptsize{Intensities}  &\scriptsize{-} \\
\scriptsize{Paul et al. (2013) \cite{P5C1-VC-1}}    &\scriptsize{H.264/AVC (CIF Format)}  &\scriptsize{MOG on decoded pixel intensities}     &\scriptsize{Selective Maintenance} &\scriptsize{Block-based Difference} &\scriptsize{Intensities}  &\scriptsize{-} \\
\scriptsize{Paul et al. (2013) \cite{P0C0-A-310-2}}    &\scriptsize{H.264/AVC}  &\scriptsize{MOG \cite{P1C2-MOG-132}}     &\scriptsize{-} &\scriptsize{-} &\scriptsize{Color}  &\scriptsize{-} \\
\scriptsize{Zhang et al. (2010) \cite{P0C0-A-320-1}} &\scriptsize{H.264/AVC}   &\scriptsize{Non-Parametric Background Generation \cite{P0C0-A-320-1}} &\scriptsize{Idem BG \cite{P0C0-A-320-1}} &\scriptsize{Idem BG \cite{P0C0-A-320-1}}  &\scriptsize{Color}  &\scriptsize{-} \\
\scriptsize{Zhang et al. (2012) \cite{P0C0-A-320}} &\scriptsize{H.264/AVC}   &\scriptsize{SWRA \cite{P0C0-A-320}} &\scriptsize{Selective Maintenance} &\scriptsize{-} &\scriptsize{Color}  &\scriptsize{-} \\
\scriptsize{Chen et al. (2012) \cite{P0C0-A-321}}   &\scriptsize{H.264/AVC}  &\scriptsize{Average}  &\scriptsize{Selective Maintenance} &\scriptsize{-} &\scriptsize{Color}  &\scriptsize{-} \\
\scriptsize{Han et al. (2012)\cite{P0C0-A-322}}     &\scriptsize{H.264/AVC} &\scriptsize{-} &\scriptsize{-} &\scriptsize{-} &\scriptsize{Color} &\scriptsize{Panorama Background/Motion Compensation} \\
\scriptsize{Zhang et al. (2012) \cite{P0C0-A-323}}  &\scriptsize{H.264/AVC}  &\scriptsize{MSBDC \cite{P0C0-A-323}}     &\scriptsize{Selective Maintenance} &\scriptsize{-} &\scriptsize{Color}  &\scriptsize{-} \\
\scriptsize{Geng et al. (2012)\cite{P0C0-A-324}}   &\scriptsize{H.264/AVC}   &\scriptsize{SWRA \cite{P0C0-A-320}}     &\scriptsize{Selective Maintenance} &\scriptsize{-} &\scriptsize{Color}  &\scriptsize{-} \\
\scriptsize{Zhang et al. (2014) \cite{P5C1-VC-10}}  &\scriptsize{}  &\scriptsize{BMAP \cite{P5C1-VC-10}} &\scriptsize{Selective Maintenance} &\scriptsize{-} &\scriptsize{Color}  &\scriptsize{-} \\
\scriptsize{Zhao et al. (2014) \cite{P5C1-VC-11}}   &\scriptsize{HEVC}  &\scriptsize{BFDS \cite{P5C1-VC-11}}     &\scriptsize{-} &\scriptsize{-} &\scriptsize{Color}  &\scriptsize{-} \\
\scriptsize{Zhang et al. (2014) \cite{P0C0-A-327}}  &\scriptsize{HEVC}  &\scriptsize{Running Average}     &\scriptsize{-} &\scriptsize{-} &\scriptsize{Color}  &\scriptsize{-} \\
\scriptsize{Chakraborty  et al. (2014) \cite{P5C1-VC-2}}    &\scriptsize{HEVC}  &\scriptsize{KDE/Median}     &\scriptsize{Selective Maintenance} &\scriptsize{-} &\scriptsize{Intensity}  &\scriptsize{-} \\
\scriptsize{Chakraborty  et al. (2014) \cite{P5C1-VC-3}}    &\scriptsize{HEVC}  &\scriptsize{KDE/Median}     &\scriptsize{Selective Maintenance} &\scriptsize{-} &\scriptsize{Intensity}  &\scriptsize{-} \\
\scriptsize{Chakraborty  et al. (2017) \cite{P0C0-A-312}}    &\scriptsize{HEVC}  &\scriptsize{KDE/Median}     &\scriptsize{Selective Maintenance} &\scriptsize{-} &\scriptsize{Intensity}  &\scriptsize{Scene Adaptive Non-Parametric Technique} \\
\scriptsize{Chen  et al. (2012) \cite{P3C1-OA-32}}    &\scriptsize{H.264/AVC}  &\scriptsize{RPCA (LRSD \cite{P3C1-OA-32})} &\scriptsize{-} &\scriptsize{-} &\scriptsize{Intensity}  &\scriptsize{-} \\
\scriptsize{Guo et al. (2013) \cite{P3C1-OA-31}}     &\scriptsize{H.264/AVC}  &\scriptsize{RPCA (Dictionary Learning \cite{P3C1-OA-31})} &\scriptsize{-} &\scriptsize{-} &\scriptsize{Color}  &\scriptsize{-} \\
\scriptsize{Zhao et al. (2013) \cite{P3C1-OA-30}}     &\scriptsize{HEVC}  &\scriptsize{RPCA (Adaptive Lagrange Multiplier \cite{P3C1-OA-30})} &\scriptsize{-} &\scriptsize{-} &\scriptsize{Color}  &\scriptsize{-} \\
\hline
\end{tabular}}
\caption{Background models used for optical motion capture and video coding: An Overview. "-" indicated that the background model used in not indicated in the paper.} \centering
\label{P0C2:OMC}
\end{table}
\end{landscape}

\subsection{Optical motion capture} 
\label{P0C2:sec:OMC}
The aim is to obtain a 3D model of an actors filmed by a system with multi-cameras wich can require markers \cite{P0C0-A-101} or not \cite{P0C0-A-100}\cite{P0C0-A-110}\cite{P0C0-A-130}\cite{P0C0-A-131}\cite{P0C0-A-140}\cite{P0C0-A-141}.
Because it is impossible to have a rigourous 3D reconstruction of a human model, a 3D voxel approximation \cite{P0C0-A-130}\cite{P0C0-A-131} obtained by shape-from-silhouette (also called visual hull) is computed with the silhouettes obtained from each camera. Then, the movememts are tracked and reproduced on human body model called avatar. Optical motion capture is used in computer game, virtual clothing application and virtual reality. In all optical motion capture systems, it requires in its first step to obtain a full and precise capture of human sihouette and movements from the different point of view provided by the multiple cameras. One common technique for obtaining silhouettes also used in television weather forecasts and for cinematic special effects for background substitution is chromakeying (also called bluescreen matting) which is based on the fact that the actual scene background is a single uniform color that is unlikely to appear in foreground objects. Foreground objects can then be segmented from the background by using color comparisons but chromakey techniques do not admit arbitrary backgrounds, which is a severe limitation as developed by Buehler et al. \cite{P0C0-A-150}. Thus, background subtraction is more suitable to obtain the silhouette. Practically, silhouettes are then extracted in each view by background subtraction, and thus this step is also called silhouette detection or silhouette extraction. Because the acquisition is made in indoor scenes, the background model required can be uni-modal, and shadows and highlights are the main challenges in this application. In this context, Wren et al. \cite{P1C2-A-102} used a single Gaussian model in YUV color space whilst Horprasert et al. \cite{P0C0-A-111}\cite{P0C0-A-112} used a statistical model with shadow and highlight suppression. Furthermore, Table \ref{P0C2:OMC} shows an overview of the different publications in the field with information about the background model, the background maintenance, the foreground detection, the color space and the strategies used by the authors. \\

\subsection{Human-machine interaction} 
\label{P0C2:sec:HMI}
Several applications need interactions between human and machine through a video acquired in real-time by fixed cameras such as games (Microsoft's Kinect) and ludo-applications such as Aqu@theque \cite{P0C0-A-200}\cite{P0C0-A-201}\cite{P0C0-A-202}.

\begin{itemize}
\item \textbf{Arts and Games:} First, the person's body pixels are located with background subtraction, and then this information is used as the basis for graphical responses in interactive systems as developed in Levin et al. \cite{P0C0-A-360} website\protect\footnotemark[9]). In 2003, Warren \cite{P0C0-A-361} presented a vocabulary of various essential interaction techniques which can use this kind of body-pixel data. These schema are useful in "mirror-like" contexts, such as Myron Krueger's Videoplace \protect\footnotemark[10]), or video games like the PlayStation Eye-Toy, in which the participant can observe his own image or silhouette composited into a virtual scene. 

\footnotetext[9]{http://www.flong.com/texts/essays/essaycvad/}
\footnotetext[10]{http://www.medienkunstnetz.de/works/videoplace/}

\item \textbf{Ludo-Multimedia Applications:} In this type of applications, the user can select a moving object of interest on a screen, and then information are provided. A representative example is the Aqu@theque project which allows a visitor of an aquarium to select on an interactive interface fishes that are filmed on line by a remote video camera. This interface is a touch screen divided into two parts. The first one shows the list of fishes present in the tank and is useful all the time. The filmed scene is visualized in the remaining part of the screen. The computer can supply information about fishes selected by the user with his finger. A fish is then automatically identified and some educational information about it is put on the screen. The user can also select each identified fish whose virtual representation is shown on another screen. This second screen is a virtual tank reproducing the natural environment where the fish lives in presence of it preys and predators. The behavior of every fish in the virtual tank is modeled. The project is based on two original elements: the automatic detection and recognition of fish species in a remote tank of an aquarium and the behavioral modeling of virtual fishes by multi-agents method. 
\end{itemize}

\subsection{Vision-based Hand Gesture Recognition} 
\label{P0C2:sec:HGR}
In vision-based hand gesture recognition, it is needed to detect, track and recognize hand gesture for several applications such as human-computer interface, behavior studies, sign language interpretation and learning, teleconferencing, distance learning, robotics, games selection and object manipulation in virtual environments. We have classified them as follows:

\begin{itemize}
\item \textbf{Human-Computer Interface:} 
Common HCI techniques still rely on simple devices such as keyboard, mice, and joysticks, which are not enough to convoy the latest technology. Hand gesture has become one of the most important attractive alternatives to existing traditional HCI techniques. Practically, hand gesture detection for HCI is achieved using real-time video streaming by removing the background using a background algorithm. Then, every hand gesture can be used  for augmented screen as in Park and Hyun \cite{P0C0-A-370} or for computer interface in vision-based hand gesture recognition as in Stergiopoulou et al. \cite{P0C0-A-371}. 
\item \textbf{Behavior Analysis:}  Perrett et al. \cite{P0C0-A-374} analyzed which vehicle occupant is interacting with a control on the center console when it is activated, enabling the full use of dual-view touch screens and the removal of duplicate controls. The proposed method is first based on hands detection made by a background subtraction algorithm incorporating information from a superpixel segmentation stage. Practically, Perrett et al. \cite{P0C0-A-374} chose PBAS \cite{P2C1-230} as background subtraction algorithm because it allows small foreground objects to decay into the background model quickly whilst larger objects persist, and superpixel Simple Linear Iterative Clustering (SLIC) algormothm as the superpixel segmentation method. Experimental results \cite{P0C0-A-374} on the centers panel of a car show that the hands can be effectivly detected both in day and night conditions.
\item \textbf{Sign Language Interpretation and Learning:}  Elsayed et al. \cite{P0C0-A-372} proposed to detect moving hand area precisely in a real time video sequence using a threshold based on skin color values to improve the segmentation process. The initial background is the first frame without foreground objects. Then, the foreground detection is obtained with a threshold on the difference between the background and the current frame in YCrCb color space. Experimental results  \cite{P0C0-A-372} on indoor and outdoor scenes show that this method can efficiently detect the hands. 
\item \textbf{Robotics:} Khaled  et al. \cite{P0C0-A-373} used a running average to detect the hands, and the 1\$ algorithm  \cite{P0C0-A-373} for hand’s template matching. Then, five hand gestures are detected and translated into commands that can be used to control robot movements. 
\end{itemize}

\subsection{Content-based video coding} 
\label{P0C2:sec:CVC}
To generate video contents, videos have to be segmented into video objects and tracked as they transverse across the video frames. The registered background and the video objects are then encoded separately to allow the transmission of video objects only in the case when the background does not change over time as in video surveillance scenes taken by a fixed camera. So, video coding needs an effective method to separate moving objects from static and dynamic environments \cite{P0C0-A-300}\cite{P5C1-MulF-40}. 

\indent For H.264 video coding, Paul et al. \cite{P5C1-VC-1}\cite{P0C0-A-310-1} proposed a video coding method using a reference frame which is the most common frame in scene generated by dynamic background modeling based on the MOG model with decoded pixel intensities instead of the original pixel intensities. Thus, the proposed model focuses on rate-distortion optimization whereas the original MOG primarily focuses on moving object detection. In a further work, Paul et al. \cite{P0C0-A-310-2} presented an arbitrary shaped pattern-based video coding (ASPVC) for dynamic background modeling based on MD-MOG. Even if these dynamic background frame based video coding methods based on MoG based background modeling achieve better rate distortion performance compared to the H.264 standard, they need high computation time, present low coding efficiency for dynamic videos, and prior knowledge requirement of video content. To address these limitations, Chakraborty et al.  \cite{P5C1-VC-2}\cite{P5C1-VC-3} presented an Adaptive Weighted non-Parametric (WNP) background modeling technique, and further embedded it into HEVC video coding standard for better rate-distortion performance. Being non-parametric, WNP outperforms in dynamic background scenarios compared to MoG-based techniques without a priori knowledge of video data distribution. In a further work, Chakraborty et al. \cite{P0C0-A-312} improved WNP by using a scene adaptive non-parametric (SANP) technique developed to handle video sequences with high dynamic background. 

\indent To address the limitations of the H.264/AVC video coding, Zhang et al. \cite{P0C0-A-320-1} presented a coding scheme for surveillance videos captured by fixed cameras. This scheme used a nonparametric background generation proposed by Liu et al. \cite{P0C0-A-320-2}. In a further work, Zhang et al. \cite{P0C0-A-320} proposed a Segment-and-Weight based Running Average (SWRA) method for surveillance video coding. In a similar approach, Chen et al. \cite{P0C0-A-321} used a timely and bit saving background maintenance model. In a further work, Zhang et al. \cite{P0C0-A-323} used a macro-block-level selective background difference coding method (MSBDC). In an other work, Zhang et al. \cite{P5C1-VC-10} presented a Background Modeling based Adaptive Background Prediction (BMAP) method. In an other approach, Zhao et al. \cite{P5C1-VC-11} proposed a background-foreground division based search algorithm (BFDS) to address the limitations of the HECV coding whilst Zhang et al. \cite{P0C0-A-327} used a running average. For moving cameras, Han et al. \cite{P0C0-A-322} proposed to compute a panorama background with motion compensation. 

\indent Because H.264/AVC is not sufficiently efficient for encoding surveillance videos since it not exploits the strong background
temporal redundancy, Chen et al. \cite{P3C1-OA-32} used for the compression the RPCA decomposition model \cite{P3C1-PCP-1} which decomposed a surveillance video into the low-rank component (background), and the sparse component (moving objects). Then, Chen et al. \cite{P3C1-OA-32} developed different coding methods for the two different components by representing the frames of the background by very few independent frames based on their linear dependency. Experimental results \cite{P3C1-OA-32} show that the proposed RPCA method called Low-Rank and Sparse Decomposition (LRSD) outperforms H.264/AVC, up to 3 dB PSNR gain, especially at relatively low bit rate. In an other work, Guo et al. \cite{P3C1-OA-31} trained a background dictionary based on a small number of observed frames, and then separated every frame into the background and motion (foreground) by using the RPCA decomposition model \cite{P3C1-PCP-1}. In a further step, Guo et al. \cite{P3C1-OA-31} stored the compressed motion with the reconstruction coefficient of the background corresponding to the background dictionary. Experimental results \cite{P3C1-OA-31} show that this RPCA method significantly reduces the size of videos while gains much higher PSNR compared to the state-of-the-art codecs. In these RPCA video coding standards, the selection of Lagrange multiplier is crucial to achieve trade-off between the choices of low-distortion and low-bitrate prediction modes, and the rate-distortion analysis shows that a larger Lagrange multiplier should be used if the background in a coding unit took a larger proportion. To take into account this fact, Zhao et al. \cite{P3C1-OA-32} proposed a modified Lagrange multiplier for rate-distortion optimization by performing an in-depth analysis on the relationship between the optimal Lagrange multiplier and the background proportion, and then Zhao et al. \cite{P3C1-OA-32} presented a Lagrange multiplier selection model to obtain the optimal coding performance. Experimental results show that this Adaptive Lagrange Multiplier Coding Method (ALMCM) \cite{P3C1-OA-32} achieves $18.07\%$ bit-rate saving on CIF sequences and $11.88\%$ on SD sequences against the background-irrelevant Lagrange multiplier selection method. 

Table \ref{P0C2:OMC} shows an overview of the different publications in the field with information about the background model, the background maintenance, the foreground detection, the color space and the strategies used by the authors. \\

\subsection{Background substitution} 
This task is also called background cut and video matting. It consist in extracting the foreground from the input video and then combine it with a new background. Thus, background subtraction can be used in the first step as in Huang et al. \cite{P0C0-A-350}. \\

\subsection{Carried baggage detection}
Tzanidou \cite{P0C0-A-600} proposed a carried baggage detection based on background modeling  which used multi-directional gradient and phase congruency. Then, the detected foreground contours are refined by classifying the edge segments as either belonging to the foreground or background. Finally, a contour completion technique by anisotropic diffusion is employed. The proposed method targets cast shadow removal, gradual illumination change invariance, and closed contour extraction. 

\subsection{Fire detection} 
Several fire detection systems \cite{P0C0-A-700}\cite{P0C0-A-710}\cite{P0C0-A-720}\cite{P0C0-A-730} use in their first step background subtraction to detect moving pixels. Second, colors of moving pixels are checked to evaluate if they match to pre-specified fire-colors, then wavelet analysis in both temporal and spatial domains is carried out to determine high-frequency activity as developed in Toreyin et al. \cite{P0C0-A-700}. 

\subsection{OLED defect detection}   
Organic Light Emitting Diode (OLED) is a light-emitting  diode  which is  popular in the display industry due to its advantages such as colorfulness, light weight, large viewing angle, and power efficiency as developed in Jeon and Yoo \cite{P1C2-KDE-171}. But, the complex manufacturing process also produces defects. which may consistently affect the quality and life of the display. in this context, an automated inspection system based on computer vision is needed.  Practically, OLED presents a feature of gray scale and repeating patterns, but significant intensity variations are also presented. Thus, background subtraction can be used for the inspection. For example, KDE based background model \cite{P1C2-KDE-171} 
can be built by  using multiple  repetitive  images around  the  target area. Then,  the model is used  to  compute likelihood 
at each pixel of the target image.

\section{Discussion}
\label{Challenges}

\subsection{Solved and unsolved challenges}
As developed in the previous sections, all these applications present their own characteristics in terms of environments and objects of interest, and then they present a less or more complex challenge for background subtraction. In addition, they are less or more recent with the consequence that there are less or more publications that addressed the corresponding challenges. Thus, in this section, we have grouped in Table \ref{P0C2:Applications} the solved and unsolved challenges by application to give an overview in which applications investigations need to be made. We can see that the main difficult challenges are camera jitter, illumination changes and dynamic backgrounds that occur in outdoor scenes. Future research may concern these challenges.

\begin{table}
\scalebox{0.65}{
\begin{tabular}{|l|l|l|l|} 
\hline
\scriptsize{Applications} &\scriptsize{Scenes} &\scriptsize{Challenges} &\scriptsize{Solved-unsolved} \\
\hline
\hline
\scriptsize{1) Intelligent Visual Surveillance of Human Activities}      & \scriptsize{}       & \scriptsize{}   & \scriptsize{}  \\
\scriptsize{1.1) Traffic Surveillance}  & \scriptsize{Outdoor scenes}    & \scriptsize{Multi-modal backgrounds}  & \scriptsize{Partially solved}  \\
\scriptsize{}                           & \scriptsize{}                  & \scriptsize{Illumination changes}     & \scriptsize{Partially solved}  \\
\scriptsize{}                           & \scriptsize{}                  & \scriptsize{Camera jitter}            & \scriptsize{Partially solved}  \\
\scriptsize{1.2 Airport Surveillance}   & \scriptsize{Outdoor scenes}    & \scriptsize{Illumination changes}     & \scriptsize{Partially solved}  \\
\scriptsize{}                           & \scriptsize{}                  & \scriptsize{Camera jitter}            & \scriptsize{Partially solved}  \\
\scriptsize{}                           & \scriptsize{}                  & \scriptsize{Illumination changes}     & \scriptsize{Partially solved}  \\
\scriptsize{1.3) Maritime Surveillance} & \scriptsize{Outdoor scenes}    & \scriptsize{Multimodal backgrounds}   & \scriptsize{Partially solved}  \\
\scriptsize{}                           & \scriptsize{}                  & \scriptsize{Illumination changes}     & \scriptsize{Partially solved}  \\
\scriptsize{}                           & \scriptsize{}                  & \scriptsize{Camera jitter}            & \scriptsize{Partially solved}  \\
\scriptsize{1.4) Store Surveillance}    & \scriptsize{Indoor scenes}     & \scriptsize{Multimodal backgrounds}   & \scriptsize{Splved} \\
\scriptsize{1.5) Coastal Surveillance}  & \scriptsize{Outdoor scenes}    & \scriptsize{Multimodal backgrounds}   & \scriptsize{Partially solved}  \\
\scriptsize{1.6) Swimming Pools Surveillance}  & \scriptsize{Outdoor scenes}    & \scriptsize{Multimodal backgrounds}   & \scriptsize{Solved}  \\
\cline{2-4}
\scriptsize{2) Intelligent Visual Observation of Animal and Insect Behaviors} & \scriptsize{}    & \scriptsize{}  & \scriptsize{}  \\
\scriptsize{2.1) Birds Surveillance}     & \scriptsize{Outdoor scenes}    & \scriptsize{Multimodal backgrounds}   & \scriptsize{Partially solved}  \\
\scriptsize{}                            & \scriptsize{}                  & \scriptsize{Illumination changes}     & \scriptsize{Partially solved}  \\
\scriptsize{}                            & \scriptsize{}                  & \scriptsize{Camera jitter}            & \scriptsize{Partially solved}  \\
\scriptsize{2.2) Fish Surveillance}      & \scriptsize{Aquatic scenes}    & \scriptsize{Multimodal backgrounds}   & \scriptsize{Partially solved}  \\
\scriptsize{}                            & \scriptsize{}                  & \scriptsize{Illumination changes}     & \scriptsize{Partially solved}  \\
\scriptsize{2.3) Dolphins Surveillance}  & \scriptsize{Aquatic scenes}    & \scriptsize{Multimodal backgrounds}   & \scriptsize{Partially solved}  \\
\scriptsize{}                            & \scriptsize{}                  & \scriptsize{Illumination changes}     & \scriptsize{Partially solved}  \\
\scriptsize{2.4) Honeybees Surveillance} & \scriptsize{Outdoor scenes}    & \scriptsize{Small objects}            & \scriptsize{Partially solved}  \\
\scriptsize{2.5) Spiders Surveillance}   & \scriptsize{Outdoor scenes}    & \scriptsize{}                         & \scriptsize{Partially solved}  \\
\scriptsize{2.6) Lizards Surveillance}   & \scriptsize{Outdoor scenes}    & \scriptsize{Multimodal backgrounds}   & \scriptsize{Partially solved}  \\
\scriptsize{2.7) Pigs Surveillance}      & \scriptsize{Indoor scenes}     & \scriptsize{Illumination changes}     & \scriptsize{Partially solved}  \\
\scriptsize{2.7) Hinds Surveillance}     & \scriptsize{Outdoor scenes}    & \scriptsize{Multimodal backgrounds}   & \scriptsize{Partially solved}  \\
\scriptsize{}                            & \scriptsize{}                  & \scriptsize{Low-frame rate}           & \scriptsize{Partially solved}  \\
\cline{2-4}
\scriptsize{3) Intelligent Visual Observation of Natural Environments}  & \scriptsize{}    & \scriptsize{}   & \scriptsize{}  \\
\scriptsize{3.1) Forest}   & \scriptsize{Outdoor scenes}    & \scriptsize{Multimodal backgrounds}   & \scriptsize{Partially solved}  \\
\scriptsize{}              & \scriptsize{}                  & \scriptsize{Illumination changes}     & \scriptsize{Partially solved}  \\
\scriptsize{}              & \scriptsize{}                  & \scriptsize{Low-frame rate}           & \scriptsize{Partially solved}  \\
\scriptsize{3.2) River}    & \scriptsize{Aquatic scenes}    & \scriptsize{Multimodal backgrounds}   & \scriptsize{Partially solved}  \\
\scriptsize{}              & \scriptsize{}                  & \scriptsize{Illumination changes}     & \scriptsize{Partially solved}  \\
\scriptsize{3.3) Ocean}    & \scriptsize{Aquatic scenes}    & \scriptsize{Multimodal backgrounds}   & \scriptsize{Partially solved}  \\
\scriptsize{}              & \scriptsize{}                  & \scriptsize{Illumination changes}     & \scriptsize{Partially solved}  \\
\scriptsize{3.4) Submarine}  & \scriptsize{Aquatic scenes}  & \scriptsize{Multimodal backgrounds}   & \scriptsize{Partially solved}  \\
\scriptsize{}              & \scriptsize{}                  & \scriptsize{Illumination changes}     & \scriptsize{Partially solved}  \\
\cline{2-4}
\scriptsize{4) Intelligent Analysis of Human Activities} & \scriptsize{}    & \scriptsize{}         & \scriptsize{}  \\
\scriptsize{4.1) Soccer}   & \scriptsize{Outdoor scenes}    & \scriptsize{Small objects}            & \scriptsize{Solved}  \\
\scriptsize{}              & \scriptsize{}                  & \scriptsize{Illumination changes}     & \scriptsize{Solved}  \\
\scriptsize{4.2) Rowing}   & \scriptsize{Indoor scenes}     & \scriptsize{}                         & \scriptsize{Solved}  \\
\scriptsize{4.3) Surf}     & \scriptsize{Aquatic scenes}    & \scriptsize{Dynamic backgrounds}      & \scriptsize{Partially solved}  \\
\scriptsize{}              & \scriptsize{}                  & \scriptsize{Illumination changes}     & \scriptsize{Partially solved}  \\
\cline{2-4}
\scriptsize{5) Visual Hull Computing} & \scriptsize{}    & \scriptsize{}   & \scriptsize{}  \\
\scriptsize{Image-based Modeling}    & \scriptsize{Indoor scenes}    & \scriptsize{Shadows/highlights}   & \scriptsize{Solved}       \\
\scriptsize{Optical Motion Capture}  & \scriptsize{Indoor scenes}    & \scriptsize{Shadows/highlights}   & \scriptsize{Solved (SG)}  \\
\cline{2-4}
\scriptsize{6) Human-Machine Interaction (HMI)} & \scriptsize{}    & \scriptsize{}   & \scriptsize{}  \\
\scriptsize{Arts}                               & \scriptsize{Indoor scenes}    & \scriptsize{}   & \scriptsize{}  \\
\scriptsize{Games}                              & \scriptsize{Indoor scenes}    & \scriptsize{}   & \scriptsize{}  \\
\scriptsize{Ludo-Multimedia}                    & \scriptsize{Indoor scenes/Outdoor scenes}       & \scriptsize{}   & \scriptsize{}  \\
\cline{2-4}
\scriptsize{7) Vision-based Hand Gesture Recognition}            & \scriptsize{}    & \scriptsize{}   & \scriptsize{}  \\
\scriptsize{Human Computer Interface (HCI)}         & \scriptsize{Indoor scenes}    & \scriptsize{}   & \scriptsize{}  \\
\scriptsize{Behavior Analysis}    & \scriptsize{Indoor scenes/Outdoor scenes}       & \scriptsize{}   & \scriptsize{}  \\
\scriptsize{Sign Language Interpretation and Learning} & \scriptsize{Indoor scenes/Outdoor scenes}  & \scriptsize{}   & \scriptsize{}\\
\scriptsize{Robotics}             & \scriptsize{Indoor scenes}                      & \scriptsize{}   & \scriptsize{}  \\
\cline{2-4}
\scriptsize{7) Content based Video Coding}  & \scriptsize{}    & \scriptsize{Indoor scenes/Outdoor scenes}   & \scriptsize{}  \\
\hline\end{tabular}}
\caption{Solved and unsolved issues : An Overview} \centering
\label{P0C2:Applications}
\end{table}

\subsection{Prospective solutions}
Prospective solutions to handle the unsolved challenges can be the use of recent background subtraction methods based on fuzzy models \cite{P2C2-10}\cite{P5C1-FA-12}\cite{P5C1-FA-13}\cite{P5C1-FA-33}, RPCA models \cite{P3C1-PCP-940}\cite{P3C1-PCP-941}\cite{P3C1-PCP-942}\cite{P3C1-PCP-9420} and deep learning models \cite{P1C5-2100}\cite{P1C5-2150}\cite{P1C5-2162}\cite{P1C5-2163}\cite{P1C5-2163-1}\cite{P1C5-2167}. Among these recent models, several algorithms are potential usable algorithms for real applications:
\begin{itemize}
\item For fuzzy concepts, the foreground detection based on Choquet integral was tested with success for moving vehicles detection by Lu et al. \cite{P5C1-FA-31}\cite{P5C1-FA-34}.
\item For RPCA methods, Vaswani et al. \cite{P3C1-PCP-1030}\cite{P3C1-PCP-1030-1} provided a full study of robust subspace learning methods for background subtraction in terms of detection and algorithms's properties. Among online algorithms, incPCP\protect\footnotemark[11] algorithm \cite{P3C1-PCP-412} and its corresponding improvements \cite{P3C1-PCP-1061}\cite{P3C1-PCP-1062}\cite{P3C1-PCP-1063}\cite{P3C1-PCP-1064}\cite{P3C1-PCP-1064-1}\cite{P3C1-PCP-1065}\cite{P3C1-PCP-1068}\cite{P3C1-PCP-1069} as well as the ReProCS\protect\footnotemark[12] algorithm \cite{P3C1-PCP-1010} and its numerous variants \cite{P3C1-PCP-1011}\cite{P3C1-PCP-1019}\cite{P3C1-PCP-1020}\cite{P3C1-PCP-1027-1} present both advantages in terms of detection, real-time and memory requirements. In traffic surveillance, incPCP was tested with success for vehicle counting \cite{P3C1-PCP-1066}\cite{P3C1-PCP-1067} whilst an online RPCA algorithm for vehicle and person detection \cite{P3C1-PCP-1070}. In animals surveillance, Rezaei and Ostadabbas \cite{P3C1-RMC-91}\cite{P3C1-RMC-92} provided a fast Robust Matrix Completion (fRMC) algorithm for in-cage mice detection using the Caltech resident intruder mice dataset. 
\item For deep learning, only Bautista et al. \cite{P1C5-2120} tested the convolutional neural network for vehicle detection in low resolution traffic videos. But, even if the recent deep learning methods can be naturally considered due there robustness in presence of the concerned unsolved challenges, most of the time they are still to time and memory consuming to be currently employed in real application cases. 
\end{itemize}
Moreover, it is also interesting to consider improvements of the current used models (MOG \cite{P1C2-MOG-10}, codebook \cite{P1C6-100}, ViBe \cite{P2C1-200}, PBAS \cite{P2C1-230}). Instead of the original MOG, codebook and ViBe algorithms employed as in most of the reviewed works in this paper, several improvements of MOG \cite{P1C2-MOG-769}\cite{P1C2-MOG-770}\cite{P1C2-MOG-771}\cite{P1C2-MOG-773}\cite{P1C2-MOG-776}\cite{P1C2-MOG-777} as well as codebook \cite{P1C6-166}\cite{P1C6-172}\cite{P1C6-177}\cite{P1C6-190}, ViBe \cite{P2C1-217}\cite{P2C1-219}\cite{P2C1-229}\cite{P2C1-229-1} and PBAS \cite{P2C1-232} algorithms are potential usable methods for these real applications. For example, Goyal and Singhai \cite{P7C1-6000} evaluated six improvements of MOG on the CDnet 2012 dataset showing that Shah et al.'s MOG \cite{P1C2-MOG-717} and Chen et Ellis'MOG \cite{P1C2-MOG-628-1} both published in 2014 achieve significantly better detection while being usable in real applications than previously published MOG algorithms, that are MOG in 1999, Adaptive GMM {P1C2-MOG-92} in 2003, Zivkovic-Heijden GMM  \cite{P1C2-MOG-36-1} in 2004, and Effective GMM \cite{P1C2-MOG-126} in 2005. Furthermore, there also exist real-time implementation of MOG \cite{P5C3-GPU-20}\cite{P5C3-GPU-23}\cite{P5C3-EI-2}\cite{P5C3-EI-3}\cite{P5C3-AI-101}, codebook \cite{P5C3-PI-10}, ViBe \cite{P2C1-216}\cite{P2C1-216-1} and PBAS \cite{P2C1-235}\cite{P2C1-236}\cite{P2C1-237}. In addition, robust background initialization methods \cite{P1C5-2130}\cite{P1C5-2164} as well as robust deep learned features \cite{P1C5-2200}\cite{P1C5-2210} with the MOG model could also be considered for very challenging environments like maritime and submarine environments.

\footnotetext[11]{https://sites.google.com/a/istec.net/prodrig/Home/en/pubs/incpcp}
\footnotetext[12]{http://www.ece.iastate.edu/~hanguo/PracReProCS.html}

\subsection{Datasets for Evaluation}
In this part, we quickly survey available datasets to evaluate algorithms in similar conditions than the real ones. For visual surveillance of human activities, there are several available dataset. Fist, Toyama et al. \cite{P6C2-Dataset-1} provided in 1999 the Wallflower dataset but it was limited to person detection with one Ground-Truth (GT) image by video. Li et al.  \cite{P6C2-Dataset-10} developed a more larger dataset called I2R dataset with videos with both persons and cars in indoor and outdoor scenes. But, this dataset did not cover a very large spectrum of challenges and the GTs are also limited to 20 by video. In 2012, a breakthrough was done by the BMC 2012 dataset and especially by the CDnet 2012 dataset  \cite{P6C2-Dataset-1000} that are very realistic large scale datasets with a big amount of videos and corresponding GTs. In 2014, the CDnet 2012 dataset was extended with additional camera-captured videos (~70,000 new frames) spanning 5 new categories, and became the CDnet 2014 dataset \cite{P6C2-Dataset-1001}. In addition, there also dataset for RGB-D videos (SBM-RGBD dataset \cite{P6C2-Dataset-110}), infrared videos (OTCBVS 2006), and 
multi-spectral videos (FluxData FD-1665 \cite{P6C2-Dataset-200}). For visual surveillance of animals and insects, there are very less datasets. At the best of our knowledge, there are the following main datasets that are 1) Aqu@theque \cite{P6C2-Dataset-2000} for fish in tank, Fish4knowledge \cite{P6C2-Dataset-2020} for fish in open sea, 2) the Caltech resident intruder mice dataset \cite{P6C2-Dataset-20000} for social behavior recognition of mice, 3) the Caltech Camera Traps (CCT\protect\footnotemark[13]) dataset \cite{P7C1-200} which contains sequences of images taken at approximately one frame per second for census and recognition of speciesna d 4) the eMammal datasets which also camera trap sequences. All the link to these datasets are available on the Background Subtraction Website\protect\footnotemark[14]. Practically, we can note the absence of a large-scale dataset for visual surveillance of animals and insects, and for visual surveillance of natural environments.

\footnotetext[13]{https://beerys.github.io/CaltechCameraTraps/}
\footnotetext[14]{https://sites.google.com/site/backgroundsubtraction/test-sequences}

\subsection{Libraries  for Evaluation}
Most of the time, authors as for example Wei et al. \cite{P7C1-5001} in 2018 employed one of the three background subtraction algorithms based on MOG that are available in OpenCV\protect\footnotemark[15], or provided an evaluation of these three algorithms in the context of traffic surveillance like in Marcomini and Cunha \cite{P7C1-6002} in 2018. But, these algorithms are less efficient that more recent algorithms available in the BGSLibrary\protect\footnotemark[16] and LRS Library\protect\footnotemark[17]. Indeed, BGSLibrary \cite{P7C1-7000}\cite{P7C1-7001} provides a C++ framework to perform background subtraction with currently more than $43$ background subtraction algorithms. In addition, Sobral \cite{P7C1-7010} provided a study of five algorithms from BGSLibrary in the context of highway traffic congestion classification showing that these recent algorithms are more efficient than the three background subtraction algorithms available in OpenCV. For LRSLibrary, it is implemented in MATLAB and focus on decomposition in low-rank and sparse components. The LRSLibrary this provides a collection of state-of-the-art matrix-based and tensor-based factorization algorithms. Several algorithms can be implemented in C to reach real-time requirements.
  
\footnotetext[15]{https://opencv.org/}
\footnotetext[16]{https://github.com/andrewssobral/bgslibrary}
\footnotetext[17]{https://github.com/andrewssobral/lrslibrary}

\section{Conclusion}
\label{Conclusion}
In this review, we have firstly presented all the applications where background subtraction is used to detect static or moving objects of interest. The, we reviewed the challenges related to the different environments and the different moving objects. Thus, the following conclusions can be made: 
\begin{itemize}
\item  All these applications show the importance of the moving object detection in video as it is the first step that is followed by tracking, recognition or behavior analysis. So, the foreground mask needs to be the most precise as possible and quickly as possible for a issue of real-time constraint. A valuable study of the influence of background subtraction on the further steps can be found in Varona et al. \cite{P0C2-2070}. 
\item These different applications present several specificities and need to deal with specific critical situations due to \textbf{(1)} the location of the cameras which can generate small or large moving objects to detect in respect of the size of the images, \textbf{(2)} the type of the environments, and \textbf{(3)} the type of the moving objects. 
\item Because the environments are very different, the background model needs to handle different challenges following the application. Furthermore, the moving objects of interest present very different intrinsic properties in terms of appearance. So, it is required to develop specific background models for a specific application or to find a universal background model which can be used in all the applications. To have a universal background model, the best way may to develop a dedicated background model for particular challenges, and to pick up the suitable background model when the corresponding challenges are detected. 
\item Basic models are sufficiently robust for applications which are in controlled environments such as optical motion capture in indoor scenes. For traffic surveillance, statistical models offer a suitable framework but challenges such as illumination changes and sleeping/beginning foreground objects need to add specific developments. For natural environments and in particular maritime and aquatic environments, more robust background methods than the top methods of ChangeDetection.net competition are required for maritime and submarine environment as developed in Prasad et al. \cite{P0C0-A-29-2} in 2017. Thus, recent RPCA and deep learning models have to be considered for this kind of environments.
\end{itemize}

\bibliographystyle{plain}
\bibliography{ChapterP0C2/mybibP0C0,ChapterP0C2/mybibP0C1,ChapterP0C2/mybibP0C2,ChapterP0C2/mybibP0C3,ChapterP0C2/mybibP1C1,ChapterP0C2/mybibP1C2,ChapterP0C2/mybibP1C3,ChapterP0C2/mybibP1C4,ChapterP0C2/mybibP1C5,ChapterP0C2/mybibP1C6,ChapterP0C2/mybibP1C7,ChapterP0C2/mybibP2C0,ChapterP0C2/mybibP2C1,ChapterP0C2/mybibP2C2,ChapterP0C2/mybibP2C3,ChapterP0C2/mybibP2C4,ChapterP0C2/mybibP2C5,ChapterP0C2/mybibP3C1,ChapterP0C2/mybibP3C2,ChapterP0C2/mybibP3C3,ChapterP0C2/mybibP3C4,ChapterP0C2/mybibP3C5,ChapterP0C2/mybibP4C1,ChapterP0C2/mybibP4C2,ChapterP0C2/mybibP4C2-1,ChapterP0C2/mybibP4C3,ChapterP0C2/mybibP5C1-1,ChapterP0C2/mybibP5C1-2,ChapterP0C2/mybibP5C1-3,ChapterP0C2/mybibP5C1-4,ChapterP0C2/mybibP5C1-5,ChapterP0C2/mybibP5C2,ChapterP0C2/mybibP5C3,ChapterP0C2/mybibP6C1,ChapterP0C2/mybibP6C2,ChapterP0C2/mybibP6C3,ChapterP0C2/mybibP7C7}

\end{document}